\documentclass[10pt,twocolumn,letterpaper]{article}
\newcommand{\ARXIV}[2]{#1} %

\usepackage{templates/ICCV/iccv}
\iccvfinalcopy
\makeatletter%
\@namedef{ver@everyshi.sty}{}
\makeatother%

\PassOptionsToPackage{hyphens}{url}
\usepackage{hyperref}
\hypersetup{pagebackref=true,breaklinks=true,colorlinks,bookmarks=false}

\usepackage{times}
\usepackage{epsfig}
\usepackage{graphicx}
\usepackage{amsmath}
\usepackage{amssymb}
\usepackage{times}
\usepackage{graphicx}
\usepackage{mathtools}
\usepackage{amssymb}
\usepackage{booktabs}
\usepackage{multirow}
\usepackage{tabularx}
\usepackage{xcolor}
\usepackage{lipsum}
\usepackage{soul}
\usepackage{filecontents}

\usepackage{balance}
\usepackage{placeins}
\usepackage{subcaption}
\usepackage{enumitem}
\usepackage{arydshln}
\usepackage{algorithm}
\usepackage{algpseudocode}
\usepackage{tikz}
\usepackage{pgfplots}
\usepackage{styles}
\usepackage{efbox}
\usepackage{microtype}
\usepackage{subcaption}
\usetikzlibrary{shapes,plotmarks,positioning, decorations.pathreplacing, calc, intersections, pgfplots.groupplots,arrows.meta}

\usepackage[hang,flushmargin]{footmisc}
\usepackage[accsupp]{axessibility}  %
\efboxsetup{linecolor=white,linewidth=0.5pt,margin=-.25pt}

\def\iccvPaperID{1800}

\begin{document}

\title{Efficient 3D Semantic Segmentation with \METHOD}

\author{
    Damien Robert\textsuperscript{1, 2}\\
    {\tt\small damien.robert@ign.fr}
    \and
    Hugo Raguet\textsuperscript{3}\\
    {\tt\small hugo.raguet@insa-cvl.fr}
    \and
    Loic Landrieu\textsuperscript{2,4}\\
    {\tt\small loic.landrieu@enpc.fr}
    \and
    {\textsuperscript{1}CSAI, ENGIE Lab CRIGEN, France}\\
    {\textsuperscript{2} LASTIG, IGN, ENSG, Univ Gustave Eiffel, France}\\
    {\textsuperscript{3}INSA Centre Val-de-Loire Univ de Tours, LIFAT, France}\\
    {{\textsuperscript{4}LIGM, Ecole des Ponts, Univ Gustave Eiffel, CNRS, France}}\\
}

\maketitle

\begin{abstract}
We introduce a novel superpoint-based transformer architecture for efficient semantic segmentation of large-scale 3D scenes. Our method incorporates a fast algorithm to partition point clouds into a hierarchical superpoint structure, which makes our preprocessing 
$7$ times faster than existing superpoint-based approaches. Additionally, we leverage a self-attention mechanism to capture the relationships between superpoints at multiple scales, leading to state-of-the-art performance on three challenging benchmark datasets: S3DIS (76.0\% mIoU 6-fold validation), KITTI-360 (63.5\% on Val), and DALES (79.6\%). 
With only $212$k parameters, our approach is up to $200$ times more compact than other state-of-the-art models while maintaining similar performance. Furthermore, our model can be trained on a single GPU in $3$ hours for a fold of the S3DIS dataset, which is $7\times$ to $70\times$ fewer GPU-hours than the best-performing methods.
Our code and models are accessible at \GITHUB.
\end{abstract}
\setlength{\parskip}{-0.13em}

\section{Introduction}
As the expressivity of deep learning models increases rapidly, so do their complexity and resource requirements \cite{owidartificialintelligence}. In particular, vision transformers have demonstrated remarkable results for 3D point cloud semantic segmentation \cite{zhao2021point,park2022fast,guo2021pct,lai2022stratified,loiseau2022online}, but their high computational requirements make them challenging to train effectively. 
Additionally, these models rely on regular grids or point samplings, which do not adapt to the varying complexity of 3D data: the same computational effort is allocated everywhere, regardless of the local geometry or radiometry of the point cloud.
This issue leads to needlessly high memory consumption, limits the number of points that can be processed simultaneously, and hinders the modeling of long-range interactions.

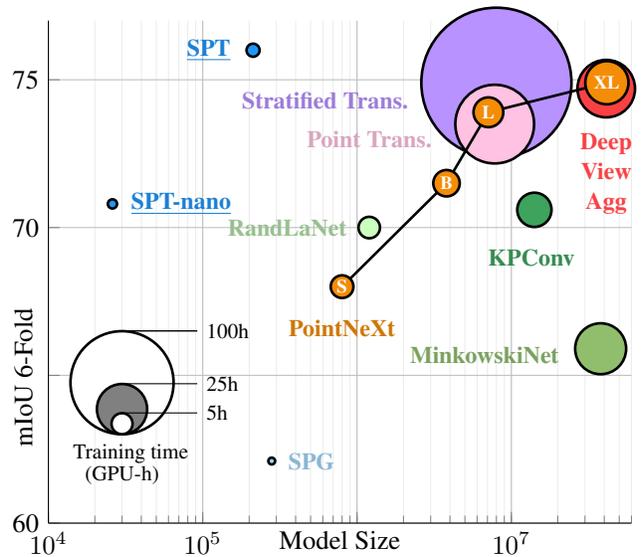
\begin{figure}[t]
\centering
\hspace{-0.40cm}
\begin{tikzpicture}
\begin{semilogxaxis}[%
    width=.93\linewidth,
    scale only axis,
    ytick={60,65,70,75},
    yticklabels={60,,70,75},
    xtick={10000, 100000,1000000,10000000,100000000},
    xticklabels={$10^4$,$10^5$,,$10^7$,$10^8$},
    xmin=10000,
    xmax=60000000,
    ymin=60,
    ymax=77,
    axis x line*=bottom,
    axis y line*=left,
    grid={major},
    xminorgrids=true,
    minor grid style={gray!15,line width=.25pt},
    legend pos=south east,
    ylabel={mIoU 6-Fold},
    xlabel={Model Size},
    ylabel style={xshift=-1.4cm,yshift=-0.9cm},
    xlabel style={xshift=-0cm,yshift=+0.55cm},
    clip marker paths=false,
    clip mode=individual,
    ]
]

\addplot[
    scatter=true,
    line width=0pt,
    mark=*,
    nodes near coords*={\size},
    scatter/@pre marker code/.code={
    \def\markopts{fill=SPTCOLOR, draw opacity=\teaseropacity, line width=1pt}
    \expandafter\scope\expandafter[\markopts]},
    scatter/@pre marker code/.append style={/tikz/mark size=2*sqrt(\size)-1pt},
    scatter/@post marker code/.code={\endscope},
    visualization depends on={value \thisrow{f} \as \size} 
] table [x={s},y={m}] {
    f        m          s
    3.02     76.0      212000
};

\addplot[
    scatter=true,
    line width=0pt,
    mark=*,
    nodes near coords*={\size},
    scatter/@pre marker code/.code={
    \def\markopts{fill=SPTCOLOR, draw opacity=\teaseropacity, line width=1pt}
    \expandafter\scope\expandafter[\markopts]},
    scatter/@pre marker code/.append style={/tikz/mark size=2*sqrt(\size)-1pt},
    scatter/@post marker code/.code={\endscope},
    visualization depends on={value \thisrow{f} \as \size} 
] table [x={s},y={m}] {
    f        m          s
    1.88     70.8      26000
};
\node[SPTCOLOR!85!black, left] at (axis cs:212000,76.0){\small{\bf \underline{SPT}}\;\;};
\node[SPTCOLOR!85!black, right] at (axis cs:30000,70.8){\small{\bf \underline{SPT-nano}}};

\addplot[
    scatter=true,
    line width=0pt,
    mark=*,
    nodes near coords*={\size},
    scatter/@pre marker code/.code={
    \def\markopts{fill=SPGCOLOR, draw opacity=\teaseropacity, line width=1pt}
    \expandafter\scope\expandafter[\markopts]},
    scatter/@pre marker code/.append style={/tikz/mark  size=2*sqrt(\size)-1pt},
    scatter/@post marker code/.code={\endscope},
    visualization depends on={value \thisrow{f} \as \size} 
] table [x={s},y={m}] {
    f         m       s
    1.3       62.1       280000
};
\node[SPGCOLOR!85!black,right] at (axis cs:280000,62.1){\;\small{\bf SPG}};

\addplot[
    scatter=true,
    line width=0pt,
    mark=*,
    nodes near coords*={\size},
    scatter/@pre marker code/.code={
    \def\markopts{fill=KPCONVCOLOR,, draw opacity=\teaseropacity, line width=1pt}
    \expandafter\scope\expandafter[\markopts]},
    scatter/@pre marker code/.append style={/tikz/mark  size=2*sqrt(\size)-1pt},
    scatter/@post marker code/.code={\endscope},
    visualization depends on={value \thisrow{f} \as \size} 
] table [x={s},y={m}] {
    f         m       s
    14      70.6      14100000
};
\node[KPCONVCOLOR!85!black,,below] at (axis cs:13500000,69.6){\small{\bf KPConv}};

\addplot[
    scatter=true,
    line width=0pt,
    mark=*,
    nodes near coords*={\size},
    scatter/@pre marker code/.code={
    \def\markopts{fill=STRATTRANSCOLOR, draw opacity=\teaseropacity, line width=1pt}
    \expandafter\scope\expandafter[\markopts]},
    scatter/@pre marker code/.append style={/tikz/mark  size=2*sqrt(\size)-1pt},
    scatter/@post marker code/.code={\endscope},
    visualization depends on={value \thisrow{f} \as \size} 
] table [x={s},y={m}] {
    f         m       s
    216     74.9     8000000
};
\node[STRATTRANSCOLOR!85!black,left] at (axis cs:3000000, 74.3){\small{\bf Stratified Trans.\;\;}};

\addplot[
    scatter=true,
    line width=0pt,
    mark=*,
    nodes near coords*={\size},
    scatter/@pre marker code/.code={
    \def\markopts{fill=RANDLACOLOR, draw opacity=\teaseropacity, line width=1pt}
    \expandafter\scope\expandafter[\markopts]},
    scatter/@pre marker code/.append style={/tikz/mark  size=2*sqrt(\size)-1pt},
    scatter/@post marker code/.code={\endscope},
    visualization depends on={value \thisrow{f} \as \size} 
] table [x={s},y={m}] {
    f         m       s
    6.4     70.0     1200000
};
\node[RANDLACOLOR!75!black,left] at (axis cs:1200000, 70.0){\small{\bf RandLaNet\;\;}};

   \addplot[
    scatter=true,
    line width=0pt,
    mark=*,
    nodes near coords*={\size},
    scatter/@pre marker code/.code={
    \def\markopts{fill=POINTTRANSCOLOR, draw opacity=\teaseropacity, line width=1pt}
    \expandafter\scope\expandafter[\markopts]},
    scatter/@pre marker code/.append style={/tikz/mark  size=2*sqrt(\size)-1pt},
    scatter/@post marker code/.code={\endscope},
    visualization depends on={value \thisrow{f} \as \size} 
] table [x={s},y={m}] {
    f         m       s
    63.12     73.5     7800000
};
\node[POINTTRANSCOLOR!85!black,left] at (axis cs:3500000, 73){\small{\bf Point Trans.}};

\addplot[
    scatter=true,
    line width=0pt,
    mark=*,
    nodes near coords*={\size},
    scatter/@pre marker code/.code={
    \def\markopts{fill=MINKOCOLOR, draw opacity=\teaseropacity, line width=1pt}
    \expandafter\scope\expandafter[\markopts]},
    scatter/@pre marker code/.append style={/tikz/mark  size=2*sqrt(\size)-1pt},
    scatter/@post marker code/.code={\endscope},
    visualization depends on={value \thisrow{f} \as \size} 
] table [x={s},y={m}] {
    f         m       s
    28.3      65.9    37900000  
};
\node[MINKOCOLOR!85!black,,left] at (axis cs:37900000,65.7){\small{\bf MinkowskiNet \;\;\;\;}};

\addplot[
    scatter=true,
    line width=0pt,
    mark=*,
    nodes near coords*={\size},
    scatter/@pre marker code/.code={
    \def\markopts{fill=DVACOLOR, draw opacity=\teaseropacity, line width=1pt}
    \expandafter\scope\expandafter[\markopts]},
    scatter/@pre marker code/.append style={/tikz/mark  size=2*sqrt(\size)-1pt},
    scatter/@post marker code/.code={\endscope},
    visualization depends on={value \thisrow{f} \as \size} 
] table [x={s},y={m}] {
    f         m       s
    35.28     74.7     41200000
};
\node[DVACOLOR,below] at (axis cs:41200000, 73.5){\small{\bf Deep}};
\node[DVACOLOR,below] at (axis cs:41200000, 72.5){\small{\bf View}};
\node[DVACOLOR,below] at (axis cs:41200000, 71.5){\small{\bf Agg}};

\addplot[
    black,
    scatter=true,
    line width=1pt,
    mark=*,
    nodes near coords*={\size},
    scatter/@pre marker code/.code={
    \def\markopts{fill=POINTNEXTCOLOR, draw=black, draw opacity=\teaseropacity, line width=1pt}\expandafter\scope\expandafter[\markopts]},
    scatter/@pre marker code/.append style={/tikz/mark  size=2*sqrt(\size)-1pt},
    scatter/@post marker code/.code={\endscope},
    visualization depends on={value \thisrow{f} \as \size} 
] table [x={s},y={m}] {
    f        m       s
    7       68         800000
    9       71.5       3800000
    11      73.9       7100000
    21      74.9       41600000
};

\node[POINTNEXTCOLOR!85!black,below] at (axis cs:800000, 67.2){\small{\bf PointNeXt}};
\node[white] at (axis cs:800000, 68){\scriptsize{\bf S}};
\node[white] at (axis cs:3800000, 71.5){\scriptsize{\bf B}};
\node[white] at (axis cs:7100000, 73.9){\scriptsize{\bf L}};
\node[white] at (axis cs:41600000, 74.9){\scriptsize{\bf XL}};

\node[draw=black,right, circle, fill=white, minimum size = 20*sqrt(100)-5pt, scale=0.2, anchor=south,line width=1pt] at (axis cs:30000,63) (n1){};
\node[draw=black,right, circle, fill=black!50!white, minimum size = 20*sqrt(25)-5pt, scale=0.2, anchor=south,line width=1pt] at (axis cs:30000,63) (n2) {};
\node[draw=black,right, circle, fill=white, minimum size = 20*sqrt(5)-5pt, scale=0.2, anchor=south,line width=1pt] at (axis cs:30000,63) (n4) {};

\draw [-] (n1.north) -- ++ (1cm,0cm);
\draw [-] (n2.north) -- ++ (1cm,0cm);
\draw [-] (n1.north) -- ++ (1cm,0cm);
\draw [-] (n4.north) -- ++ (1cm,0cm);

\node[draw=none,anchor=north, right=of n1.north] (x) {\footnotesize 100h};
\node[draw=none,anchor=west, right=of n2.north] (x) {\footnotesize  25h};
\node[draw=none,anchor=west, right=of n4.north] (x) {\footnotesize 5h};
\node[draw=none,anchor=south, below=of n4, yshift=1cm] (x) { \footnotesize \;\;\;Training time};
\node[draw=none,anchor=south, below=of n4, yshift=0.7cm] (x) { \footnotesize (GPU-h)};
     
\end{semilogxaxis}
\end{tikzpicture}
\caption{
{\bf Model Size vs. Performance.} We visualize the performance of different methods on the S3DIS dataset (6-fold validation) in relation to their model size in log-scale. The area of the markers indicates the GPU-time to train on a single fold. 
Our proposed method \METHOD (SPT) achieves state-of-the-art with a reduction of up to $200$-fold in model size and $70$-fold in training time (in GPU-h) compared to recent methods. The even smaller SPT-nano model achieves a fair performance with $26$k parameters only.
}
\label{fig:teaser}
\end{figure}

Superpoint-based methods \cite{landrieu2018large,landrieu2019point,hui2021superpoint,quana2016international} address the limitation of regular grids by partitioning large point clouds into sets of points--- superpoints---which adapt to the local complexity. By directly learning the interaction between superpoints instead of individual points, these methods enable the analysis of large scenes with compact and parsimonious models that can be trained faster than standard approaches.
However, superpoint-based methods often require a costly preprocessing step, and their range and expressivity are limited by their use of local graph-convolution schemes \cite{simonovsky2017dynamic}.

In this paper, we propose a novel superpoint-based transformer architecture that overcomes the limitations of both approaches, see \figref{fig:teaser}. 
Our method starts by partitioning a 3D point cloud into a hierarchical superpoint structure that adapts to the local properties of the acquisition at multiple scales simultaneously. To compute this partition efficiently, we propose a new algorithm that is an order of magnitude faster than existing superpoint preprocessing algorithms.
Next, we introduce the \METHOD (SPT) architecture, which uses a sparse self-attention scheme to learn relationships between superpoints at multiple scales. 
By viewing the semantic segmentation of large point clouds as the classification of a small number of superpoints, our model can accurately classify millions of 3D points simultaneously without relying on sliding windows. \SHORTHAND achieves near state-of-the-art accuracy on various open benchmarks while being significantly more compact and able to train much quicker than common approaches. The main contributions of this paper are as follows:
\begin{itemize}[align=right,itemindent=1em,labelsep=3pt,labelwidth=1em,leftmargin=2pt,nosep]
    \item {\bf Efficient Superpoint Computation:} 
    We propose a new method to compute a hierarchical superpoint structure for large point clouds, which is more than 
    {7 times faster}
    than existing superpoint-based methods. Our preprocessing time is also 
    {comparable or faster}
    than standard approaches, addressing a significant drawback of superpoint methods.
    \item {\bf State-of-the-Art Performance:} Our model reaches performance at or close to the state-of-the-art for three open benchmarks with distinct settings: S3DIS for indoor scanning \cite{armeni20163d}, KITTI-360 for outdoor mobile acquisitions \cite{liao2022kitti}, and DALES for city-scale aerial LiDAR \cite{varney2020dales}.
    \item {\bf Resource-Efficient Models:} \SHORTHAND is particularly resource-efficient as it only has $212$k parameters for S3DIS and DALES, a $200$-fold reduction compared to other state-of-the-art models such as PointNeXt \cite{qian2022pointnext} and takes $70$ times
    fewer GPU-h to train than Stratified Transformer \cite{lai2022stratified}. The even more compact SPT-nano reaches $70.8\%$ 6-Fold mIoU on S3DIS with only $26$k parameters, making it the smallest model to reach above $70\%$ by a factor of almost $300$.
\end{itemize}

\section{Related Work}
This section provides an overview of the main inspirations for this paper, which include 3D vision transformers, partition-based methods, and efficient learning for 3D data.

\paragraph{3D Vision Transformers.}
Following their adoption for image processing~\cite{dosovitskiy2020image,liu2021swin}, Transformer architectures~\cite{vaswani2017attention} designed explicitly for 3D analysis have shown promising results in terms of performance \cite{zhao2021point,guo2021pct} and speed \cite{park2022fast,loiseau2022online}.
In particular, the Stratified Transformer of Lai \etal  uses a specific sampling scheme \cite{lai2022stratified} to model long-range interactions. However, the reliance of 3D vision transformers on arbitrary K-nearest or voxel neighborhoods leads to high memory consumption, which hinders the processing of large scenes and the ability to leverage global context cues.

\paragraph{Partition-Based Methods.} 
Partitioning images into superpixels has been studied extensively to simplify image analysis, both before and after the widespread use of deep learning \cite{achanta2012slic, tu2018learning}. 
Similarly, superpoints are used for 3D point cloud segmentation~\cite{papon2013voxel,lin2018toward} and object detection \cite{han2020occuseg,engelmann20203d}. SuperPointGraph \cite{landrieu2018large} proposed to learn the relationship between superpoints using graph convolutions \cite{simonovsky2017dynamic} for semantic segmentation. While this method trains fast,
its preprocessing is slow and its expressivity and range are limited, as it operates on a single partition.
Recent works have proposed ways of learning the superpoints themselves \cite{landrieu2019point,hui2021superpoint,thyagharajan2022segment}, which yields improved results but at the cost of an extra training step or a large point-based backbone~\cite{kang2023region}.

Hierarchical partitions are used for image processing \cite{arbelaez2006boundary,xu2016hierarchical,zhang2022nested} and 3D analysis tasks such as point cloud compression \cite{fan2013point} and object detection \cite{chen2021hierarchical,liang2021instance}. 
Hierarchical approaches for semantic segmentation use Octrees with fixed grids \cite{narasimhamurthy2023hierarchical,Riegler2017OctNet}. On the contrary, \SHORTHAND uses a multi-scale hierarchical structure that adapts to the local geometry of the data. This leads to partitions that conform more closely to semantic boundaries, enabling the network to model the interactions between objects or object parts.

\paragraph{Efficient 3D Learning.}
As 3D scans of real-world scenes can contain hundreds of millions of points, optimizing the efficiency of 3D analysis is an essential area of research. PointNeXt \cite{qian2022pointnext} proposes several effective techniques that allow simple and efficient methods \cite{qi2017pointnetpp} to achieve state-of-the-art performance. RandLANet \cite{hu2020randla} demonstrates that efficient sampling strategies can yield excellent results. Sparse \cite{SubmanifoldSparseConvNetCVPR} or hybrid \cite{liu2019point} point cloud representations have also helped reduce memory usage. However, by leveraging the local similarity of dense point clouds, superpoint-based methods can achieve an input reduction of several orders of magnitude, resulting in unparalleled efficiency.

\section{Method}
\begin{figure}[t]
\centering
\begin{tabular}{@{}c@{}c@{}}
    \begin{subfigure}[b]{0.5\linewidth}
        \includegraphics[width=\columnwidth]{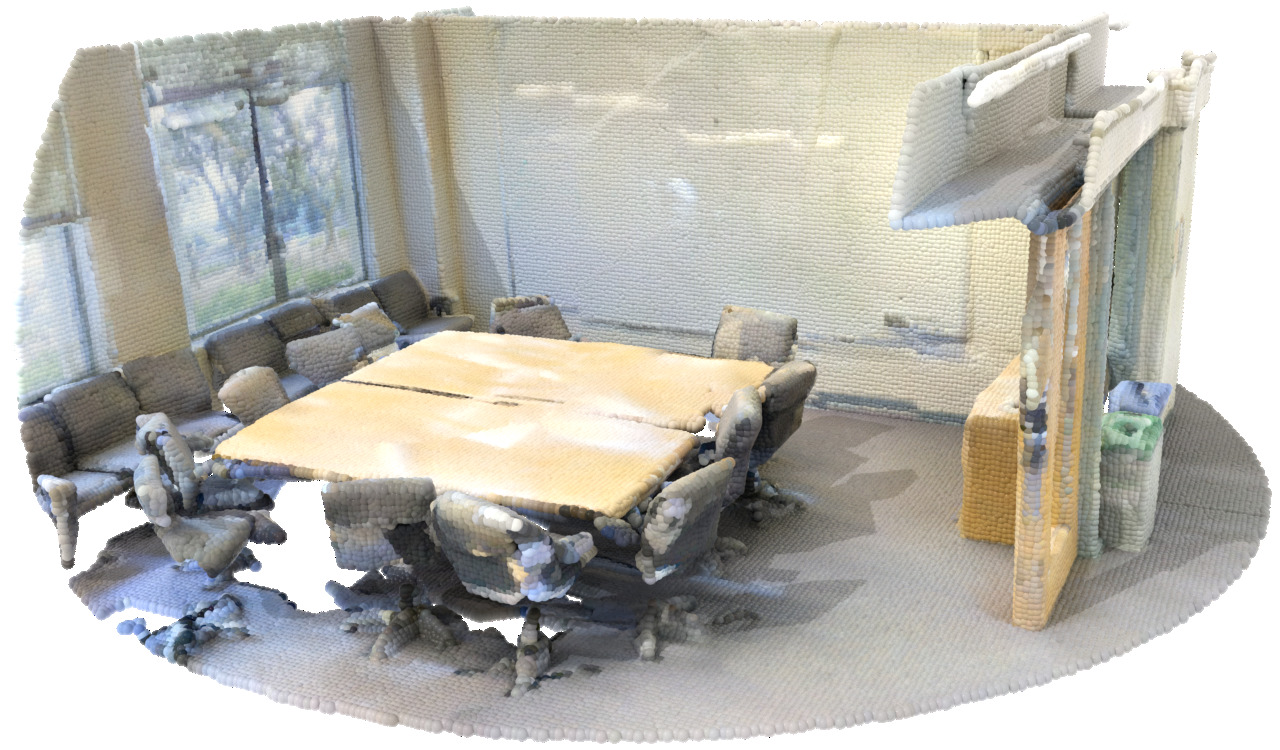}
        \caption{Input point cloud}
        \label{fig:teaser:rgb}
    \end{subfigure}
     &  
     \begin{subfigure}[b]{0.5\linewidth}
        \includegraphics[width=\columnwidth]{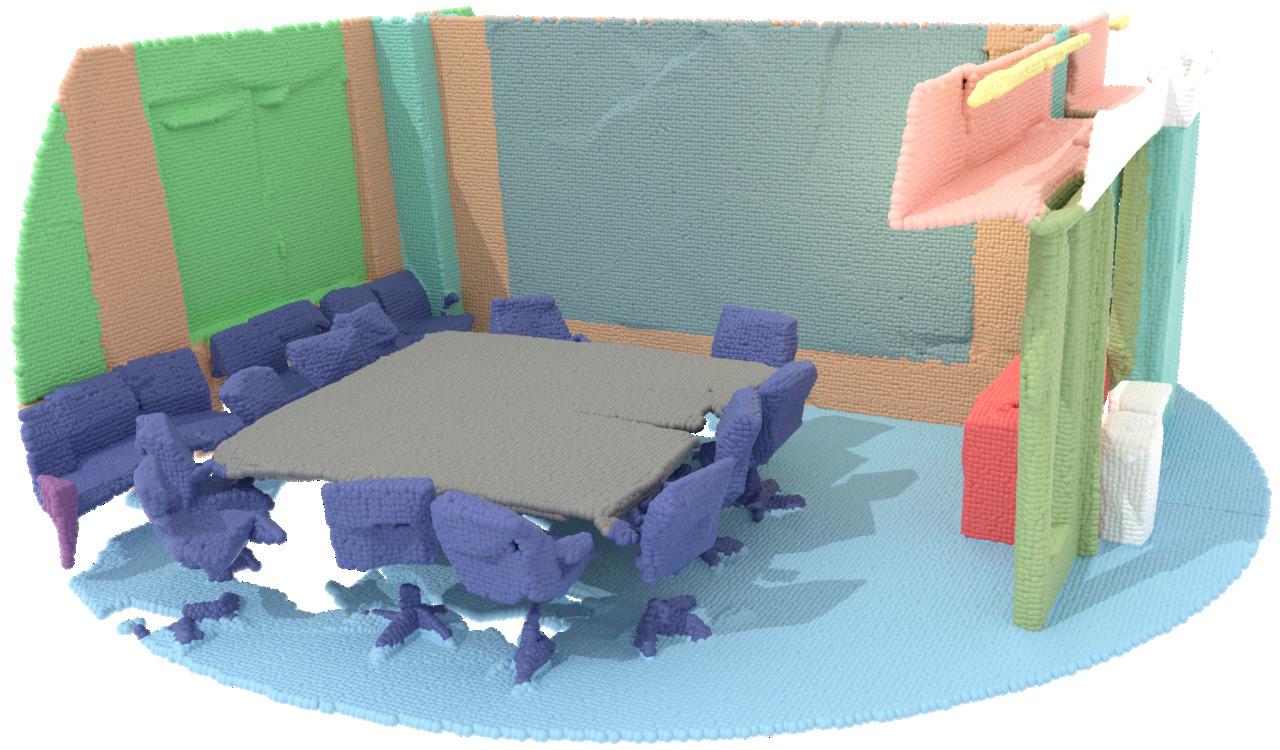}
        \caption{Ground truth labels}
        \label{fig:teaser:gt}
    \end{subfigure}
     \\
     \begin{subfigure}[b]{0.5\linewidth}
        \includegraphics[width=\columnwidth]{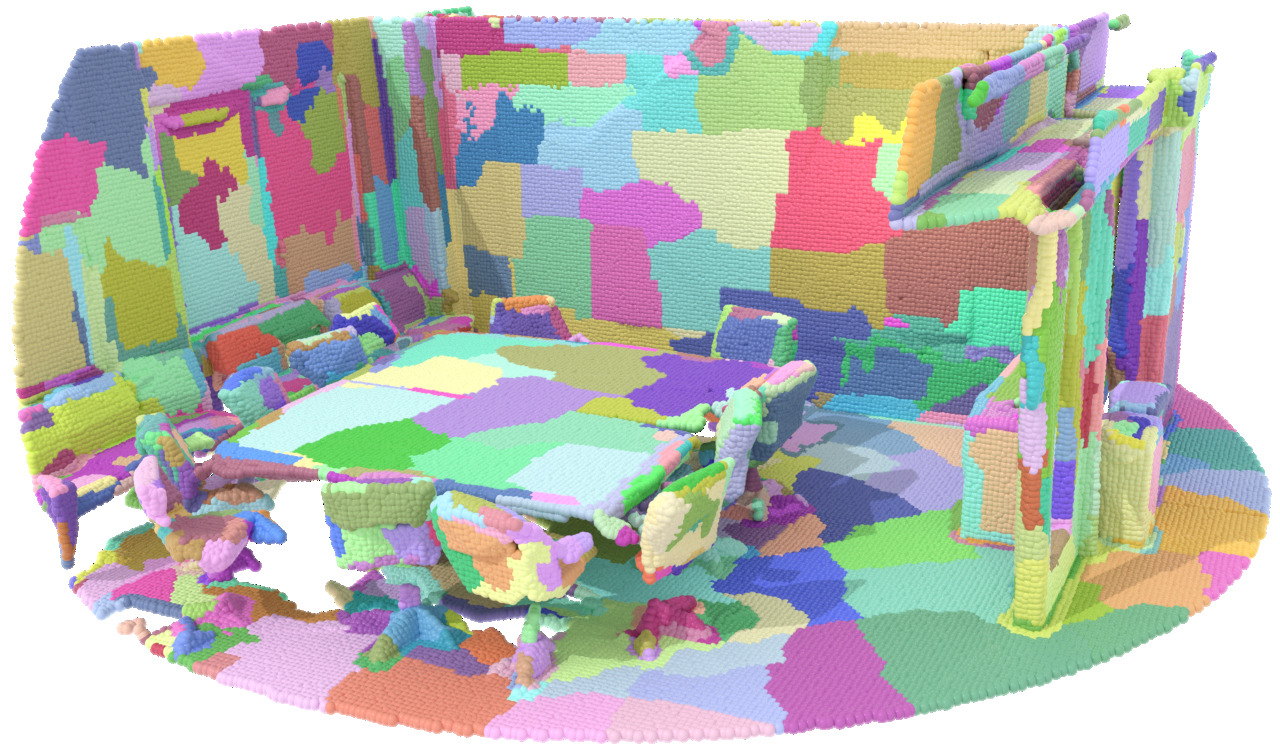}
        \caption{First partition level}
        \label{fig:teaser:pone}
    \end{subfigure}
     &  
     \begin{subfigure}[b]{0.5\linewidth}
        \includegraphics[width=\columnwidth]{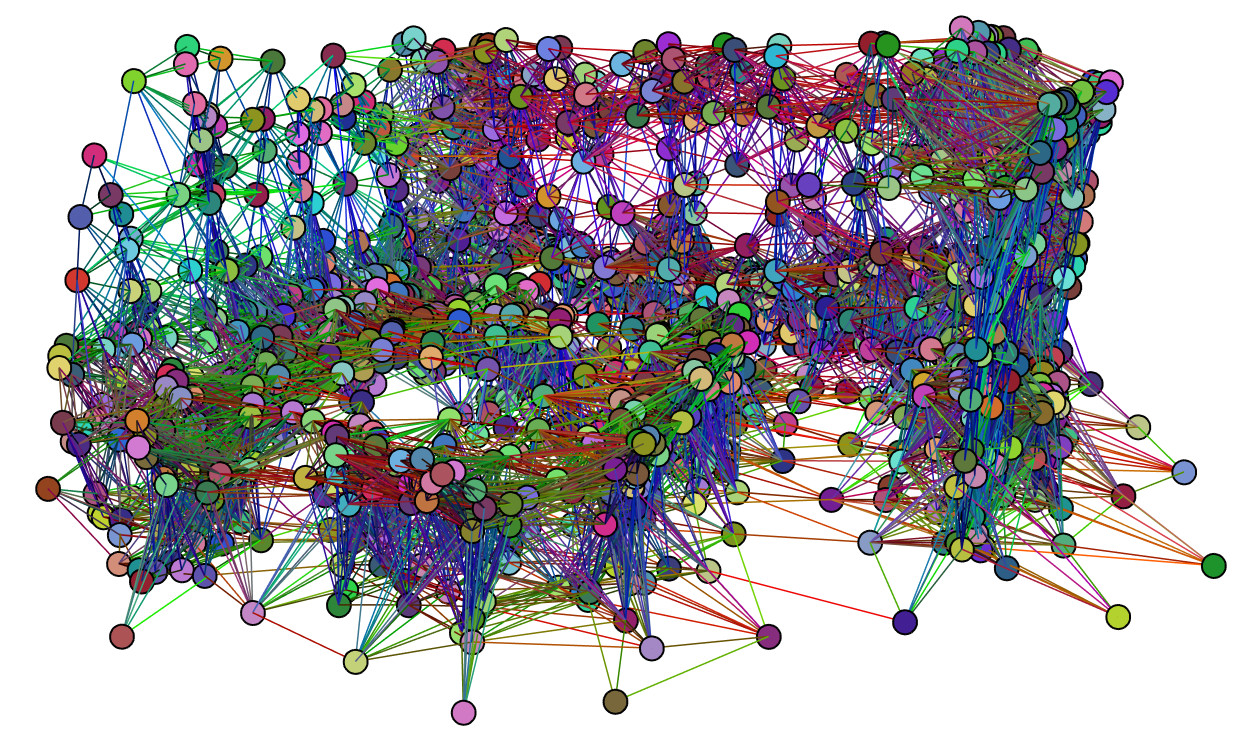}
        \caption{First superpoint-graph}
        \label{fig:teaser:gone}
    \end{subfigure}
     \\
     \begin{subfigure}[b]{0.5\linewidth}
        \includegraphics[width=\columnwidth]{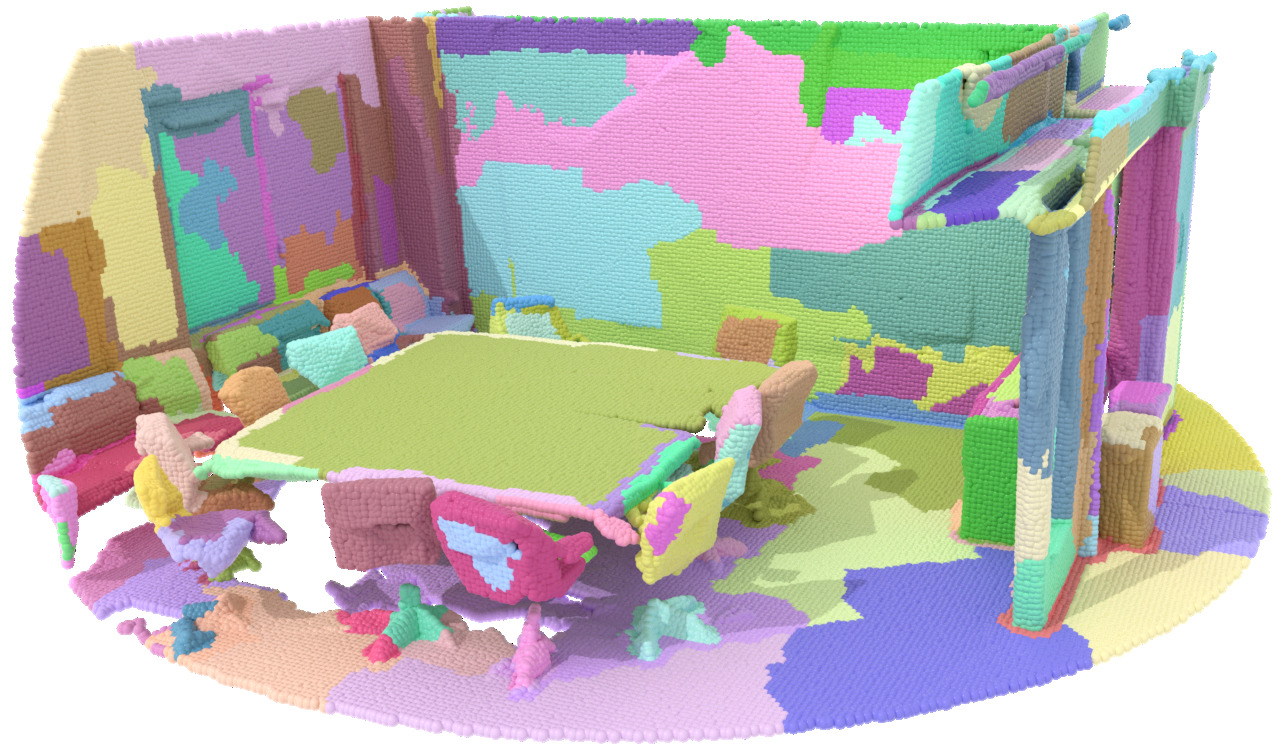}
        \caption{Second partition level}
        \label{fig:teaser:ptwo}
    \end{subfigure}
     &  
     \begin{subfigure}[b]{0.5\linewidth}
        \includegraphics[width=\columnwidth]{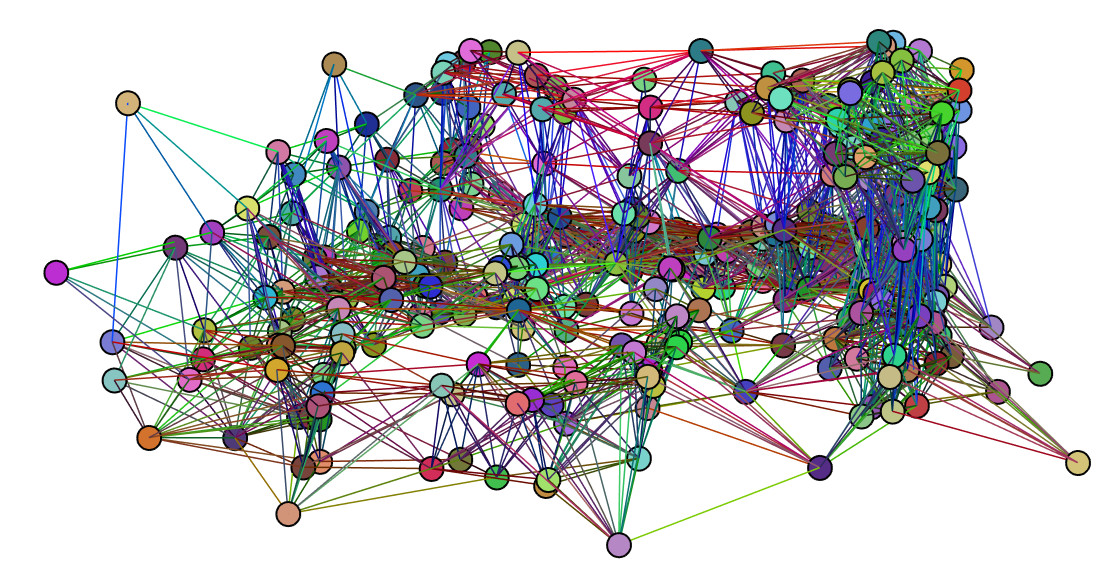}
        \caption{Second superpoint-graph}
        \label{fig:teaser:gtwo}
    \end{subfigure}
\end{tabular}
\caption{
{\bf Superpoint Transformer.} 
Our method takes as input a point cloud \Subref{fig:teaser:rgb} and computes its hierarchical partition into geometrically homogeneous superpoints at multiple scales: \Subref{fig:teaser:pone} and \Subref{fig:teaser:ptwo}. For all partition levels, we construct superpoint adjacency graphs \Subref{fig:teaser:gone} and \Subref{fig:teaser:gtwo}, which are used by an attention-based network to classify the finest superpoints.
} 
\label{fig:graphs}
\end{figure}

Our method has two key components. First, we use an efficient algorithm to segment an input point cloud into a compact multi-scale hierarchical structure. Second, a transformer-based network leverages this structure to classify the elements of the finest scale.

\subsection{Efficient Hierarchical Superpoint Partition}
We consider a point cloud $\cC$ with positional and radiometric information. To learn multiscale interactions, we compute a hierarchical partition of $\cC$ into {geometrically}-homogeneous superpoints of increasing coarseness; see \figref{fig:graphs}. We first define the concept of hierarchical partitions.%
\begin{definition} {\bf Hierarchical Partitions.} 
A partition of a set $\cX$ is a collection of subsets of $\cX$ such that each element of $\cX$ is in one and only one of such subsets. $\cP\eqdef[\cP_0, \cdots, \cP_I]$ is a hierarchical partition of $\cX$ if $\cP_0=\cX$, and $\cP_{i+1}$ is a partition of $\cP_{i}$ for $i\in[0,I-1]$.
\end{definition}

Throughout this paper, all functions or tensors related to a specific partition level $i$ are denoted with an exponent $i$.
\paragraph{Hierarchical Superpoint Partitions.}
We propose an efficient approach for constructing hierarchical partitions of large point clouds. First, we associate each point $c$ of $\cC$ with features $f_c$ representing its local geometric and radiometric information. These features can be handcrafted \cite{guinard2017weakly} or learned \cite{landrieu2019point,hui2021superpoint}.
See the Appendix for more details on point features.
We also define a graph $\cG$ encoding the adjacency between points usually based on spatial proximity, \eg $k$-nearest neighbors.

We view the features $f_c$ for all $c$ of $\cC$ as a signal $f$ defined on the nodes of the graph $\cG$. Following the ideas of SuperPoint Graph \cite{landrieu2018large}, we compute an approximation of $f$ into constant components by solving an energy minimization problem penalized with a graph-based notion of \emph{simplicity}. The resulting constant components form a partition whose granularity is determined by a regularization strength $\lambda>0$: higher values yield fewer and coarser components.

For each component of the partition, we can compute the mean position (centroid) and feature of its elements, defining a coarser point cloud on which we can repeat the partitioning process.
We can now compute a hierarchical partition $\cP\eqdef[\cP_0, \cdots, \cP_I]$ of $\cC$ from a list of regularization strengths $\lambda_1, \cdots, \lambda_{I}$. %
First, we set $\cP_0$ as the point cloud $\cC$ and $f^0$ as the point features $f$. Then, for $i=1$ to $I$, we compute (i) a partition $\cP_{i}$ of $f^{i-1}$ penalized with $\lambda_i$;
(ii) the mean signal $f^i$ for all components of $\cP_{i}$. The coarseness of the resulting partitions $[\cP_0, \cdots, \cP_I]$ is thus strictly increasing.
See the Appendix for a more detailed description of this process.

\paragraph{Hierarchical Graph Structure.} A hierarchical partition defines a polytree structure across the different levels. Let $p$ be an element of $\cP_i$. 
If $i \in [0,I-1]$, $\parent{p}$ is the component of $\cP_{i+1}$ which contains $p$.
If $i \in [1,I]$, $\children{p}$ is the set of components of $\cP_{i-1}$ whose parent is $p$.

Superpoints also share adjacency relationships with superpoints \emph{of the same partition level}. For each level $i\geq 1$, we build a \emph{superpoint-graph} $\cG_i$ by connecting adjacent components of $\cP_i$, \ie superpoints whose closest points are within a distance gap $\epsilon_i>0$.
For $p  \in \cP_i$, we denote $\hop{p} \subset \cP_i$ the set of neighbours of $p$ in the graph $\cG_i$. 
More details on the superpoint-graph construction can be found in the Appendix.

\paragraph{Hierarchical Parallel $\ell_0$-Cut Pursuit.}
Computing the hierarchical components involves solving a recursive sequence of non-convex, non-differentiable optimization problems on large graphs.
We propose an adaptation of the $\ell_0$-cut pursuit algorithm~\cite{landrieu2017cut} to solve this problem. To improve efficiency, we adapt the graph-cut parallelization strategy initially introduced by Raguet \etal \cite{raguet2019parallel} in the convex setting.

\subsection{\METHOD}

\input{figures/architecture}

Our proposed \SHORTHAND architecture draws inspiration from the popular U-Net \cite{ronneberger2015u,gao2019graph}.
However, instead of using grid, point, or graph subsampling, our approach derives its different resolution levels from the hierarchical partition $\cP$.%

\paragraph{General Architecture.}
As represented in \figref{fig:nested}, \SHORTHAND comprises an encoder with $I$ stages and a decoder with $I-1$ stages: the prediction takes place at the level $\cP_1$ and not on individual points. 
We start by computing the relative positions $x$ of all points and superpoints with respect to their parent. 
For a superpoint $p \in \cP_i$, we define $x^i_p$ as the position of the centroid of $p$ relative to its parent's. The coarsest superpoints of $\cP_I$ have no parent and use the center of the scene as a reference centroid.
We then normalize these values so that the sets $\{x^{i}_p | p \in \children{q}\}$ have a radius of $1$ for all $q  \in \cP_{i+1}$.
We compute features for each 3D point by using a multi-layer perceptron (MLP) to mix their relative positions and handcrafted features: ${g}^{0}\eqdef \phi^0_\text{enc}([x^0,f^0])$, with $[\cdot, \cdot]$ the channelwise concatenation operator.

Each level $i\geq 1$ of the encoder maxpools the features of the finer partition level $i-1$, adds relative positions $x^i$ and propagates information between neighboring superpoints in $\cG_i$.
For a superpoint $p$ in $\cP_i$, this translates as:
\begin{align} \label{eq:encoder}
    {g}^{i}_p &= \cT^i_{\text{enc}} \circ \phi^i_\text{enc}\left(\left[x^i_p, \max_{q \in \children{p}}\left({g}^{i-1}_q\right)\right]\right)
\end{align}
with $\phi^i_\text{enc}$ an MLP and
$\cT^i_{\text{enc}}$ a transformer module explained below.
By avoiding communication between the 3D points of $\cP_0$, we bypass a potential computational bottleneck.

The decoder passes information from the coarser partition level $i+1$ to the finer level $i$. It uses the relative positions $x^i$  and the encoder features $g^i$ to improve the spatial resolution of its feature maps $h^i$ \cite{ronneberger2015u}.
For a superpoint $p$ in partition $\cP_i$ with $1\leq i<I-1$, this can be expressed as:
\begin{align} \label{eq:decoder}
    {h}^i_p &= \cT^i_{\text{dec}} \circ \phi^i_\text{dec}\left(\left[x^i_p,g^i_p, h^{i+1}_{\parent{p}}\right]\right)
\end{align}
with $h^I=g^I$, $\phi^i_\text{dec}$ an MLP, and $\cT^i_{\text{dec}}$ an attention-based module similar to $\cT^i_{\text{enc}}$.

\paragraph{Self-Attention Between Superpoints.} We propose a variation of graph-attention networks \cite{velivckovic2018graph} to propagate information between neighboring superpoints of the same partition level.
For each level of the encoder and decoder, we associate to superpoint $p \in \cP_i$ a triplet of key, query, value vectors 
$\key_p, \que_p, \val_p$ of size $D_\text{key}, D_\text{key}$ and $ D_\text{val}$. These values are obtained by applying a linear layer to the corresponding feature map $m$ after GraphNorm normalization \cite{cai2021graphnorm}.

We then characterize the relationship between two superpoints $p,q$ of $\cP_i$ adjacent in $\cG_i$ by a triplet of features $a^\text{key}_{p,q}, a^\text{que}_{p,q},a^\text{val}_{p,q}$ of dimensions $D_\text{key}, D_\text{key}$ and $ D_\text{val}$, and whose computation is detailed in the next section.
Given a superpoint $p$, we stack the vectors $a^\text{key}_{p,q}, a^\text{que}_{p,q},a^\text{val}_{p,q}$ for $q \in \hop{p}$ in matrices  $A^\text{key}_p, A^\text{que}_p,A^\text{val}_p$ of dimensions $\vert  \hop{p} \vert\times D_\text{key}$ or $\vert  \hop{p} \vert\times D_\text{val}$.
The modules $\cT^i_{\text{enc}}$ and $\cT^i_{\text{dec}}$ gather contextual information %
as follows: 
\begin{align}
\!\!\!\!{[\cT(m)]}_p\!\!\plusequal \attention(\que_p^\intercal\oplus\! A^\text{que}_p\!,\key_{\hop{p}}\!+\!A^\text{key}_p\!,\val_{\hop{p}}\!+\!A^\text{val}_p)~,
\end{align}
with $\plusequal$ a residual connection \cite{he2016deep}, $\oplus$ the addition operator with broadcasting on the first dimension, and $\key_{\hop{p}}$ the matrix of stacked vectors $\key_q$ for $q\in \hop{p}$. The attention mechanism writes as follows: 
\begin{align}\label{eq:att}
    &\attention(\que,\key,\val) \eqdef 
    \val^\intercal
    \softmax
    \left(
        \frac{\que \odot \key \mathbf{1}}
             {\sqrt{\vert  \hop{p} \vert} }
    \right)~,
\end{align}
with $\odot$ the Hadamard termwise product and $\mathbf{1}$ a column-vector with $D_\text{key}$ ones. Our proposed scheme is similar to classic attention schemes with two differences: (i) the queries adapt to each neighbor, and (ii) we normalize the softmax with the neighborhood size instead of the key dimension.
In practice, we use multiple independent attention modules in parallel (multi-head attention) and several consecutive attention blocks. 

\subsection{Leveraging the Hierarchical Graph Structure}
The hierarchical superpoint partition $\cP$ can be used for more than guidance for graph pooling operations. Indeed, we can learn expressive adjacency encodings capturing the complex adjacency relationships between superpoints and employ powerful supervision and augmentation strategies based on the hierarchical partitions.

\paragraph{Adjacency Encoding.}
While the adjacency between two 3D points is entirely defined by their distance vector, the relationships between superpoints are governed by additional factors such as their alignment, proximity, and difference in sizes or shapes.
We characterize the adjacency of pairs of adjacent superpoints of the same partition level using a set of  handcrafted features based on: (i) the relative positions of centroids, (ii) position of paired points in each superpoints, (iii) the superpoint principal directions, and (iv) the ratio between the superpoints' length, volume, surface, and point count. These features are efficiently computed only once during preprocessing. 

For each pair of superpoints $(p,q$) adjacent in $\cG_i$,
we jointly compute the concatenated $a^\text{key}_{p,q}, a^\text{que}_{p,q}, a^\text{val}_{p,q}$ by applying an MLP $\phi_\text{adj}^i$ to the handcrafted adjacency features defined above.
Further details on the superpoint-graph construction and specific adjacency features are provided in the Appendix.
\if 1 0
\paragraph{Superpoint Supervision.}
The superpoints are designed to be geometrically and radiometrically homogeneous and are thus generally semantically consistent. We do not make point-level predictions but only classify the much sparser superpoints of $\cP_1$, hence saving computation and memory.
However, superpoints may sometimes overlap different objects, making the supervision of their classification ambiguous. To account for this, we compute for each superpoint $p$ of $\cP_1$ the frequency histograms $z_p$ of the labels of the points it contains. We supervise the prediction $y_c$ with the Kullback–Leibler divergence \cite{joyce2011kullback} with $z_p$ weighted by the number of points $N^1_p$ in the superpoint $p$:
\begin{align}
    \mathcal{L}(y)  = \frac1{\mid\cC\mid}  \sum_{p \in \cP_1} N^1_p \KL(z_p,y_p)~,
\end{align}
with $\mid\cC\mid$ the cardinal of $\cC$.
\fi

\paragraph{Hierarchical Supervision.}
We propose to take advantage of the nested structure of the hierarchical partition $\cP$ into the supervision of our model. 
We can naturally associate the superpoints of any level $i \geq 1$ with a set of 3D points in $\cP_0$.
The superpoints at the finest level $i=1$ are almost semantically pure (see \figref{fig:purity}), while the superpoints at coarser levels $i>1$ typically encompass multiple objects.
Therefore, we use a dual learning objective: (i) we predict the most frequent label within the superpoints of $\cP_1$ %
, and (ii) we predict the label distribution for the superpoints of $\cP_i$ with $i>1$. We supervise both predictions with the cross-entropy loss.%

Let $y^i_p$ denote the true label distribution of the 3D points within a superpoint $p \in \cP_i$, and $\hat{y}^i_p$ a one-hot-encoding of its most frequent label. We use a dedicated linear layer at each partition level to map the decoder feature $g^i_p$ to a predicted label distribution $z^i_p$. Our objective function can be formulated as follows:
\begin{align}
    \!\!\!\mathcal{L}=\sum_{p \in \cP_1}\!\!\frac{-N^1_p}{\mid\cC\mid}H(\hat{y}^1_p,z^1_p)
    \!+\!\!\sum_{i = 2}^I  \sum_{p \in \cP_i} \frac{\mu^i N^i_p}{\mid\cC\mid} H(y^i_p,z^i_p)~,
\end{align}
where $\mu^2, \cdots, \mu^I$ are positive weights, $N^i_p$ represents the number of points within a superpoint $p \in \cP_i$, and $|\cC|$ is the total number of points in the point cloud, and $H(y,z)=-\sum_{k \in \cK}y_k \log (z_k)$ and $\cK$ the class set.

\if 1 0
Since the superpoints of $\cP$ are geometrically homogeneous by construction, they should be somewhat semantically consistent. At the finer levels, the superpoints should mostly represent parts or entire object. Superpoints of coarser levels may contain several distinct but related objects or instances. We can predict their content at all levels simultaneously to achieve supervision across scales.

The decoder yields a feature map $g^i$ for each level $\cP_i$ of the hierarchical partition for $i=1, \cdots, I$. We use a dedicated layer for each partition level to obtain a probabilistic prediction $y^i_p$ of each superpoint $p$ of each level $i$. During inference, we only compute the prediction for $i=1$ and associate each 3D point with the class of its parent in $\cP_1$.
The last layer of the decoder yields a feature map  $g^1$ for all superpoints of the finest partition level $\cP_1$. To save computation and memory, we perform the classification at this level with a linear layer, but not for all individual 3D points. This gives us a prediction $y_c$ in the $K$-simplex for all $c$ of $\cP_1$ with $K$ the size of the class set.
 
The superpoints of the partition $\cP_i$ are not necessarily semantically pure. For each superpoint $p$ of $\cP_i$, we compute the frequency histograms $z^i_p$ of the labels of the points it contains. We supervize the prediction $y^i_c$ with the Kullback–Leibler divergence \cite{joyce2011kullback} with $z^i_p$ weighted by the number of points $N^i_p$ in the superpoint $p$:
\begin{align}
    \mathcal{L}(y)  = \frac1{\mid\cC\mid} \sum_{i = 1}^I \mu^i \sum_{p \in \cP_i} N^i_p \KL(z^i_p,y^i_p)~,
\end{align}
with $\mu^1, \cdots, \mu^I$ positive weights and $\mid\cC\mid$ the cardinal of $\cC$.

\fi

\paragraph{Superpoint-Based Augmentations.}
Although our approach classifies superpoints rather than individual 3D points, we still need to load the points of $\cP_0$ in memory to embed the superpoints from $\cP_1$. However, since superpoints are designed to be geometrically simple, only a subset of their points is needed to characterize their shape. Therefore, when computing the feature $g^1_p$ of a superpoint $p$ of $\cP_1$ containing $n$ points with \eqref{eq:encoder}, we sample only a portion $\tanh(n / n_\text{max})$ of its points, with a minimum of $n_\text{min}$. This sampling strategy reduces the memory load and acts as a powerful data augmentation. The lightweight version of our model \SHORTHANDNANO goes even further. It ignores the points entirely and only use handcrafted features to embed the superpoints of $\cP_1$, thus avoiding entirely the complexity associated with the size of the input point cloud $\cP_0$.

To further augment the data, we exploit the geometric consistency of superpoints and their hierarchical arrangement. During the batch construction, we randomly drop each superpoint with a given probability at all levels. %
Dropping superpoints at the fine levels removes random objects or object parts, while dropping superpoints at the coarser levels removes entire structures such as walls, buildings, or portions of roads, for example.

\section{Experiments}
We evaluate our model on three diverse datasets described in \secref{sec:data}. In \secref{sec:quant}, we evaluate our approach in terms of precision, but also quantify the gains in terms of pre-processing, training, and inference times. Finally, we propose an extensive ablation study in \secref{sec:abla}.

\begin{table}
    \caption{{\bf Partition Configuration.} We report the point count of different datasets %
    before and after subsampling, as well as the size of the partitions.}
    \label{tab:datas}
    \centering\small{
    \begin{tabular}{@{}lllll@{}}
    \toprule
    \multirow{1}{*}{Dataset} & Points & Subsampled & $\mid \cP_1 \mid$ & $\mid \cP_2 \mid$ \\ \midrule
    S3DIS \cite{armeni20163d} & 273m & \;\;\;\;32m\;\; & 979k & 292k \\
    DALES \cite{varney2020dales} & 492m & \;\;\;\;449m & 14.8m & 2.56m  \\
    KITTI-360 \cite{liao2022kitti} & 919m & \;\;\;\;432m & 16.2m & 2.98m \\
    \bottomrule
    \end{tabular}}
\end{table}

\subsection{Datasets and Models}
\label{sec:data}
\paragraph{Datasets.} To demonstrate its versatility, we evaluate \SHORTHAND on three large-scale datasets of different natures.\\% of indoor, outdoor, and aerial scans.
\noindent\textbf{S3DIS} \cite{armeni20163d}. This indoor dataset of office buildings contains over $274$ million points across $6$ building floors---or areas. The dataset is organized by individual rooms, but can also be processed by considering entire areas at once.

\begin{figure*}
    \centering
    \begin{tabular}{@{}lccc@{}}
    \rotatebox{90}{ \qquad\qquad Input}
    &
    \begin{subfigure}[b]{0.28\textwidth}
      \includegraphics[width=\linewidth]{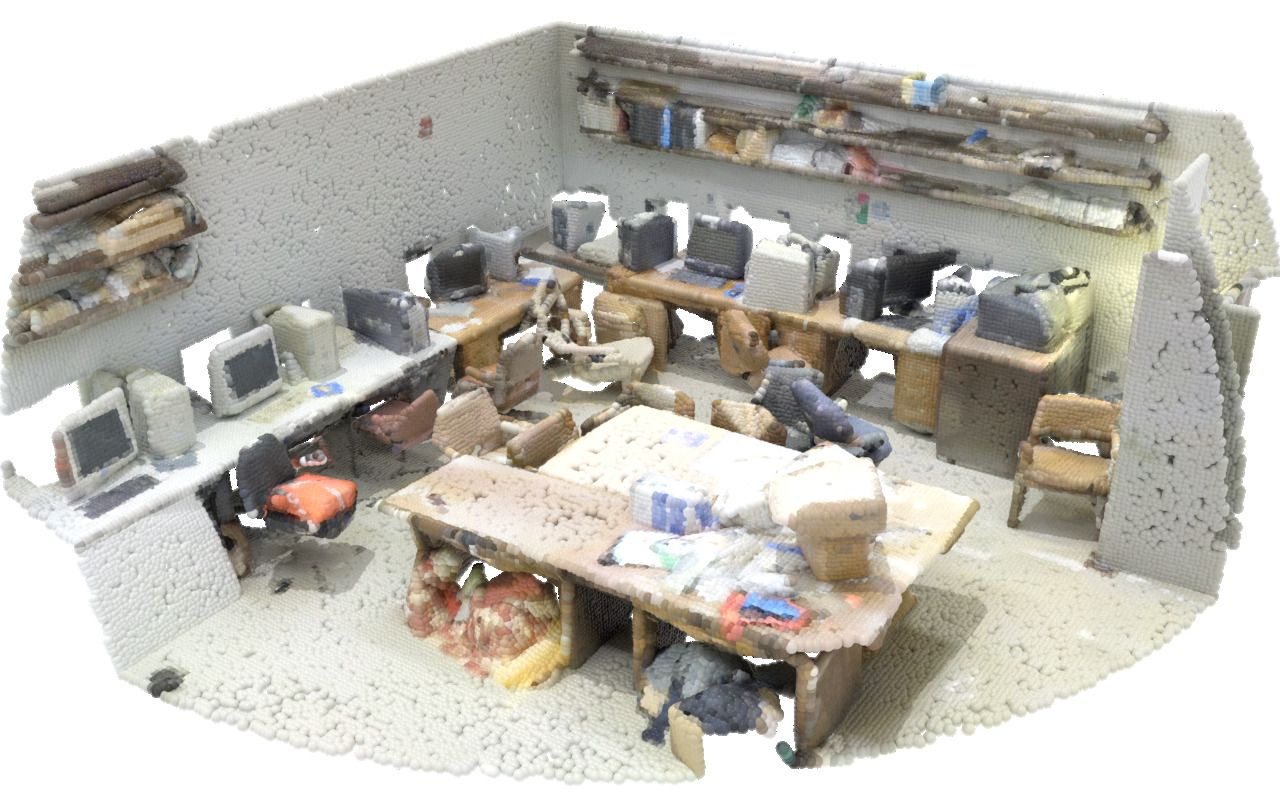}
    \end{subfigure}
         & 
    \begin{subfigure}[b]{0.28\textwidth}
      \includegraphics[width=\linewidth]{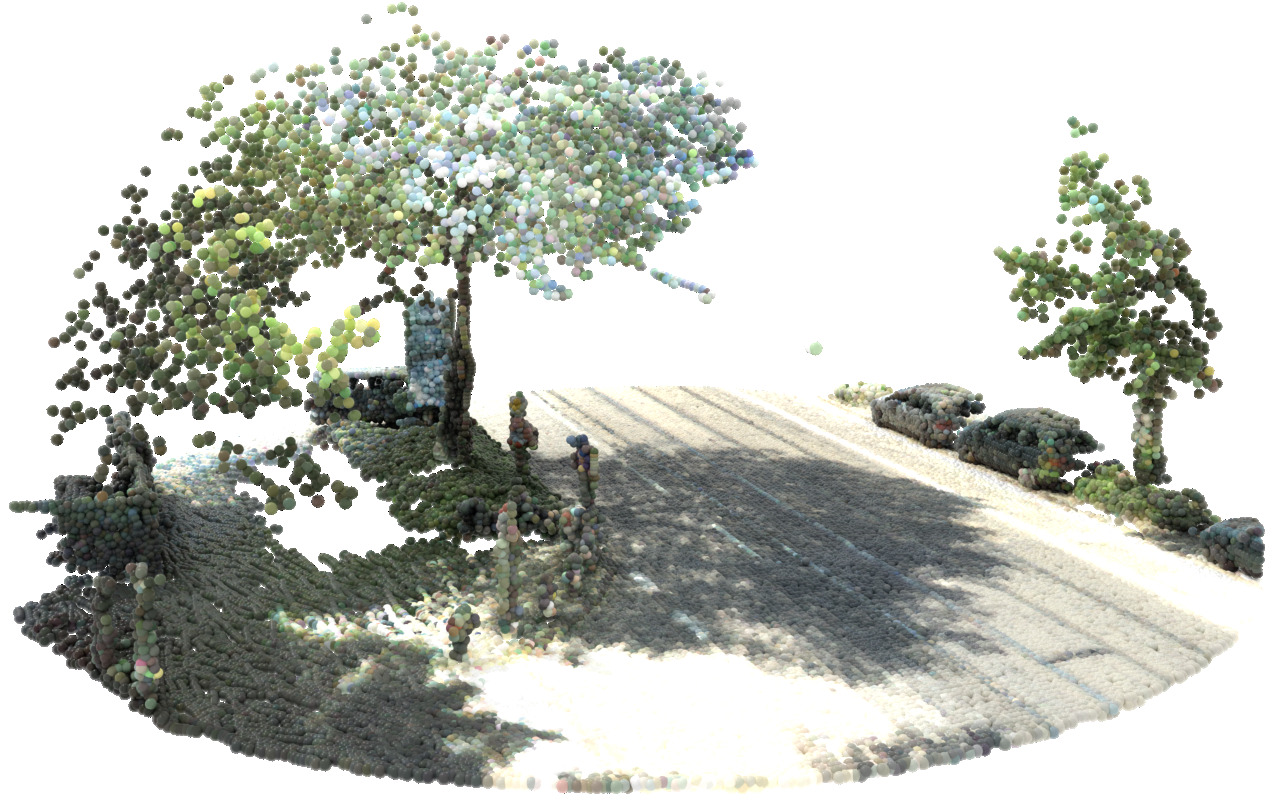}
    \end{subfigure}
         & 
    \begin{subfigure}[b]{0.28\textwidth}
      \includegraphics[width=\linewidth]{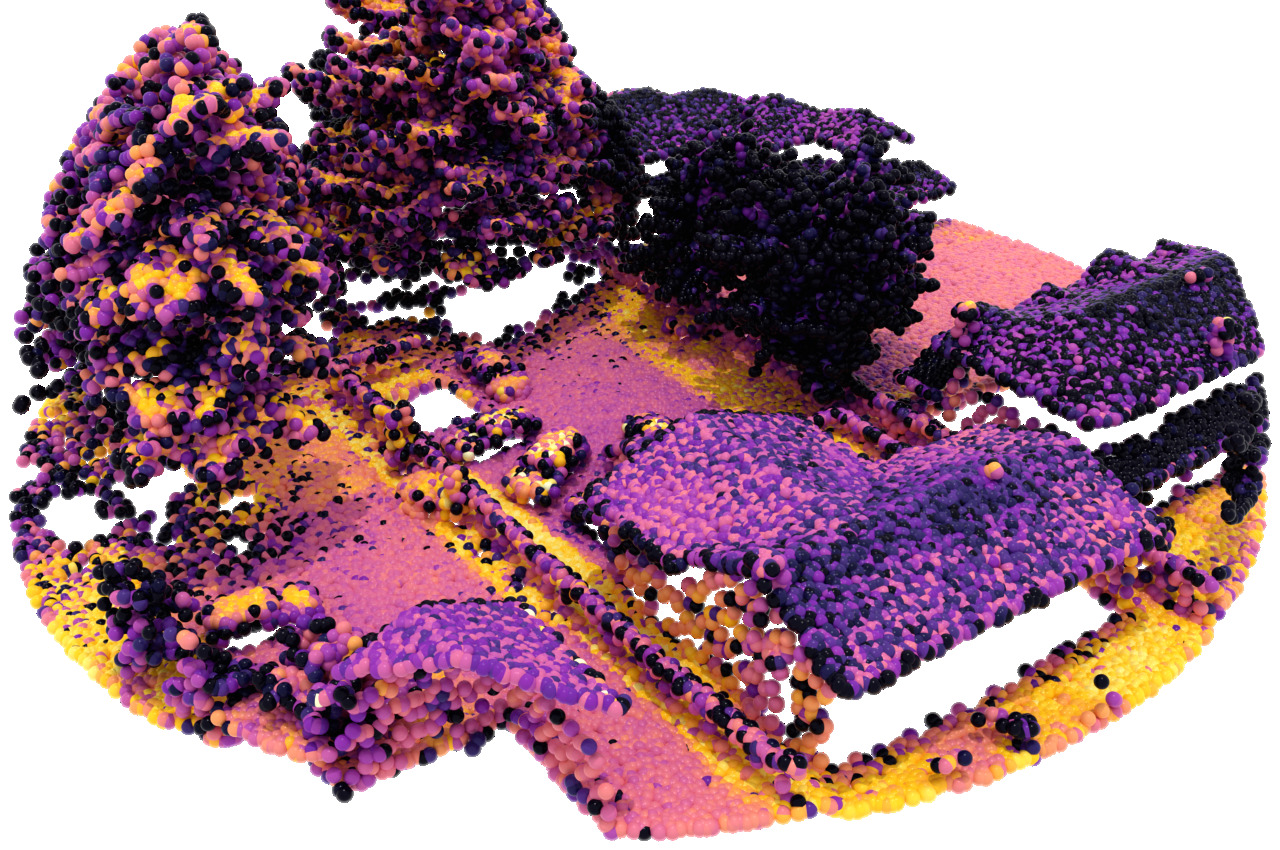}
    \end{subfigure}  \\
    \rotatebox{90}{ \qquad Partition $\cP_2$}
    &
    \begin{subfigure}[b]{0.28\textwidth}
      \includegraphics[width=\linewidth]{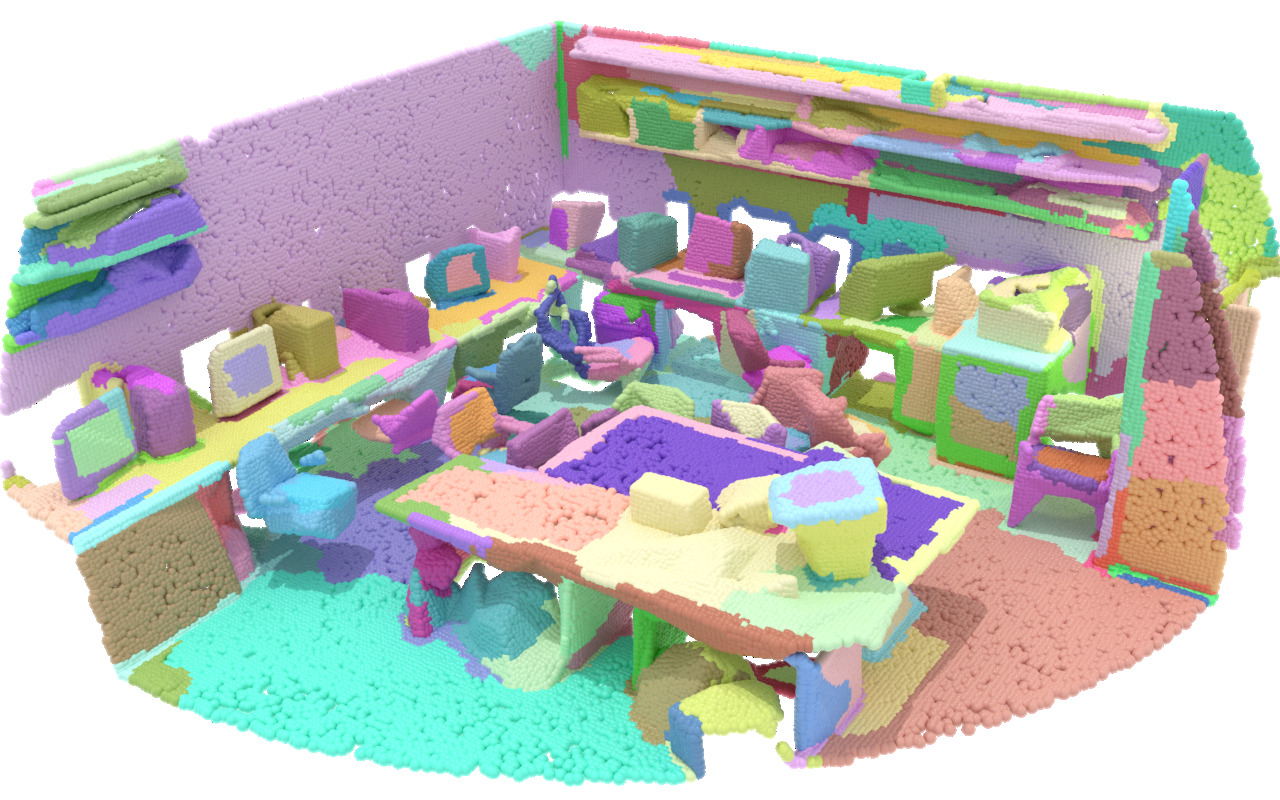}
    \end{subfigure}
         & 
    \begin{subfigure}[b]{0.28\textwidth}
 \begin{tikzpicture}
        \node[anchor=south west,inner sep=0] (image) at (0,0) {\includegraphics[width=\linewidth]{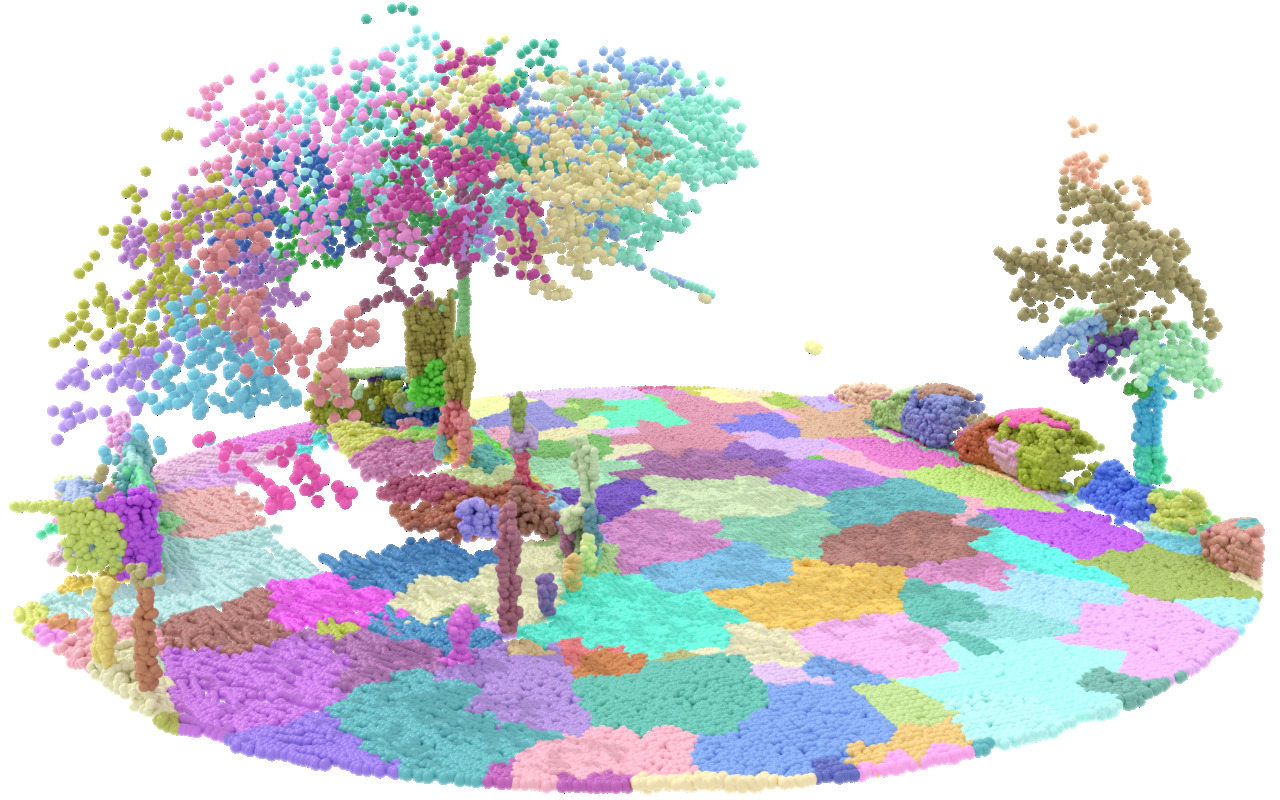}};
        \begin{scope}[x={(image.south east)},y={(image.north west)}]
        \end{scope}
        \end{tikzpicture}
     
    \end{subfigure}
         & 
    \begin{subfigure}[b]{0.28\textwidth}
      \includegraphics[width=\linewidth]{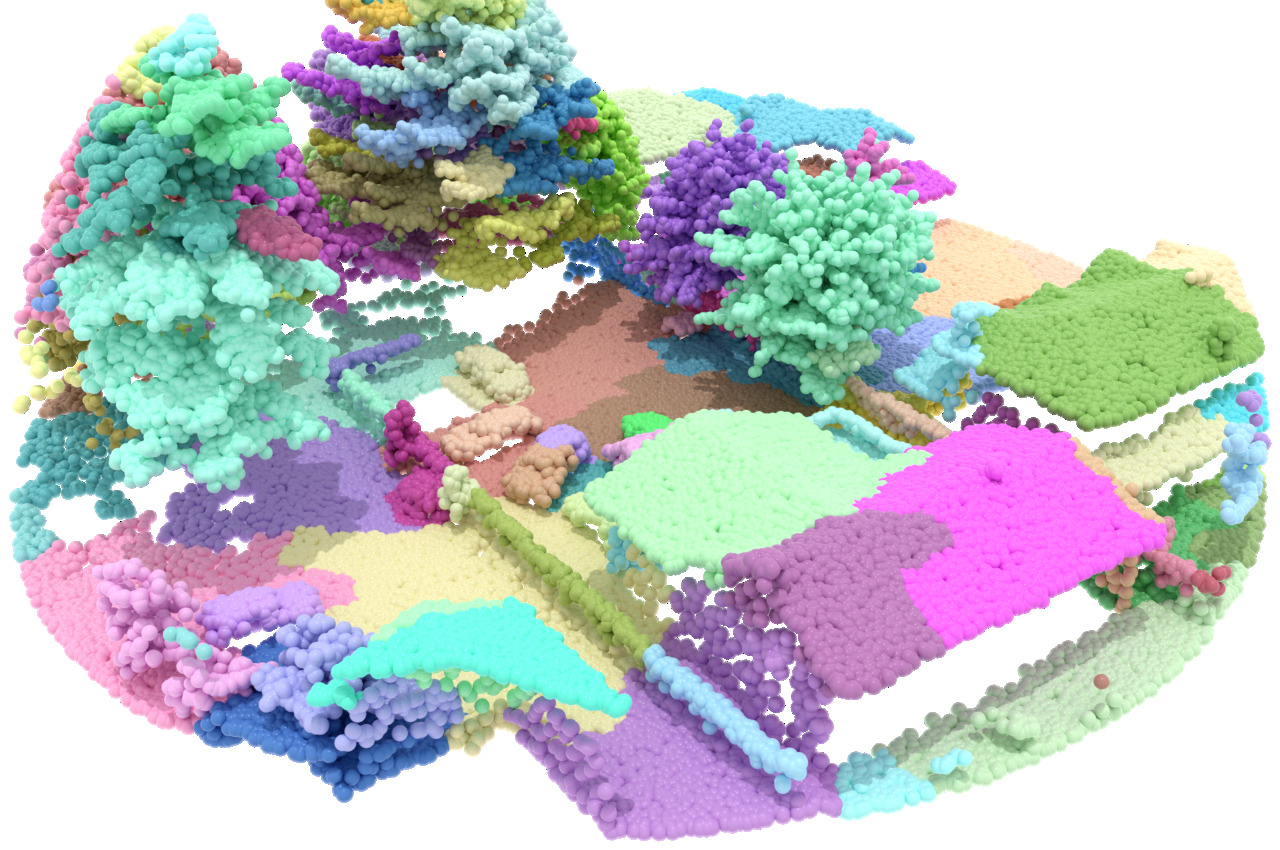}
    \end{subfigure}  \\
    \rotatebox{90}{ \quad Ground Truth}
    &
    \begin{subfigure}[b]{0.28\textwidth}
      \includegraphics[width=\linewidth]{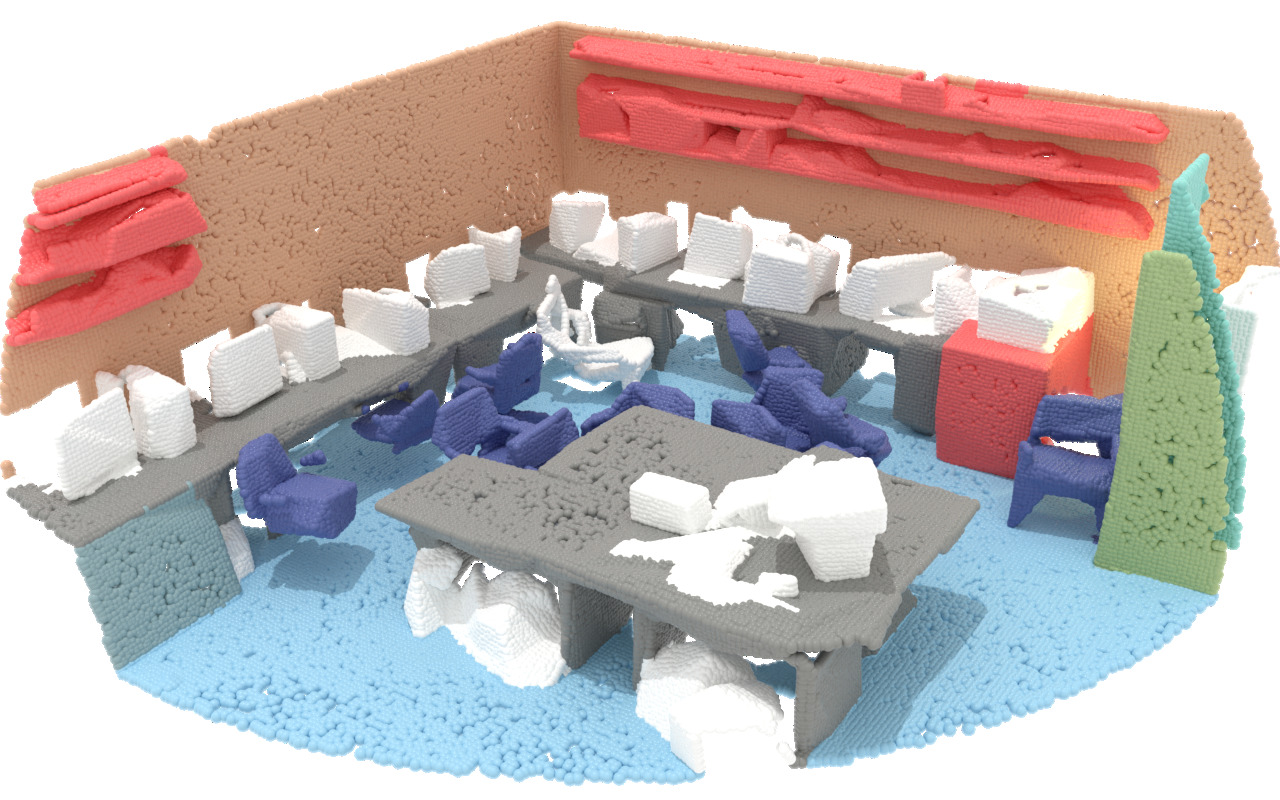}
    \end{subfigure}
         & 
    \begin{subfigure}[b]{0.28\textwidth}
      \includegraphics[width=\linewidth]{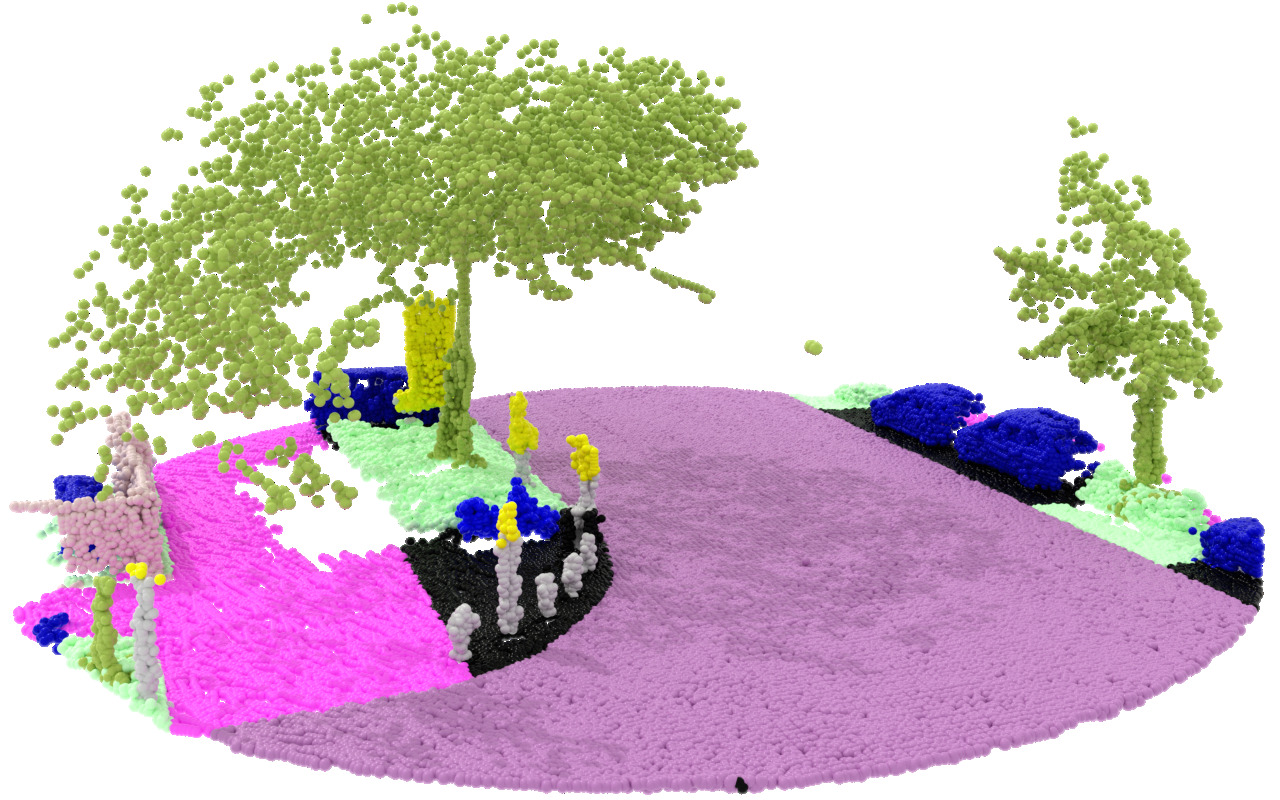}
    \end{subfigure}
         & 
    \begin{subfigure}[b]{0.28\textwidth}
      \includegraphics[width=\linewidth]{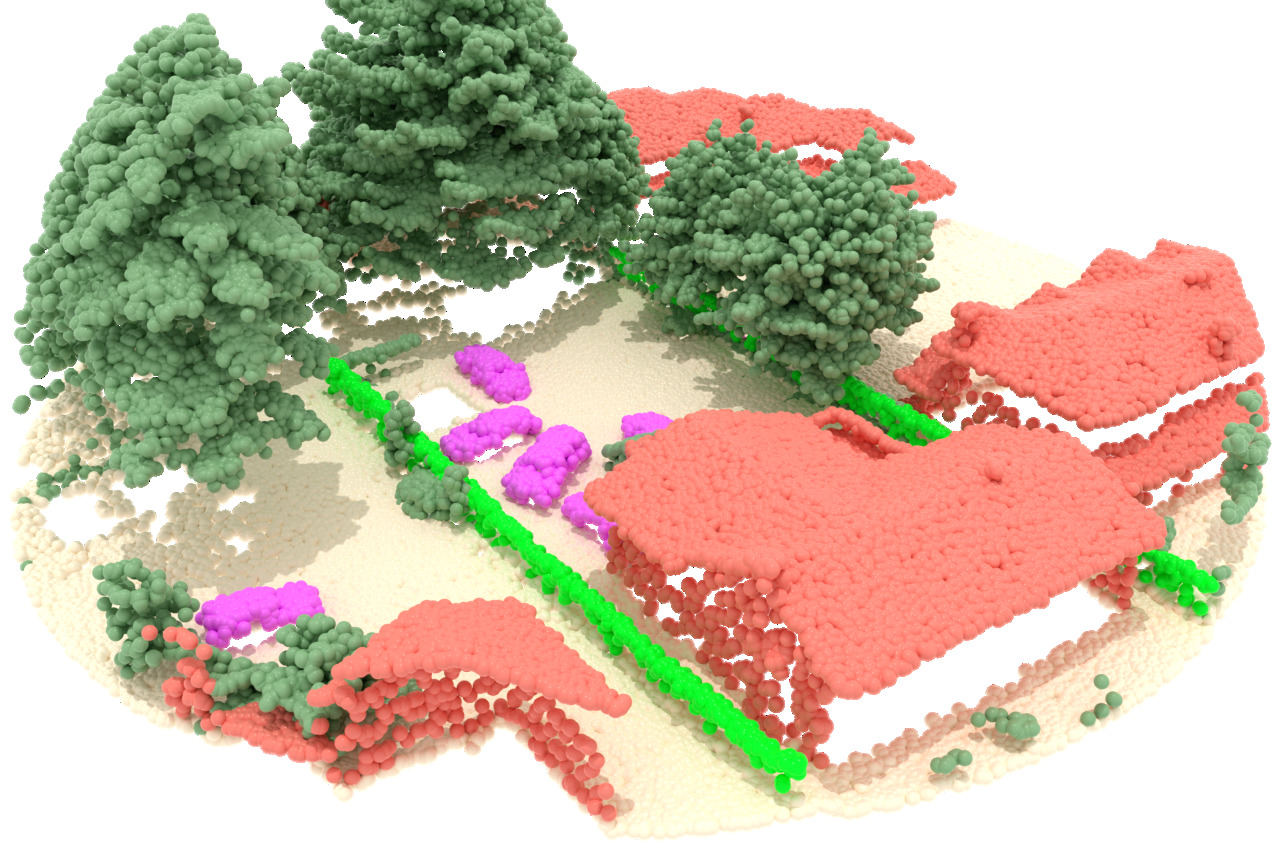}
    \end{subfigure}  \\
    \rotatebox{90}{ \qquad\quad Prediction}
    &
    \begin{subfigure}[b]{0.28\textwidth}
      \includegraphics[width=\linewidth]{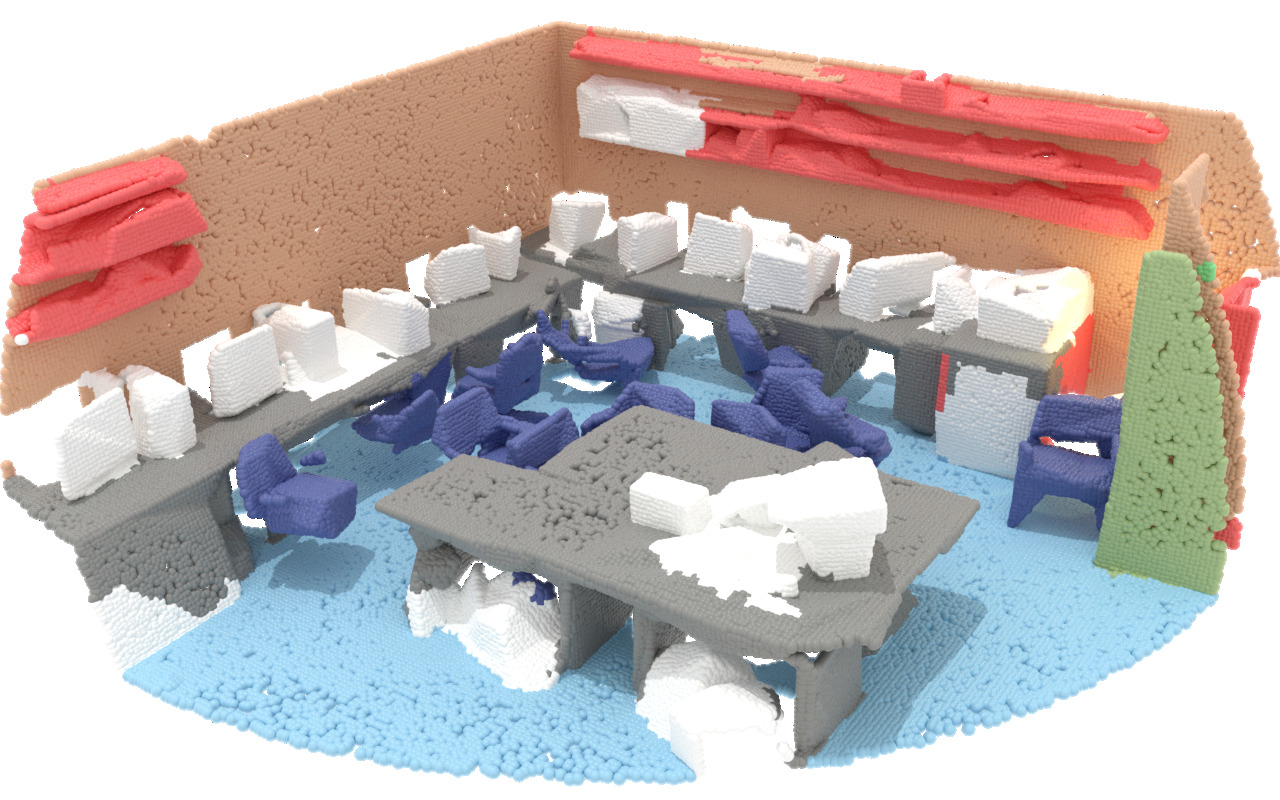}
      \caption{S3DIS}
      \label{fig:quali:stdis}
    \end{subfigure}
         & 
    \begin{subfigure}[b]{0.28\textwidth}
      \includegraphics[width=\linewidth]{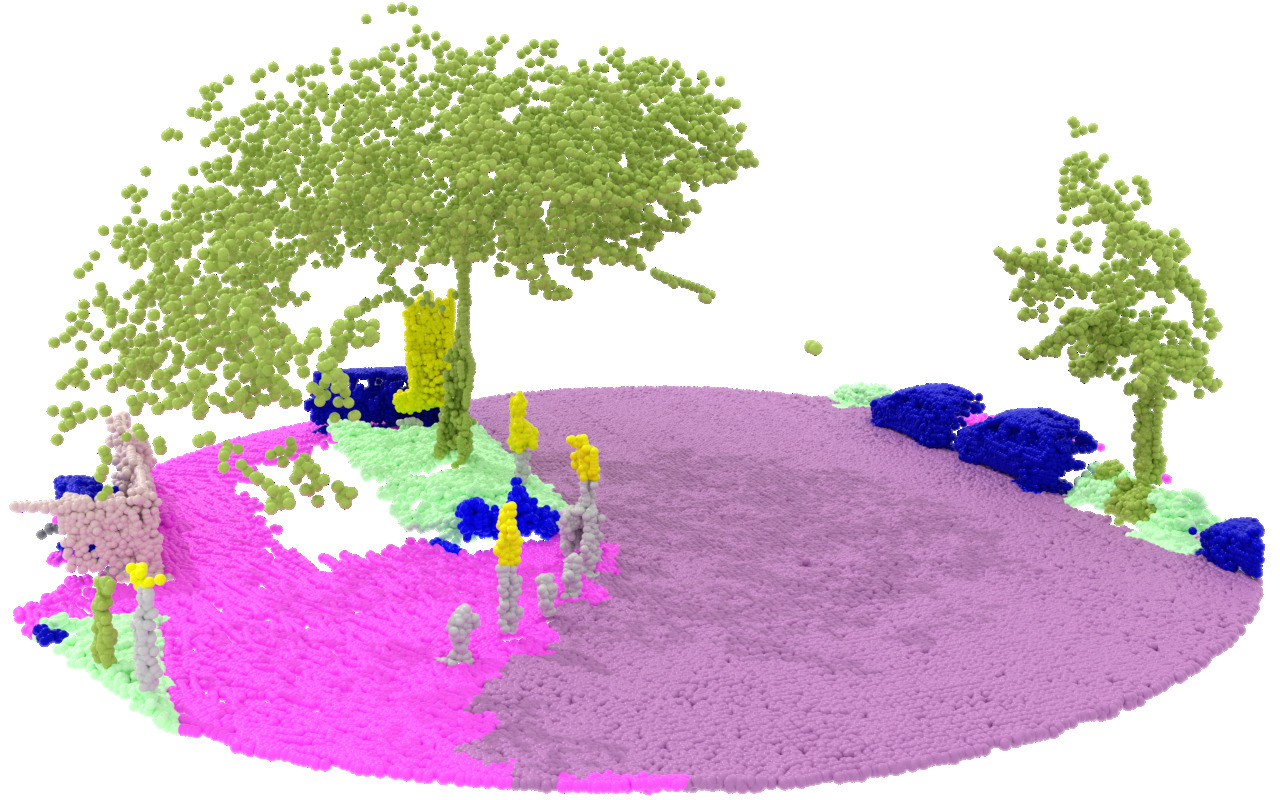}
      \caption{KITTI-360}
      \label{fig:quali:kitti}
    \end{subfigure}
         & 
    \begin{subfigure}[b]{0.28\textwidth}
      \includegraphics[width=\linewidth]{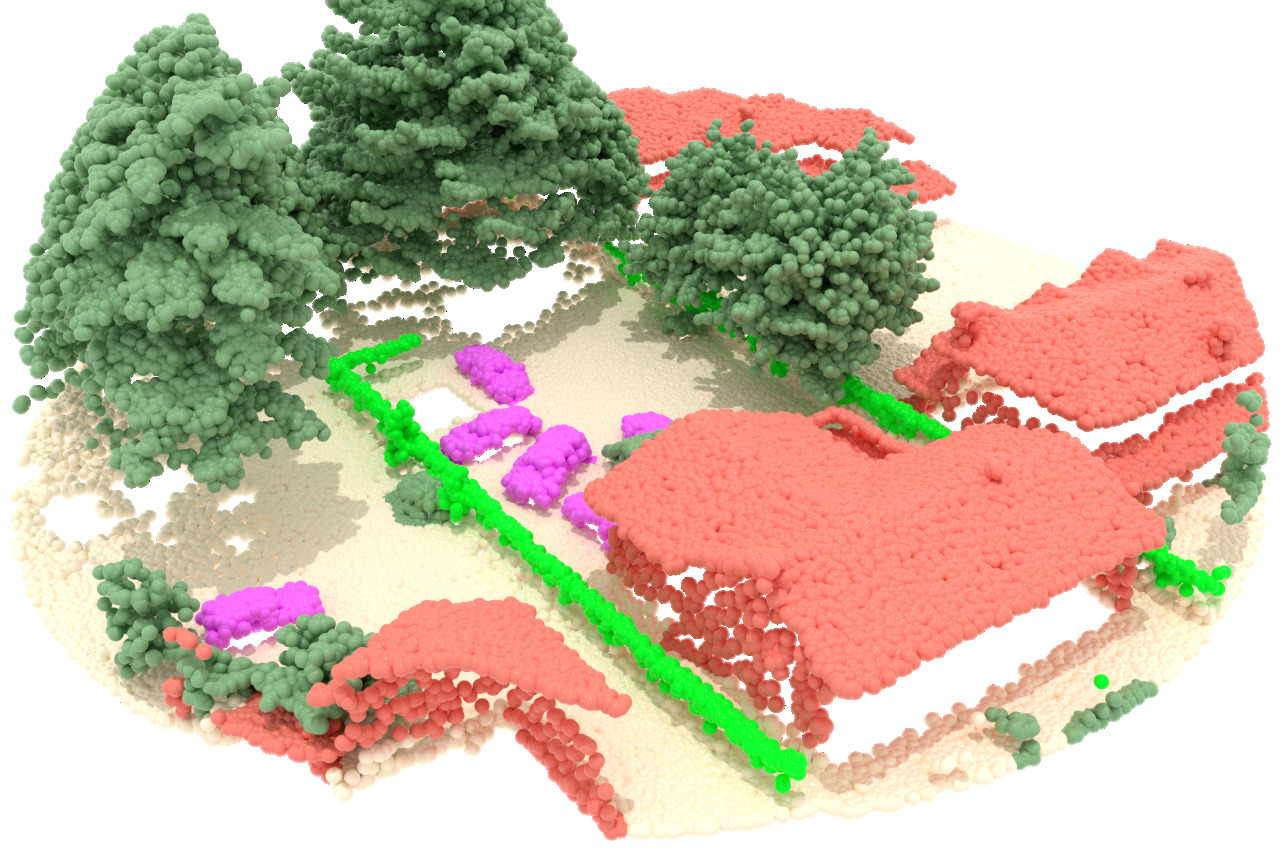}
      \caption{DALES}
      \label{fig:quali:dales}
    \end{subfigure}  \\
    \end{tabular}
    \caption{{\bf Qualitative Results.} We represent   input samples (with color or intensity) of our approach and its predictions for all three datasets. Additionally, we show the coarsest partition level and demonstrate how superpoints can accurately capture the contours of complex objects and classify them accordingly. Black points are unlabeled in the ground truth.}
    \label{fig:quali}
\end{figure*}
\noindent \textbf{KITTI-360 }\cite{liao2022kitti}. This outdoor dataset contains more than $100$ k laser scans acquired in various urban settings on a mobile platform. We use the \emph{accumulated point clouds} format, which consists of large scenes with around $3$ million points. There are $239$ training scenes and 
{$61$ for validation}.

\noindent\textbf{DALES} \cite{varney2020dales}. This $10$ km$^2$  aerial LiDAR dataset contains $500$ millions of points across $40$ urban and rural scenes, including $12$ for evaluation.

We subsample the datasets using a $3$cm grid for S3DIS, and $10$cm for KITTI-360 and DALES. All accuracy metrics are reported for the full, unsampled point clouds. 
We use a two-level partition ($I=2$) with $\mu^2=50$ for all datasets and select the partition parameters to obtain a $30$-fold reduction between $\cP_1$ and $\cP_0$ and a further $5$-fold reduction for $\cP_2$. See \tabref{tab:datas} for more details.

\paragraph{Models.} We use the same model configuration for all three datasets with minimal adaptations.  All transformer modules have a {shared} width $D_\text{val}$, a small key space of dimension $D_\text{key}=4$, $16$ heads, with $3$ blocks in the encoder and $1$ in the decoder. We set $D_\text{val}=64$ for S3DIS and DALES {($210$k parameters)}, and $D_\text{val}=128$ ($777$k parameters) for KITTI360. See the Appendix and our open repository for the detailed configuration of all modules.

We also propose SPT-nano, a lightweight version of our model that does not compute point-level features but operates directly on the first partition level $\cP_1$. The value of the maxpool over points in \eqref{eq:encoder} for $i=1$ is replaced by $f^1$, the
aggregated handcrafted point features at the level $1$ of the partition.
This model 
never considers the full point cloud $\cP_0$ but only operates on the partitions.
For this model, we set $D_\text{val}=16$ for S3DIS and DALES ($26$k parameters), and $D_\text{val}=32$ for KITTI360 ($70$k parameters).

\paragraph{Batch Construction.} Batches are sampled from large \emph{tiles}: entire building floors for S3DIS, and full scenes for KITTI-360 or DALES. Each batch is composed of $4$ randomly sampled portions of the partition with a radius of  
{$7$m for S3DIS and $50$m for KITTI and DALES, allowing us to model long-range interactions.}
During training, we apply a superpoint dropout rate of $0.2$ for each superpoint at all hierarchy levels, as well as random rotation, tilting, point jitter and handcrafted features dropout. When sampling points within each superpoint, we set $n_\text{min}=32$ and $n_\text{max}=128$.

\paragraph{Optimization.} 
We use the ADAMW optimizer \cite{loshchilov2017decoupled} with default parameters, a weight decay of $10^{-4}$, a learning rate of $10^{-2}$ for DALES and KITTI-360 on and $10^{-1}$ for S3DIS. The learning rate for the attention modules is $10$ times smaller than for other weights.
Learning rates are warmed up from $10^{-6}$ for $20$ epochs and progressively reduced to $10^{-6}$ with cosine annealing \cite{loshchilov10sgdr}.

\begin{table}
    \caption{\textbf{Performance Evaluation.} We report the Mean Intersection-over-Union of different methods on three different datasets. \SHORTHAND performs on par or better than recent methods with significantly fewer parameters. $\dagger$ superpoint-based. $\star$/$\ast$ model with $777$k/$70$k parameters.}
    \label{tab:benchmark}
    \centering\small{
     {\setlength{\tabcolsep}{4pt} 
    \begin{tabular}{@{}lllllll@{}}\toprule
        \multirow{2}{*}{Model} 
        &
        Size
        &
        \multicolumn{2}{c}{S3DIS}
        &
        \multicolumn{1}{c}{KITTI}
        &
        \multirow{2}{*}{DALES}
        \\%\cline{2-3}
        &  $\times10^6$
        & \footnotesize{6-Fold}  & \footnotesize{Area~5} & \multicolumn{1}{c}{360 val} & %
        \\ \midrule 
        PointNet++ \cite{qi2017pointnetpp}& 3.0& 56.7   & - & - & 68.3  \\
        $\dagger$ SPG \cite{landrieu2018large}& 0.28 & 62.1 & 58.0 & - & 60.6  \\
        ConvPoint \cite{boulch2020convpoint} &4.7& 68.2 & - & - & 67.4 \\
        $\dagger$ {SPG + SSP \cite{landrieu2019point}}& 0.29 & 68.4 & 61.7 & - & -  \\
        $\dagger$ SPNet \cite{hui2021superpoint} & 0.32 & 68.7 &-&-&- \\
        MinkowskiNet \cite{choy20194d,chaton2020torch}& 37.9& 69.1 & 65.4& 58.3 & -  \\
        RandLANet \cite{hu2020randla}&1.2 & 70.0&- & - & - \\
        KPConv \cite{thomas2019kpconv}& 14.1& 70.6 & 67.1 & - & \bf 81.1 \\
        Point Trans.\cite{zhao2021point}&7.8 & 73.5 & 70.4  & - & - \\
        RepSurf-U \cite{ran2022surface}& 0.97 &74.3 & 68.9 & - & - \\
        DeepViewAgg \cite{robert2022learning}& 41.2 & 74.7 & 67.2  & 62.1 & -\\ 
        Strat. Trans. \cite{lai2022stratified,Wang2022WindowNE} & 8.0 & 74.9 & \bf 72.0 & - &- \\ 
        PointNeXt-XL \cite{qian2022pointnext}& 41.6 & 74.9 & 71.1  & - & - 
        \\\midrule
        $\dagger$ \textbf{\SHORTHAND} (ours) &  0.21 & \bf 76.0 & 68.9 & \bf 63.5$^{\star}$ & 79.6 \\
        $\dagger$ \textbf{\SHORTHANDNANO} (ours) & \bf 0.026 & 70.8 & 64.9 & 57.2$^\ast$ & 75.2 \\
        \bottomrule
    \end{tabular}}}
\end{table}

\subsection{Quantitative Evaluation}
\label{sec:quant}
\begin{figure}
    \centering
    \tikzset{every mark/.append style={scale=1.75}}
    \hspace{-3mm}
    \begin{tikzpicture}
	\begin{semilogxaxis}[
        width=.94\linewidth,
        scale only axis,
		xlabel=Training time (GPU-h),
		ylabel=Area5 test mIoU,
		legend style={at={(0.27,0.0)}, font=\scriptsize,nodes={scale=0.83, transform shape},anchor= south west},
		xmajorgrids=true,
		xminorgrids=true,
		major grid style={line width=1.3pt,draw=gray!70},
		minor grid style={line width=1pt,draw=gray!15},
		every axis plot/.append style={ultra thick},
		xmin=0.2, xmax=230,
		ymin=30, ymax=71,
		legend columns=1,
		xtick={0.1,1,10,100,1000},
		xticklabels={,1,,100,},
            ytick={40,50,60,70},
		yticklabels={40,,60,70},
            extra x ticks={0.2},
            extra x tick labels={\;0.2},
		extra tick style={grid=minor},
		y label style={at={(axis description cs:0.09,0.67)},rotate=00,anchor=south},
        legend cell align={left},
        legend columns=2,
        ylabel style={xshift=-1.2cm,yshift=-0.5cm},
        xlabel style={xshift=+0.3cm,yshift=+0.55cm},
		]
    \if 1 0		
    \addplot[line width=1pt,draw=gray!50,forget plot,forget plot,forget plot] coordinates {(0.2,0)(0.2,100)};
    \addplot[line width=1pt,draw=gray!50,forget plot,forget plot] coordinates {(0.4,0)(0.4,100)};
    \addplot[line width=1pt,draw=gray!50,forget plot,forget plot] coordinates {(0.6,0)(0.6,100)};
    \addplot[line width=1pt,draw=gray!50,forget plot,forget plot] coordinates {(0.8,0)(0.8,100)};
    \addplot[line width=1pt,draw=gray!50,forget plot,forget plot] coordinates {(2,0)(2,100)};
    \addplot[line width=1pt,draw=gray!50,forget plot] coordinates {(4,0)(4,100)};
    \addplot[line width=1pt,draw=gray!50,forget plot] coordinates {(6,0)(6,100)};
    \addplot[line width=1pt,draw=gray!50,forget plot] coordinates {(8,0)(8,100)};
    \addplot[line width=1pt,draw=gray!50,forget plot] coordinates {(20,0)(20,100)};
    \addplot[line width=1pt,draw=gray!50,forget plot] coordinates {(40,0)(40,100)};
    \addplot[line width=1pt,draw=gray!50,forget plot] coordinates {(60,0)(60,100)};
    \addplot[line width=1pt,draw=gray!50,forget plot] coordinates {(80,0)(80,100)};
    \fi

    \addplot[color=SPTCOLOR, smooth, tension=0.05] table [ x expr = \thisrow{time} + 0.2, y=iou] {figures/data/speed/spt.csv};

    \addplot[color=SPTCOLOR!75!black, smooth, tension=0.05] table [x expr =  \thisrow{time} + 0.2, y=iou] {figures/data/speed/nano.csv};

    \addplot[color=SPGCOLOR, smooth] table [x expr = \thisrow{time} + 1.5, y=iou] {figures/data/speed/spg.csv};
    \addplot[color=POINTNETPPCOLOR, smooth] table [x expr = \thisrow{time} + 0.13, y=iou] {figures/data/speed/ptn.csv};
    \addplot[color=KPCONVCOLOR, smooth] table [x expr = \thisrow{time} + 0.38, y=iou] {figures/data/speed/kpconv.csv};
    \addplot[color=MINKOCOLOR, smooth] table [x expr = \thisrow{time} + 0.35, y=iou] {figures/data/speed/minko.csv};
    \addplot[color=DVACOLOR, smooth, tension=0.05] table [x expr = \thisrow{time} + 0.66, y=iou] {figures/data/speed/dva.csv};
    \addplot[color=POINTTRANSCOLOR, smooth,tension=0.05] table [x expr = \thisrow{time} + 0.13, y=iou] {figures/data/speed/pointtrans.csv};
    \addplot[color=STRATTRANSCOLOR, smooth,tension=0.05] table [x expr = \thisrow{time} + 0.13, y=iou] {figures/data/speed/strat.csv};

    \addplot[color=SPTCOLOR, mark=oplus] table [x expr = \thisrow{time} + 0.2, y=iou] {
        time	iou
        3.02	68.7
    };
    \addplot[color=SPTCOLOR!75!black, mark=|] table [x expr = \thisrow{time} + 0.2, y=iou] {
        time	iou
        1.50    64.0
    };
    \addplot[color=SPGCOLOR, mark=square] table [x expr = \thisrow{time} + 1.5, y=iou] {
        time	iou
        1.292	62.86
    };
    \addplot[color=POINTNETPPCOLOR, mark=triangle] table [x expr = \thisrow{time} + 0.13, y=iou] {
        time	iou
        6.250	52.37
    };
    \addplot[color=KPCONVCOLOR, mark=diamond] table [x expr = \thisrow{time} + 0.38, y=iou] {
        time	iou
        14.17	63.2
    };
    \addplot[color=MINKOCOLOR, mark=o] table [x expr = \thisrow{time} + 0.35, y=iou] {
        time	iou
        28.33	60.7
    };
    \addplot[color=DVACOLOR, mark=pentagon] table [x expr = \thisrow{time} + 0.66, y=iou] {
        time	iou
        35.28   66.50
    };
    \addplot[color=POINTTRANSCOLOR, mark=star] table [x expr = \thisrow{time} + 0.13, y=iou] {
        time	iou
        63.12	69.8
    };
    \addplot[color=STRATTRANSCOLOR, mark=|] table [x expr = \thisrow{time} + 0.13, y=iou] {
        time	iou
        216.24	70.53
    };
 
    {\legend{, , , , , , , , , {\footnotesize \bf  \SHORTHAND}, \footnotesize \bf SPT-nano ($\times 0.5$), \footnotesize SPG ($\times 0.9$), \footnotesize  PointNet++ ($\times 2$), \footnotesize  KPConv ($\times 5$), \footnotesize MinkowskiNet ($\times 9$),  \footnotesize DeepViewAgg ($\times 11$),  \footnotesize Point Trans ($\times 20$), \footnotesize Strat. Trans. $\ast$  ($\times 67$)}
    }

	\end{semilogxaxis}%
\end{tikzpicture}
    \caption{{\bf Training Speed.} We report the evolution of the test mIoU for S3DIS Area~5 for different methods 
    {\emph{until the best epoch is reached}}. The curves are shifted right according to the preprocessing time. We report in parenthesis the time ratio compared to SPT. }
    \label{fig:training_speed}
\end{figure}
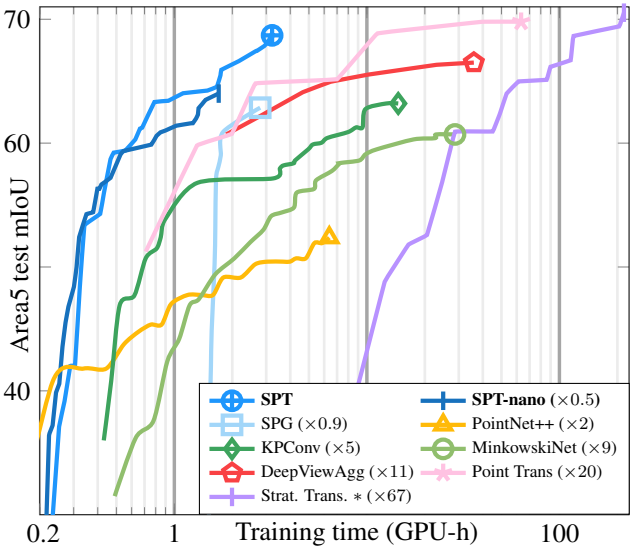

\paragraph{Performance Evaluation.} As seen in \tabref{tab:benchmark}, \SHORTHAND performs at the state-of-the-art on two of three datasets despite being a significantly smaller model. On S3DIS, \SHORTHAND beats PointNeXt-XL with $196 \times$ fewer parameters. 
On KITTI-360, \SHORTHAND outperforms MinkowskiNet despite a size ratio of $49$, and surpasses the performance of the even larger multimodal point-image model DeepViewAgg.
On DALES, \SHORTHAND outperforms ConvPoint by more than $12$ points with over $21$ times fewer parameters. 
Although \SHORTHAND is $1.5$ points behind KPConv on this dataset, it achieves these results with $67$ times fewer parameters.
\SHORTHAND achieves significant performance improvements over all superpoint-based methods on all datasets, ranging from $7$ to $14$ points.
SPT overtakes the SSP and SPNet superpoint methods that \textit{learn} the partition in a two-stage training setup, leading to pre-processing times of several hours.

Interestingly, the lightweight \SHORTHANDNANO model matches KPConv and MinkowskiNet with only  $26$k parameters.%

See Figure \ref{fig:quali} for qualitative illustrations.

\paragraph{Preprocessing Speed.} 
As reported in \tabref{tab:efficiency}, our implementation of the preprocessing step is highly efficient. We can compute partitions, superpoint-graphs, and handcrafted features, and perform I/O operations quickly: 
{$12.4$min for S3DIS, $117$ for KITTI-360, and $148$ for DALES using a server with a 48-core CPU.}
An 8-core workstation can preprocess S3DIS in $26.6$min.
Our preprocessing time is as fast or faster than point-level methods and $7\times$ faster than SuperPoint Graph's, thus alleviating one of the main drawbacks of superpoint-based methods.

\paragraph{Training Speed.}
We trained several state-of-the-art methods from scratch and report in \figref{fig:training_speed} the evolution of test performance as a function of training time.
We used the official training logs for the multi-GPU Point Transformer and Stratified Transformer.
\SHORTHAND can train much faster than all methods not based on superpoints while attaining similar performance.
Although Superpoint Graph trains even faster, its performance
saturates earlier,
$6.0$ mIoU points below \SHORTHAND.
We also report the inference time of our method in \tabref{tab:efficiency}, which is significantly lower than competing approaches, with a speed-up factor ranging from $8$ to $80$. 
All speed measurements were conducted on a single-GPU server (48 cores, 512Go RAM, A40 GPU). Nevertheless, our model can be trained on a 
{standard workstation (8 cores, 64Go, 2080Ti)}
with smaller batches, taking only $1.5$ times longer and with comparable results.

SPT performs on par or better than complex models with
up to two orders of magnitude more parameters and significantly longer training times. 
Such efficiency and compactness have many benefits for real-world scenarios where hardware, time, or energy may be limited.

\begin{table}
    \caption{\textbf{Efficiency Analysis.} We report the preprocessing time for the entire S3DIS dataset and the training and inference time for Area~5. \SHORTHAND and \SHORTHANDNANO shows significant speedups in pre-processing, training, and inference times. %
    }
    \label{tab:efficiency}
    \centering
    \small{
    \begin{tabularx}{\linewidth}{@{}lccc@{}}
        \toprule
        &  Preprocessing & Training &  Inference\\
        &  in min &  in GPU-h &   in s \\\midrule

        PointNet++ \cite{qi2017pointnetpp} &  8.0 & 6.3 & 42   \\
        KPConv \cite{thomas2019kpconv} & 23.1 & 14.1 & 162 \\
        MinkowskiNet \cite{choy20194d} & 20.7  & 28.8 & 83 \\
        Stratified Trans. \cite{lai2022stratified} & 8.0 & 216.4 &  30  \\
        Superpoint Graph \cite{landrieu2018large} &  89.9 & 1.3 & 16\\%106.26
        \midrule
        \textbf{\SHORTHAND} (ours) & 12.4 &  3.0  &  2\\
        \textbf{\SHORTHANDNANO} (ours)  &  12.4 & 1.9 & 1 \\

        \bottomrule
    \end{tabularx}}
\end{table}

\subsection{Ablation Study}
\label{sec:abla}
We evaluate the impact of several design choices in \tabref{tab:ablation} and reports here our observations.

\paragraph{a) Handcrafted features.} 
Without handcrafted point features, our model perform worse on all datasets.
This observation is in line with other works which also remarked the positive impact of well-designed handcrafted features on the performance of smaller models \cite{hsu2020incorporating,ran2022surface}.

\paragraph{b) Influence of Edges.}

Removing the adjacency encoding between superpoints leads to a significant drop of $6.3$ points on S3DIS; characterizing the relative position and relationship between superpoints appears crucial to exploiting their context. 
We also find that pruning the $50$\% longest edges of each superpoint results in a systematic performance drop, highlighting the importance of modeling long relationships.

\begin{table}
    \caption{\textbf{Ablation Study.} Impact of some of our design choices on the mIoU 
    {for all tested datasets.}}
    \label{tab:ablation}
    \centering
    \small{
    \begin{tabular}{@{}lccc@{}}
        \toprule
        Experiment & S3DIS & KITTI & DALES \\
         & 6-Fold & 360 Val & \\
        \midrule

        Best Model                 & 76.0 & 63.5 & 79.6 \\
        \midrule
        a) No handcrafted features & -0.7 & -4.1 & -1.4 \\
        b) No adjacency encoding   & -6.3 & -5.4 & -3.0 \\
        b) Fewer edges             & -3.5 & -1.1 & -0.3 \\
        c) No point sampling       & -1.3 & -0.9 & -0.5 \\
        c) No superpoint sampling  & -2.7 & -2.5 & -0.7 \\
        c) Only 1 partition level  & -8.4 & -5.1 & -0.9 \\
     
        \bottomrule
    \end{tabular}}
\end{table}

\paragraph{c) Partition-Based Improvements.} We assess the impact of several improvements made possible by using hierarchical superpoints. 
First, we find that using all available points when embedding the superpoints of $\cP_1$ instead of random sampling resulted in a small performance drop.
Second, setting the superpoint dropout rate to $0$ worsens the performance by over  
$2.5$ points on S3DIS and KITTI-360.

While we did not observe better results with three or more partition levels, only using one level leads to a significant loss of performance for all datasets.

\begin{figure}
\centering
\begin{tikzpicture}
\begin{semilogxaxis}[
    xlabel=Number of superpoints / nonempty voxels,
    ylabel=Area5 test mIoU,
    legend pos = {south east},
    xmajorgrids=true,
    xminorgrids=true,
    major grid style={line width=1.3pt,draw=gray!70},
    minor grid style={line width=1pt,draw=gray!15},
    every axis plot/.append style={ultra thick},
    xmin=5000, xmax=12000000,
    ymin=40, ymax=99.5,
    extra tick style={grid=minor},
    y label style={at={(axis description cs:0.1,.5)},rotate=00,anchor=south},
    legend cell align={left},
    mark options={solid},
    legend style={at={(1.00,0.0)}}
]

\addplot[color=VOXELPURITYCOLOR, smooth,tension=0.5, mark=*, densely dotted] table [x=sps, y=iou] {figures/data/purity/voxel.csv};

\addlegendimage{legend image with text=\textcolor{VOXELPURITYCOLOR!80!black}{\small $\mathbf{x}$ cm}}

\addplot[color=PARTITIONPURITYCOLOR, smooth,tension=0.05, mark=x, mark size=3pt] table [x=sps, y=iou] {figures/data/purity/partition_oracle.csv};

\addplot[color=SPTPURITYCOLOR, smooth,tension=0.05, mark=square*, only marks] table [x=sps, y=iou] {figures/data/purity/spt.csv};

\addlegendimage{legend image with text=\textcolor{SPTPURITYCOLOR}{$\mathbf{\times n}$}}

\addplot[densely dotted, SPTPURITYCOLOR] coordinates {(10,10) (20,20)};

\node[draw=none] (x1a) at (axis cs: 278167, 68.9) {};
\node[draw=none, text=SPTPURITYCOLOR] (x1b) at ($(x1a.center)+(0.4, -0)$) {\footnotesize $\mathbf{\times1}$};
\draw [-, SPTPURITYCOLOR, densely dotted, thick] (x1a) -- (axis cs: 278167, 96.40) ;

\node[draw=none] (x1_5a) at (axis cs: 169772, 66.7) {};
\node[draw=none, text=SPTPURITYCOLOR] (x1_5b) at ($(x1_5a.center)+(0, -30)$) {\footnotesize $\mathbf{\times1.5}$};
\draw [-, SPTPURITYCOLOR, densely dotted, thick] (x1_5a) --  (axis cs: 169772, 94.21)  ;

\node[draw=none] (x3a) at (axis cs: 93223, 64.2) {};
\node[draw=none, text=SPTPURITYCOLOR] (x3b) at ($(x3a.center)+(0, -30)$) {\footnotesize $\mathbf{\times3}$};
\draw [-, SPTPURITYCOLOR, densely dotted, thick] (x3a) -- (axis cs: 93223, 91.53) ;

\node[draw=none] (x10a) at (axis cs: 26714, 56.8) {};
\node[draw=none, text=SPTPURITYCOLOR] (x10b) at ($(x10a.center)+(0, -30)$) {\footnotesize $\mathbf{\times10}$};
\draw [-, SPTPURITYCOLOR, densely dotted, thick] (x10a) -- (axis cs: 26714, 83.40);

\node[draw=none] (thirtya) at (axis cs: 108000,81.12) {};
\node[draw=none, text=VOXELPURITYCOLOR!80!black] (thirtyb) at ($(thirtya.center)+(0,-70)$) {\footnotesize \bf $\mathbf{30}$ cm};
\draw [-, VOXELPURITYCOLOR!80!black, very thick] (thirtyb) -- ($(thirtya.center)+(0,-15)$);

\node[draw=none] (twentya) at (axis cs: 220727,87.55) {};
\node[draw=none, text=VOXELPURITYCOLOR!80!black] (twentyb) at ($(twentya.center)+(0,-60)$) {\footnotesize \bf $\mathbf{20}$ cm};
\draw [-, VOXELPURITYCOLOR!80!black, very thick] (twentyb) -- ($(twentya.center)+(0,-15)$);

\node[draw=none] (tena) at (axis cs: 981205,94.98) {};
\node[draw=none, text=VOXELPURITYCOLOR!80!black] (tenb) at ($(tena.center)+(0,-60)$) {\footnotesize \bf $\mathbf{10}$ cm};
\draw [-, VOXELPURITYCOLOR!80!black, very thick] (tenb) -- ($(tena.center)+(0,-15)$);

\node[draw=none] (fivea) at (axis cs: 3621421,97.65) {};
\node[draw=none, text=VOXELPURITYCOLOR!80!black] (fiveb) at ($(fivea.center)+(0,-60)$) {\footnotesize \bf $\mathbf{5}$ cm};
\draw [-, VOXELPURITYCOLOR!80!black, very thick] (fiveb) -- ($(fivea.center)+(0,-15)$);

\node[draw=none] (threea) at (axis cs: 9308118,98.59) {};
\node[draw=none, text=VOXELPURITYCOLOR!80!black] (threeb) at ($(threea.center)+(0,-60)$) {\footnotesize \bf $\mathbf{3}$ cm};
\draw [-, VOXELPURITYCOLOR!80!black, very thick] (threeb) -- ($(threea.center)+(0,-15)$);

\legend{
    {\footnotesize Voxel grid oracle},
    {\footnotesize  Grid size},
    {\footnotesize Partition oracle},
    {\footnotesize \SHORTHAND performance},
    {\footnotesize  Coarseness ratio},
    {\footnotesize  Performance gap},
}
\end{semilogxaxis}%
\end{tikzpicture}

\caption{{\bf Partition Purity.} 
    {
    We plot the highest achievable ``oracle'' prediction for our partitions and a regular voxel grid.
    We also show the performance of \SHORTHAND for $4$ partitions with a coarseness ratio from $\times1$ to $\times 10$. \vspace{-3mm}
    } 
}

\label{fig:purity}
\end{figure}
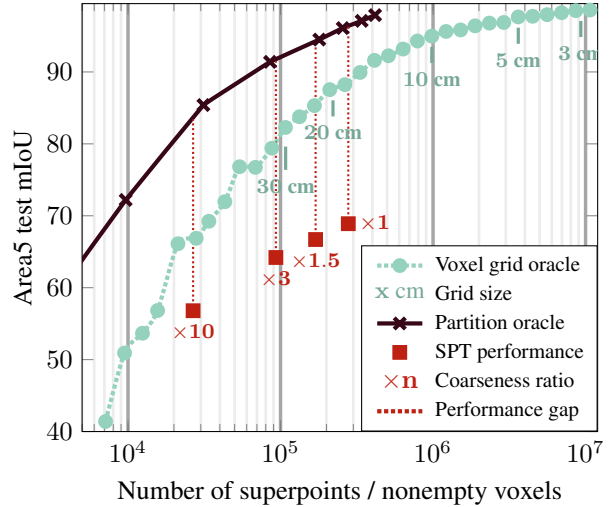

\paragraph{d) Influence of Partition {Purity}.}
In \figref{fig:purity}, we plot the performance of the ``oracle'' model which associates to each superpoint of $\cP_1$ with its most frequent true label. This model acts as an upper bound on the achievable performance with a given partition. 
Our proposed partition has significantly higher semantic purity than a regular voxel grid with as many nonempty voxels as superpoints. This implies that our superpoints adhere better to semantic boundaries between objects.

We also report the performance of our model for different partitions of varying coarseness, measured as the number of superpoints in $\cP_1$.
Using, respectively, $\times$1.5 ($\times$3) fewer superpoints leads to a performance drop of $2.2$ ($4.7$) mIoU points, but reduce the training time to $2.4$ ($1.6$) hours.
The performance of SPT is more than $20$ points below the oracle, suggesting that the partition does not strongly limit its performance.

\paragraph{Limitations.} See the Appendix.

\section{Conclusion}
We have introduced the Superpoint Transformer approach for semantic segmentation of large point clouds, combining superpoints and transformers to achieve state-of-the-art results with significantly reduced training time, inference time, and model size.
This approach particularly benefits large-scale applications and computing with limited resources. 
More broadly, we argue that small, tailored models can offer a more flexible and sustainable alternative to large, generic models for 3D learning.
With training times of a few hours on a single GPU, our approach allows practitioners to easily customize the models to their specific needs, enhancing the overall usability and accessibility of 3D learning.

\paragraph*{Acknowledgements.} 
This work was funded by ENGIE Lab CRIGEN. %
This work was supported by ANR project READY3D ANR-19-CE23-0007,
and was granted access to the HPC resources of IDRIS under the allocation AD011013388R1 made by GENCI.
We thank Bruno Vallet, Romain Loiseau and Ewelina Rupnik for inspiring discussions and valuable feedback.

\FloatBarrier
{\small
\balance
\bibliographystyle{templates/ICCV/ieee_fullname}
\bibliography{mybib}

\begin{thebibliography}{10}\itemsep=-1pt

\bibitem{achanta2012slic}
Radhakrishna Achanta, Appu Shaji, Kevin Smith, Aurelien Lucchi, Pascal Fua, and
  Sabine S{\"u}sstrunk.
\newblock {SLIC} superpixels compared to state-of-the-art superpixel methods.
\newblock {\em TPAMI}, 2012.

\bibitem{arbelaez2006boundary}
Pablo Arbelaez.
\newblock Boundary extraction in natural images using ultrametric contour maps.
\newblock {\em CVPR Workshop}, 2006.

\bibitem{armeni20163d}
Iro Armeni, Ozan Sener, Amir~R Zamir, Helen Jiang, Ioannis Brilakis, Martin
  Fischer, and Silvio Savarese.
\newblock {3D} semantic parsing of large-scale indoor spaces.
\newblock {\em CVPR}, 2016.

\bibitem{boulch2020convpoint}
Alexandre Boulch.
\newblock {ConvPoint}: Continuous convolutions for point cloud processing.
\newblock {\em Computers \& Graphics}, 2020.

\bibitem{cai2021graphnorm}
Tianle Cai, Shengjie Luo, Keyulu Xu, Di He, Tie-yan Liu, and Liwei Wang.
\newblock Graph{N}orm: {A} principled approach to accelerating graph neural
  network training.
\newblock {\em ICML}, 2021.

\bibitem{chaton2020torch}
Thomas Chaton, Nicolas Chaulet, Sofiane Horache, and Loic Landrieu.
\newblock Torch-{P}oints{3D}: {A} modular multi-task framework for reproducible
  deep learning on 3{D} point clouds.
\newblock {\em 3DV}, 2020.

\bibitem{chen2021hierarchical}
Shaoyu Chen, Jiemin Fang, Qian Zhang, Wenyu Liu, and Xinggang Wang.
\newblock Hierarchical aggregation for 3{D} instance segmentation.
\newblock {\em CVPR}, 2021.

\bibitem{choy20194d}
Christopher Choy, JunYoung Gwak, and Silvio Savarese.
\newblock 4{D} spatio-temporal {ConvNets}: {M}inkowski convolutional neural
  networks.
\newblock {\em CVPR}, 2019.

\bibitem{demantke2011dimensionality}
J{\'e}r{\^o}me Demantk{\'e}, Cl{\'e}ment Mallet, Nicolas David, and Bruno
  Vallet.
\newblock Dimensionality based scale selection in {3D LiDAR} point clouds.
\newblock In {\em Laserscanning}, 2011.

\bibitem{dosovitskiy2020image}
Alexey Dosovitskiy, Lucas Beyer, Alexander Kolesnikov, Dirk Weissenborn,
  Xiaohua Zhai, Thomas Unterthiner, Mostafa Dehghani, Matthias Minderer, Georg
  Heigold, Sylvain Gelly, et~al.
\newblock An image is worth 16x16 words: {T}ransformers for image recognition
  at scale.
\newblock {\em ICLR}, 2020.

\bibitem{engelmann20203d}
Francis Engelmann, Martin Bokeloh, Alireza Fathi, Bastian Leibe, and Matthias
  Nie{\ss}ner.
\newblock {3D-MPA: M}ulti-proposal aggregation for 3{D} semantic instance
  segmentation.
\newblock {\em CVPR}, 2020.

\bibitem{fan2013point}
Yuxue Fan, Yan Huang, and Jingliang Peng.
\newblock Point cloud compression based on hierarchical point clustering.
\newblock In {\em APSIPA ASC}, 2013.

\bibitem{fischler1981random}
Martin~A Fischler and Robert~C Bolles.
\newblock Random sample consensus: a paradigm for model fitting with
  applications to image analysis and automated cartography.
\newblock {\em Communications of the ACM}, 1981.

\bibitem{gao2019graph}
Hongyang Gao and Shuiwang Ji.
\newblock Graph {U-Nets}.
\newblock {\em ICML}, 2019.

\bibitem{owidartificialintelligence}
Charlie Giattino, Edouard Mathieu, Julia Broden, and Max Roser.
\newblock Artificial intelligence.
\newblock {\em Our World in Data}, 2022.
\newblock https://ourworldindata.org/artificial-intelligence.

\bibitem{SubmanifoldSparseConvNetCVPR}
Benjamin Graham, Martin Engelcke, and Laurens van~der Maaten.
\newblock 3{D} semantic segmentation with submanifold sparse convolutional
  networks.
\newblock {\em CVPR}, 2018.

\bibitem{guinard2017weakly}
St{\'e}phane Guinard and Loic Landrieu.
\newblock Weakly supervised segmentation-aided classification of urban scenes
  from {3D LiDAR} point clouds.
\newblock {\em ISPRS Workshop}, 2017.

\bibitem{guo2021pct}
Meng-Hao Guo, Jun-Xiong Cai, Zheng-Ning Liu, Tai-Jiang Mu, Ralph~R Martin, and
  Shi-Min Hu.
\newblock {PCT:} {P}oint cloud transformer.
\newblock {\em CVM}, 2021.

\bibitem{han2020occuseg}
Lei Han, Tian Zheng, Lan Xu, and Lu Fang.
\newblock Occuseg: Occupancy-aware {3D} instance segmentation.
\newblock {\em CVPR}, 2020.

\bibitem{he2016deep}
Kaiming He, Xiangyu Zhang, Shaoqing Ren, and Jian Sun.
\newblock Deep residual learning for image recognition.
\newblock {\em CVPR}, 2016.

\bibitem{hsu2020incorporating}
Pai-Hui Hsu and Zong-Yi Zhuang.
\newblock Incorporating handcrafted features into deep learning for point cloud
  classification.
\newblock {\em Remote Sensing}, 2020.

\bibitem{hu2020randla}
Qingyong Hu, Bo Yang, Linhai Xie, Stefano Rosa, Yulan Guo, Zhihua Wang, Niki
  Trigoni, and Andrew Markham.
\newblock Rand{LA-N}et: {E}fficient semantic segmentation of large-scale point
  clouds.
\newblock {\em CVPR}, 2020.

\bibitem{hui2021superpoint}
Le Hui, Jia Yuan, Mingmei Cheng, Jin Xie, Xiaoya Zhang, and Jian Yang.
\newblock Superpoint network for point cloud oversegmentation.
\newblock {\em ICCV}, 2021.

\bibitem{kang2023region}
Xin Kang, Chaoqun Wang, and Xuejin Chen.
\newblock Region-enhanced feature learning for scene semantic segmentation.
\newblock {\em arXiv preprint arXiv:2304.07486}, 2023.

\bibitem{lai2022stratified}
Xin Lai, Jianhui Liu, Li Jiang, Liwei Wang, Hengshuang Zhao, Shu Liu, Xiaojuan
  Qi, and Jiaya Jia.
\newblock Stratified transformer for 3{D} point cloud segmentation.
\newblock {\em CVPR}, 2022.

\bibitem{landrieu2019point}
Loic Landrieu and Mohamed Boussaha.
\newblock Point cloud oversegmentation with graph-structured deep metric
  learning.
\newblock {\em CVPR}, 2019.

\bibitem{landrieu2016cut}
Loic Landrieu and Guillaume Obozinski.
\newblock Cut pursuit: fast algorithms to learn piecewise constant functions.
\newblock {\em AISTATS}, 2016.

\bibitem{landrieu2017cut}
Loic Landrieu and Guillaume Obozinski.
\newblock Cut pursuit: Fast algorithms to learn piecewise constant functions on
  general weighted graphs.
\newblock In {\em SIAM Journal on Imaging Sciences}, 2017.

\bibitem{landrieu2018large}
Loic Landrieu and Martin Simonovsky.
\newblock Large-scale point cloud semantic segmentation with superpoint graphs.
\newblock {\em CVPR}, 2018.

\bibitem{li2018pointcnn}
Yangyan Li, Rui Bu, Mingchao Sun, Wei Wu, Xinhan Di, and Baoquan Chen.
\newblock Pointcnn: Convolution on $\chi$-transformed points.
\newblock {\em NeurIPS}, 2018.

\bibitem{liang2021instance}
Zhihao Liang, Zhihao Li, Songcen Xu, Mingkui Tan, and Kui Jia.
\newblock Instance segmentation in 3{D} scenes using semantic superpoint tree
  networks.
\newblock {\em CVPR}, 2021.

\bibitem{liao2022kitti}
Yiyi Liao, Jun Xie, and Andreas Geiger.
\newblock {KITTI-360: A} novel dataset and benchmarks for urban scene
  understanding in {2D and 3D}.
\newblock {\em TPAMI}, 2022.

\bibitem{lin2018toward}
Yangbin Lin, Cheng Wang, Dawei Zhai, Wei Li, and Jonathan Li.
\newblock Toward better boundary preserved supervoxel segmentation for 3{D}
  point clouds.
\newblock {\em ISPRS journal of photogrammetry and remote sensing}, 2018.

\bibitem{liu2021swin}
Ze Liu, Yutong Lin, Yue Cao, Han Hu, Yixuan Wei, Zheng Zhang, Stephen Lin, and
  Baining Guo.
\newblock Swin transformer: Hierarchical vision transformer using shifted
  windows.
\newblock {\em CVPR}, 2021.

\bibitem{liu2019point}
Zhijian Liu, Haotian Tang, Yujun Lin, and Song Han.
\newblock Point-voxel {CNN} for efficient 3{D} deep learning.
\newblock {\em NeurIPS}, 2019.

\bibitem{loiseau2022online}
Romain Loiseau, Mathieu Aubry, and Lo{\"\i}c Landrieu.
\newblock Online segmentation of {LiDAR} sequences: Dataset and algorithm.
\newblock {\em ECCV}, 2022.

\bibitem{loshchilov10sgdr}
Ilya Loshchilov and Frank Hutter.
\newblock {SGDR}: {S}tochastic gradient descent with warm restarts.
\newblock {\em ICLR}, 2017.

\bibitem{loshchilov2017decoupled}
Ilya Loshchilov and Frank Hutter.
\newblock Decoupled weight decay regularization.
\newblock {\em ICLR}, 2019.

\bibitem{narasimhamurthy2023hierarchical}
J Narasimhamurthy, Karthikeyan Vaiapury, Ramanathan Muthuganapathy, and
  Balamuralidhar Purushothaman.
\newblock Hierarchical-based semantic segmentation of 3{D} point cloud using
  deep learning.
\newblock {\em Smart Computer Vision}, 2023.

\bibitem{papon2013voxel}
Jeremie Papon, Alexey Abramov, Markus Schoeler, and Florentin Worgotter.
\newblock Voxel cloud connectivity segmentation-supervoxels for point clouds.
\newblock {\em CVPR}, 2013.

\bibitem{park2022fast}
Chunghyun Park, Yoonwoo Jeong, Minsu Cho, and Jaesik Park.
\newblock Fast point transformer.
\newblock {\em CVPR}, 2022.

\bibitem{qi2017pointnet}
Charles~R Qi, Hao Su, Kaichun Mo, and Leonidas~J Guibas.
\newblock Point{N}et: {D}eep learning on point sets for 3{D} classification and
  segmentation.
\newblock {\em CVPR}, 2017.

\bibitem{qi2017pointnetpp}
Charles~R Qi, Li Yi, Hao Su, and Leonidas~J Guibas.
\newblock Point{N}et++: Deep hierarchical feature learning on point sets in a
  metric space.
\newblock {\em NeurIPS}, 2017.

\bibitem{qian2022pointnext}
Guocheng Qian, Yuchen Li, Houwen Peng, Jinjie Mai, Hasan Hammoud, Mohamed
  Elhoseiny, and Bernard Ghanem.
\newblock {PointNeXt}: {R}evisiting {P}oin{N}et++ with improved training and
  scaling strategies.
\newblock {\em NeurIPS}, 2022.

\bibitem{quana2016international}
Xingwen Quana, Binbin Hea, Marta Yebrab, Changmin Yina, Zhanmang Liaoa, Xueting
  Zhanga, and Xing Lia.
\newblock Hierarchical semantic segmentation of urban scene point clouds via
  group proposal and graph attention network.
\newblock {\em International Journal of Applied Earth Observations and
  Geoinformation}, 2016.

\bibitem{raguet2019parallel}
Hugo Raguet and Loic Landrieu.
\newblock Parallel cut pursuit for minimization of the graph total variation.
\newblock {\em ICML Workshop on Graph Reasoning}, 2019.

\bibitem{ran2022surface}
Haoxi Ran, Jun Liu, and Chengjie Wang.
\newblock Surface representation for point clouds.
\newblock {\em CVPR}, 2022.

\bibitem{Riegler2017OctNet}
Gernot Riegler, Ali~Osman Ulusoy, and Andreas Geiger.
\newblock {OctNet}: {L}earning deep {3D} representations at high resolutions.
\newblock {\em CVPR}, 2017.

\bibitem{robert2022learning}
Damien Robert, Bruno Vallet, and Loic Landrieu.
\newblock Learning multi-view aggregation in the wild for large-scale 3{D}
  semantic segmentation.
\newblock {\em CVPR}, 2022.

\bibitem{ronneberger2015u}
Olaf Ronneberger, Philipp Fischer, and Thomas Brox.
\newblock U-{N}et: {C}onvolutional networks for biomedical image segmentation.
\newblock {\em MICCAI}, 2015.

\bibitem{simonovsky2017dynamic}
Martin Simonovsky and Nikos Komodakis.
\newblock Dynamic edge-conditioned filters in convolutional neural networks on
  graphs.
\newblock {\em CVPR}, 2017.

\bibitem{thomas2019kpconv}
Hugues Thomas, Charles~R Qi, Jean-Emmanuel Deschaud, Beatriz Marcotegui,
  Fran{\c{c}}ois Goulette, and Leonidas~J Guibas.
\newblock {KPC}onv: Flexible and deformable convolution for point clouds.
\newblock {\em ICCV}, 2019.

\bibitem{thyagharajan2022segment}
Anirud Thyagharajan, Benjamin Ummenhofer, Prashant Laddha, Om~Ji Omer, and
  Sreenivas Subramoney.
\newblock Segment-fusion: Hierarchical context fusion for robust 3{D} semantic
  segmentation.
\newblock {\em CVPR}, 2022.

\bibitem{tu2018learning}
Wei-Chih Tu, Ming-Yu Liu, Varun Jampani, Deqing Sun, Shao-Yi Chien, Ming-Hsuan
  Yang, and Jan Kautz.
\newblock Learning superpixels with segmentation-aware affinity loss.
\newblock {\em CVPR}, 2018.

\bibitem{varney2020dales}
Nina Varney, Vijayan~K Asari, and Quinn Graehling.
\newblock {DALES: A large-scale aerial LiDAR} data set for semantic
  segmentation.
\newblock {\em CVPR Workshops}, 2020.

\bibitem{vaswani2017attention}
Ashish Vaswani, Noam Shazeer, Niki Parmar, Jakob Uszkoreit, Llion Jones,
  Aidan~N Gomez, Lukasz Kaiser, and Illia Polosukhin.
\newblock Attention is all you need.
\newblock {\em NeurIPS}, 2017.

\bibitem{velivckovic2018graph}
Petar Veli{\v{c}}kovi{\'c}, Guillem Cucurull, Arantxa Casanova, Adriana Romero,
  Pietro Lio, and Yoshua Bengio.
\newblock Graph attention networks.
\newblock {\em ICLR}, 2018.

\bibitem{Wang2022WindowNE}
Qi Wang, Shengge Shi, Jiahui Li, Wuming Jiang, and Xiangde Zhang.
\newblock Window normalization: Enhancing point cloud understanding by unifying
  inconsistent point densities.
\newblock 2022.

\bibitem{xu2016hierarchical}
Yongchao Xu, Thierry G{\'e}raud, and Laurent Najman.
\newblock Hierarchical image simplification and segmentation based on
  mumford--shah-salient level line selection.
\newblock {\em Pattern Recognition Letters}, 2016.

\bibitem{zhang2022nested}
Zizhao Zhang, Han Zhang, Long Zhao, Ting Chen, Sercan~{\"O} Arik, and Tomas
  Pfister.
\newblock Nested hierarchical transformer: Towards accurate, data-efficient and
  interpretable visual understanding.
\newblock {\em AAAI}, 2022.

\bibitem{zhao2021point}
Hengshuang Zhao, Li Jiang, Jiaya Jia, Philip~HS Torr, and Vladlen Koltun.
\newblock Point transformer.
\newblock {\em ICCV}, 2021.

\end{thebibliography}
}

\ARXIV{
    \FloatBarrier
    \pagebreak
    \balance
    \section*{\centering \LARGE Appendix}
    \setcounter{section}{0}
    \setcounter{figure}{0}
    \setcounter{table}{0}
    \renewcommand*{\theHsection}{appendix.\the\value{section}}
    \renewcommand\thefigure{\arabic{figure}}
    \renewcommand\thetable{\arabic{table}}
    \renewcommand\thefigure{A-\arabic{figure}}
\renewcommand\thesection{A-\arabic{section}}
\renewcommand\thetable{A-\arabic{table}}
\renewcommand\theequation{A-\arabic{equation}}
\renewcommand\thealgorithm{A-\arabic{algorithm}} 
\setcounter{equation}{0}
\setcounter{section}{0}
\setcounter{figure}{0}
\setcounter{algorithm}{0}
\setcounter{table}{0}

\begin{figure*}[ht!]
\centering
\begin{tabular}{@{}ccc@{}}

\begin{subfigure}[b]{0.3\textwidth}
  \includegraphics[width=\linewidth]{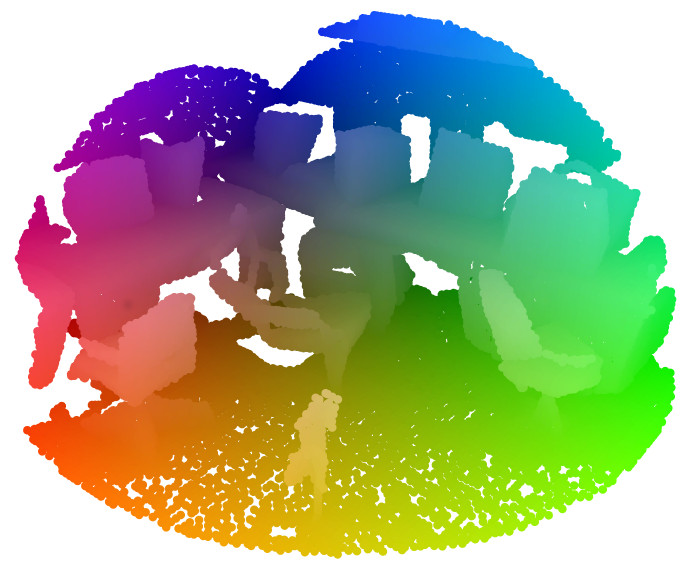}
  \caption{Position}
  \label{fig:tool:rgb}
\end{subfigure}
&
\begin{subfigure}[b]{0.3\textwidth}
  \includegraphics[width=\linewidth]{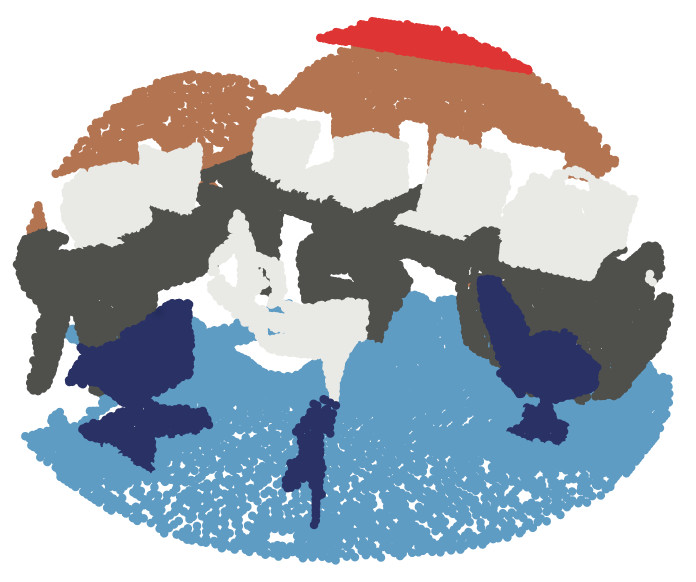}
  \caption{Ground Truth}
  \label{fig:tool:gt}
\end{subfigure}
 & 
\begin{subfigure}[b]{0.3\textwidth}
  \includegraphics[width=\linewidth]{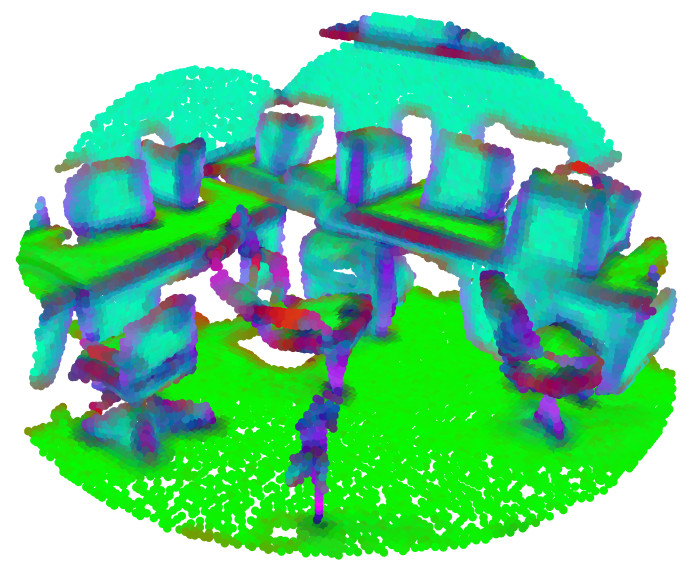}
  \caption{\textcolor{red}{Linearity}, \textcolor{green}{Planarity} \& \textcolor{blue}{Verticality}}
  \label{fig:tool:feat}
\end{subfigure}
\\

\begin{subfigure}[b]{0.3\textwidth}
  \includegraphics[width=\linewidth]{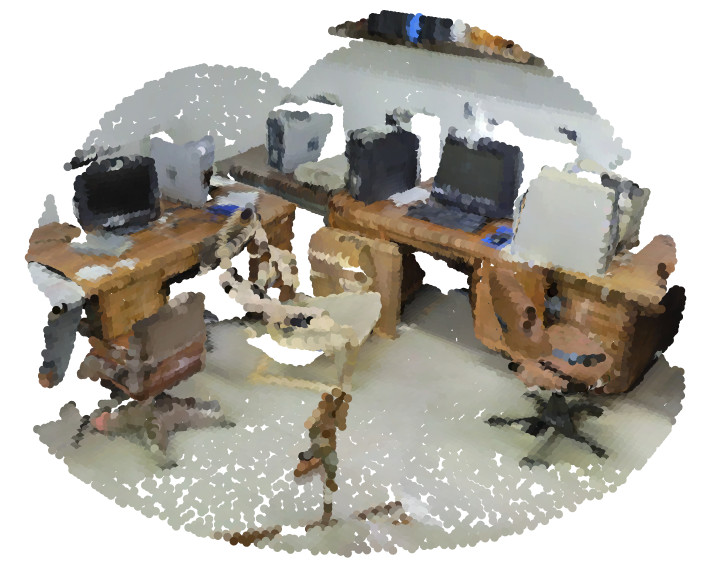}
  \caption{RGB}
  \label{fig:tool:rgb}
\end{subfigure}
&
\begin{subfigure}[b]{0.3\textwidth}
  \includegraphics[width=\linewidth]{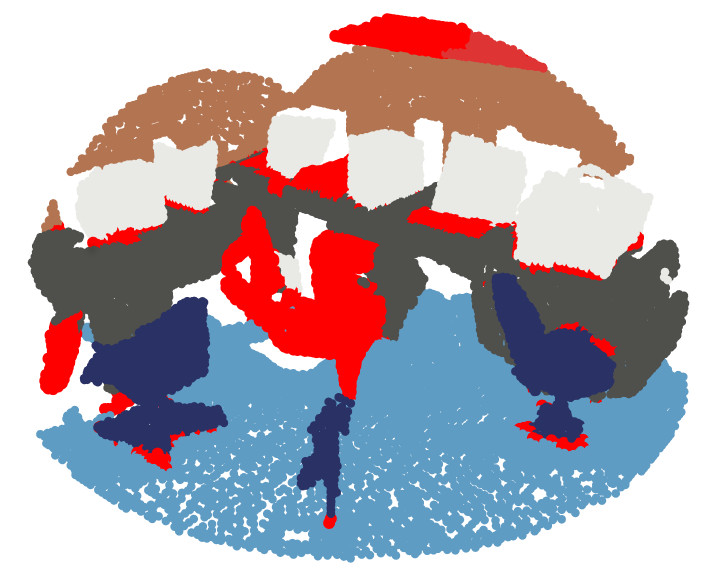}
  \caption{Predictions \& Errors}
  \label{fig:tool:pred}
\end{subfigure}
&
\begin{subfigure}[b]{0.35\textwidth}
  \includegraphics[width=\linewidth]{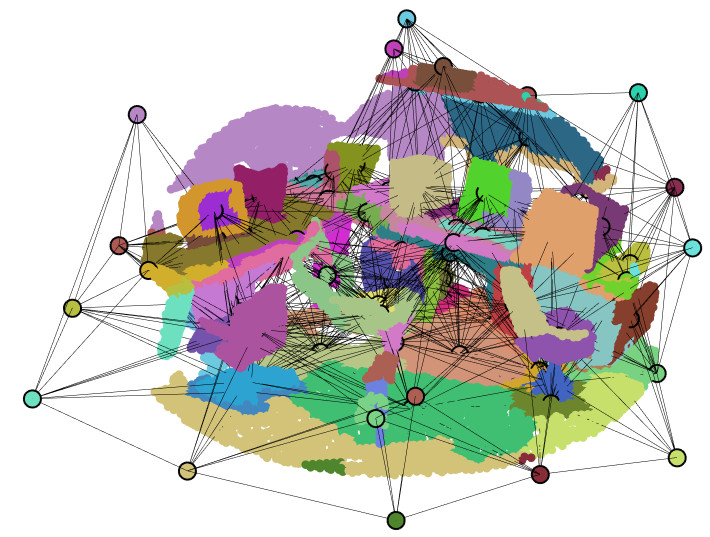}
  \caption{Level-2}
  \label{fig:tool:level2}
\end{subfigure}
\\

\end{tabular}

\caption{{\bf Interactive Visualization.} Our interactive viewing tool allows for the manipulation and visualization of sample point clouds colorized according to their position (a), semantic labels (b), selected geometric features (c), radiometry (d), and to visualize our network's prediction (e) and partitions (f).}

\label{fig:visu}
\end{figure*}

In this document, we introduce our interactive visualization tool (\secref{sec:visu}), share our source code (\secref{sec:code}), discuss limitations of our approach (\secref{sec:limitations}), provide a description (\secref{sec:hf}) and an analysis (\secref{sec:hfablation}) of all handcrafted features used by our method, detail the construction of the superpoint-graphs (\secref{sec:graphs}) and the partition process (\secref{sec:pcp}), and provide guidelines on how to choose the partition's hyperparameters (\secref{sec:hyper}). Finally, we clarify our architecture parameters (\secref{sec:implem}), explore our model's salability (\secref{sec:scalingablation}) and supervision (\secref{sec:lossablation}), detail the class-wise performance of our approach on each dataset (\secref{sec:classwise}), and the color maps used in the illustrations of the main paper (\figref{fig:colormaps}).

\section{Interactive Visualization}
\label{sec:visu}

We release for this project an interactive plotly visualization tool that produces HTML files compatible with any browser. As shown in \figref{fig:visu}, we can visualize samples from S3DIS, KITTI-360, and DALES with different point attributes and from any angle. These visualizations were instrumental in designing and validating our model and we hope that they will facilitate the reader's understanding as well.

\section{Source Code}
\label{sec:code}

We make our source code publicly available at \GITHUB.
The code provides all necessary instructions for installing and navigating the project, simple commands to reproduce our main results on all datasets, ready-to-use pretrained models, and ready-to-use notebooks. 

Our method is developed in PyTorch and relies on PyTorch Geometric, PyTorch Lightning, and Hydra. 

\section{Limitations}
\label{sec:limitations}
Our model provides significant advantages in terms of speed and compacity but  also comes with its own set of limitations.

\paragraph{Overfitting and Scaling.} The superpoint approach drastically simplifies and compresses the training sets: the $274$m 3D points of S3DIS are captured by a geometry-driven multilevel graph structure with fewer than $1.25$m nodes.
While this simplification favors the compacity and speed of the training of the model, this can lead to overfitting when using \SHORTHAND configurations with more parameters, as shown in \secref{sec:scalingablation}.
Scaling our model to millions of parameters may only yield better results for training sets that are sufficiently large, diverse, and complex.

\paragraph{Errors in the Partition.} Object boundaries lacking obvious discontinuities, such as curbs vs. roads or whiteboards vs. walls, are not well recovered by our partition. As partition errors cannot be corrected with our approach, this may lead to classification errors.
To improve this, we could replace our handcrafted point descriptors (\secref{sec:hf}) with features directly learned for partitioning \cite{landrieu2019point,hui2021superpoint}.
However, such methods significantly increase the preprocessing time, contradicting our current focus on efficiency.
In line with \cite{hsu2020incorporating,ran2022surface}, we use easy-to-compute yet expressive handcrafted features. 
Our model \SHORTHANDNANO without point encoder relies purely on such features and reaches $70.8$ mIoU on S3DIS 6-Fold with only $27$k param, illustrating this expressivity.

\paragraph{Learning Through the Partition.} The idea of learning point and adjacency features directly end-to-end is a promising research direction to improve our model. However, this implies efficiently backpropagating through superpoint hard assignments, which remains an open problem. Furthermore, such a method would consider individual 3D points during training, which would necessitate to perform the partitioning step multiple times during training time, which may negate the efficiency of our method

\paragraph{Predictions.}
Finally, our method predicts labels at the superpoint level $P_{1}$ and not individual 3D points. 
Since this may limit the maximum performance achievable by our approach, we could consider adding an upsampling layer to make point-level predictions. However, this does not appear to us as the most profitable research direction. Indeed, this may negate some of the efficiency of our method. Furthermore, as shown in the ablation study \textcolor{red}{4.3 d)} of the main paper, the ``oracle'' model outperforms ours by a large margin. This may indicate that performance improvements should primarily be searched in superpoint classification rather than in improving the partition.

Our model also learns features for superpoints and not individual 3D points. This may limit downstream tasks requiring 3D point features, such as surface reconstruction or panoptic segmentation.
However, we argue that specific adaptations could be explored to perform these tasks at the superpoint level.

\section{Handcrafted Features}
\label{sec:hf}

\begin{figure*}[ht!]
\centering
\begin{tabular}{@{}ccc@{}}

\begin{subfigure}[b]{0.3\textwidth}
  \includegraphics[width=\linewidth]{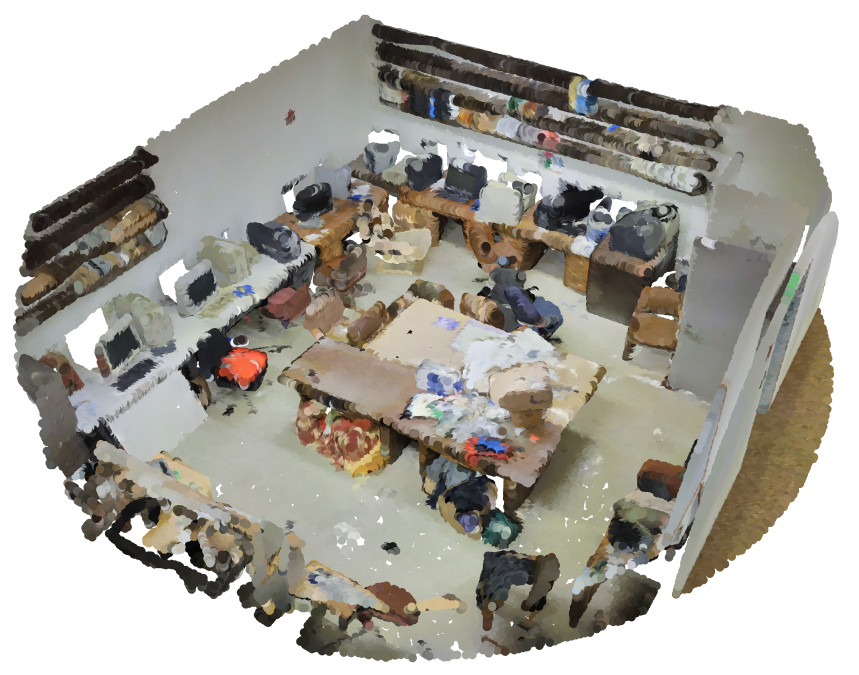}
  \caption{Input}
  \label{fig:geof:rgb}
\end{subfigure}
&
\begin{subfigure}[b]{0.3\textwidth}
  \includegraphics[width=\linewidth]{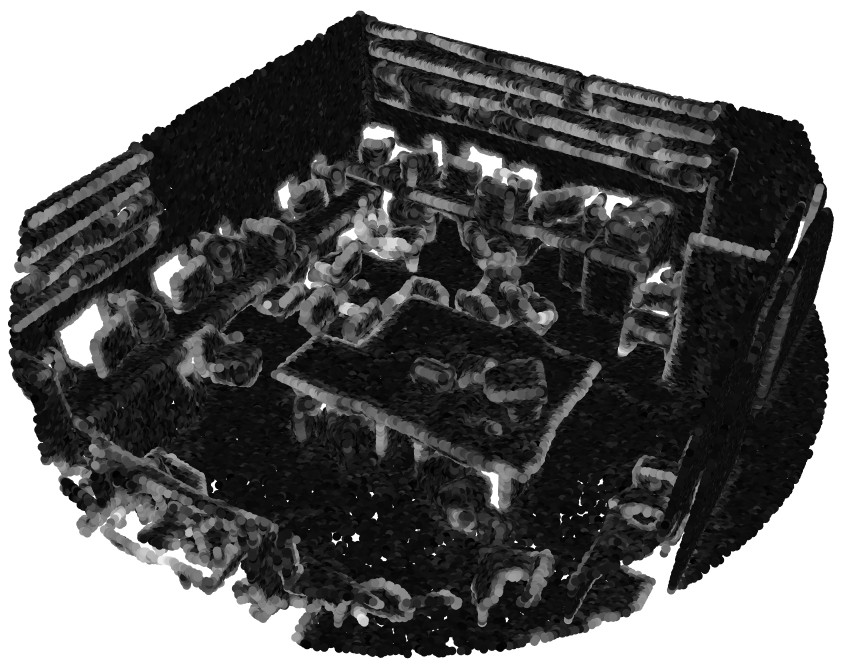}
  \caption{Linearity}
  \label{fig:geof:linearity}
\end{subfigure}
 & 
\begin{subfigure}[b]{0.3\textwidth}
  \includegraphics[width=\linewidth]{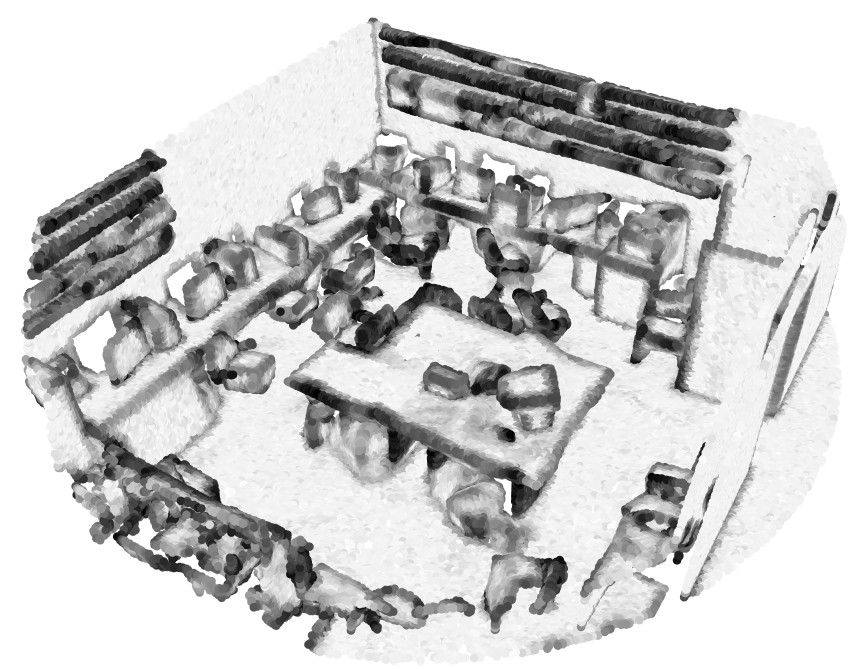}
  \caption{Planarity}
  \label{fig:geof:planarity}
\end{subfigure}
\\

\begin{subfigure}[b]{0.3\textwidth}
  \includegraphics[width=\linewidth]{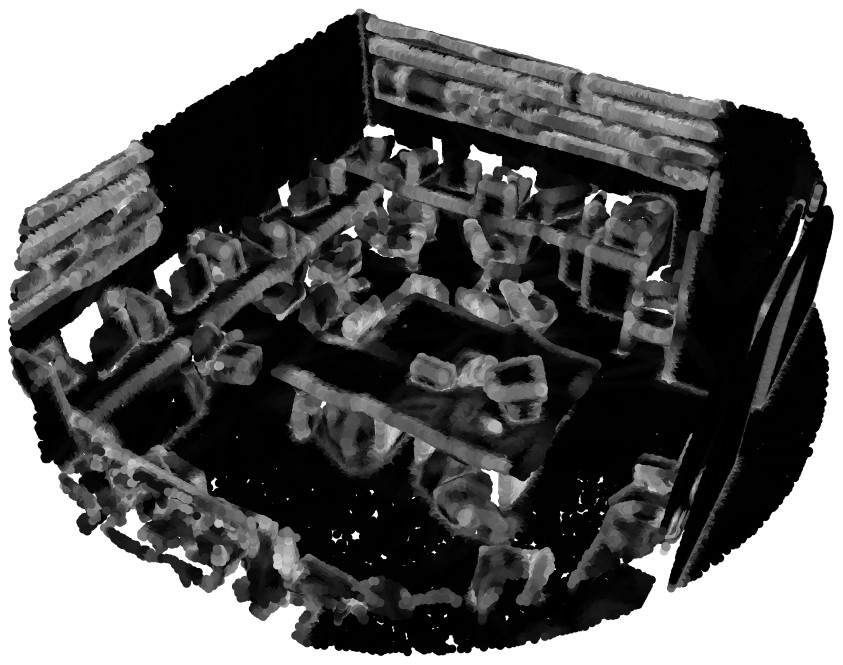}
  \caption{Scattering}
  \label{fig:geof:scattering}
\end{subfigure}
&
\begin{subfigure}[b]{0.3\textwidth}
  \includegraphics[width=\linewidth]{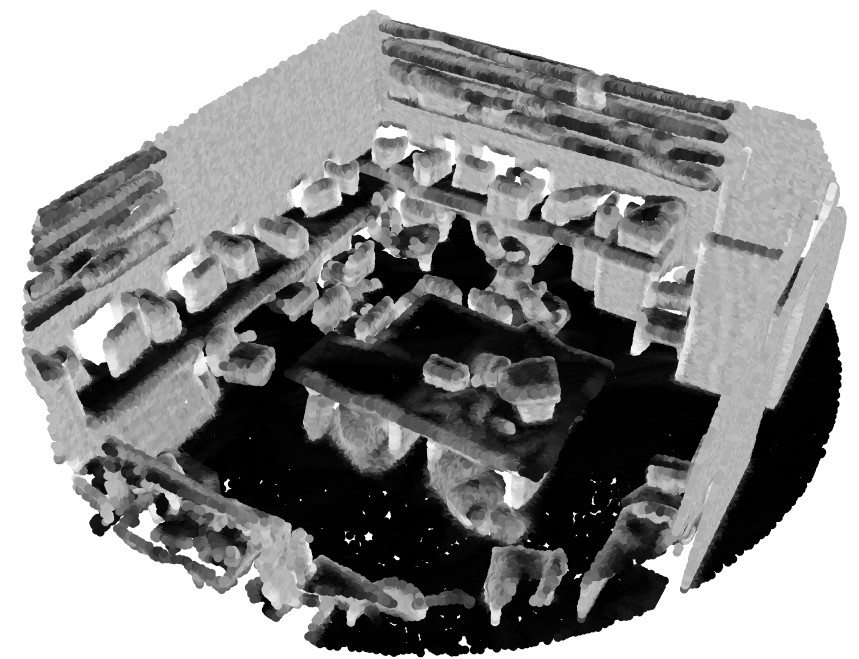}
  \caption{Verticality}
  \label{fig:geof:verticality}
\end{subfigure}
&
\begin{subfigure}[b]{0.3\textwidth}
  \includegraphics[width=\linewidth]{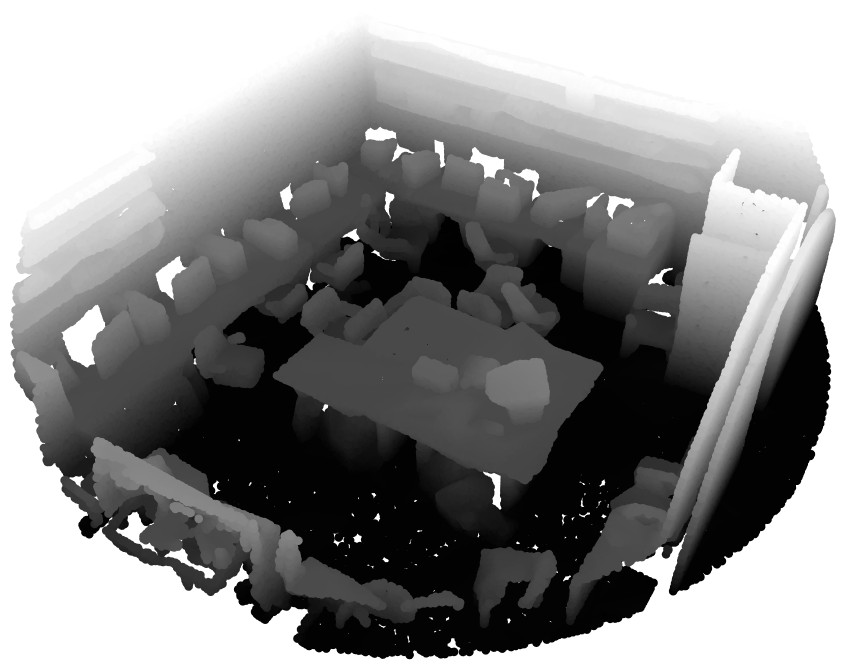}
  \caption{Elevation}
  \label{fig:geof:elevation}
\end{subfigure}
\\

\end{tabular}

\caption{{\bf Point Geometric Features.} Given an input cloud (a), the computed PCA-based geometric features (b, c, d, e) and distance to the ground (f) offer a simple characterization of the local geometry around each point.}

\label{fig:geof}
\end{figure*}

Our method relies on simple handcrafted features to build the hierarchical partition and learn meaningful %
{points and adjacency relationships.}
In this section, we provide further details on the definition of these features and how to compute them.
It is important to note that these features are only computed once during preprocessing, and thanks to our optimized implementation, this step only takes a few minutes.

\paragraph{Point Features.} We can associate each 3D point with a set of 
{$8$ easy-to-compute handcrafted features, described below.}

\begin{itemize}
    \item \textit{Radiometric features} (3 or 1): RGB colors are available for S3DIS and KITTI-360, and intensity values for DALES. These radiometric features are normalized to $[0, 1]$ at preprocessing time. 
    For KITTI-360, we find that using the HSV color model yields better results.

    \item \textit{Geometric features} (5): We use PCA-based features: \textit{linearity}, \textit{planarity}, \textit{scattering}, \cite{demantke2011dimensionality} and \textit{verticality} \cite{guinard2017weakly}, computed on the set of $50$-nearest neighbors of each point. This neighbor search is only computed once during preprocessing and is also necessary to build the graph $\cG$.
    We also define \textit{elevation} as the distance between a point and the ground below it. Since the ground is neither necessarily flat nor horizontal, we use the RANSAC algorithm~\cite{fischler1981random} on a coarse subsampling of the scene to find a ground plane. We normalize the elevation by dividing it by $4$ for S3DIS and $20$ for DALES and KITTI-360.

\end{itemize}

At preprocessing time, we only use radiometric and geometric features to compute the hierarchical partition. At training time, \SHORTHAND computes point embeddings by mapping all available point features, along with the normalized point position to a vector of size $D_\text{point}$ with a dedicated MLP $\phi^0_\text{enc}$.

We provide an illustration of the geometric point features in \figref{fig:geof}, to help the reader apprehend these simple geometric descriptors.

\paragraph{Adjacency Features.}~The relationship between adjacent superpoints provides crucial information to leverage their context. For each edge of the superpoint-graph, we compute the $18$ following features: \\
\begin{itemize}
\item \textit{Interface features (7)}: 
All adjacent superpoints share an \emph{interface}, \ie pairs of points from each superpoint that are close and share a line of sight. SuperpointGraph \cite{landrieu2018large} uses the Delaunay triangulation of the entire point cloud to compute such interfaces, while we propose a faster heuristic approach in \secref{sec:graphs} called the \emph{Approximate Superpoint Gap algorithm}. Each pair of points of an interface defines an offset, \ie a vector pointing from one superpoint to its neighbor. We compute the mean offset (dim 3), the mean offset length (dim 1), and the standard deviation of the offset in each canonical direction (dim 3).

\item \textit{Ratio features (4)}: 
As defined in \cite{landrieu2018large}, we characterize each pair of adjacent superpoints with the ratio of their \textit{lengths}, \textit{surfaces}, \textit{volumes}, and \textit{point counts}.

\item \textit{Pose features (7)}: For each superpoint, we define a normal vector as its principal component with the smallest eigenvalue. We then characterize the relative position between two superpoints with the cosine of the angle between the superpoint normal vectors (dim: 1) and between each of the two superpoints' normal and the mean offset direction (dim: 2). Additionally, the offset between the centroids of the superpoints is used to compute the centroid distance (dim: 1) and the unit-normalized centroid offset direction (dim: 3).

\end{itemize}

Note that the mean offset and the ratio features are not symmetric and imply that the edges of the superpoint-graphs are oriented.
As mentioned in Section \textcolor{red}{$3.3$}, a network $\phi_\text{adj}^i$ maps these handcrafted features to a vector of size $D_\text{key}+D_\text{que}+D_\text{val}$, for each level $i\geq1$ of the encoder and the decoder.

\section{Influence of Handcrafted Features}
\label{sec:hfablation}

\begin{table}
\caption{\textbf{Ablation on Handcrafted Features.} Impact of handcrafted features on the mIoU for all tested datasets.}
\label{tab:hfablation}
\centering
\small{
\begin{tabular}{@{}lccc@{}}
    \toprule
    Experiment & S3DIS & KITTI & DALES \\
     & 6-Fold & 360 Val & \\
    \midrule
    Best Model           & 76.0 & 63.5 & 79.6 \\
    \\
    \multicolumn{4}{c}{\textit{a)  Point Features}} \\
    \midrule
    No radiometric feat. & -2.7 & -4.0 & -1.2 \\
    No geometric feat.   & -0.7 & -4.1 & -1.4 \\
    \\
    \multicolumn{4}{c}{\textit{b) Adjacency Features}} \\
    \midrule 
    No interface feat.   & -0.2 & -0.6 & -0.7 \\
    No ratio feat.       & -1.1 & -2.2 & -0.4 \\
    No pose feat.        & -5.5 & -1.2 & -0.8 \\
    \multicolumn{4}{c}{\textit{c) Room Features}} \\
    \midrule
    Room-level samples   & -3.8 & - & - \\
    Normalized Room pos.        & -0.7 & - & - \\   
    \bottomrule
\end{tabular}}
\end{table}

In \tabref{tab:hfablation}, we quantify the impact of the handcrafted features detailed in \secref{sec:hf} on performance. To this end, we retrain \SHORTHAND without each feature group and evaluate the prediction on S3DIS Area~5. 

\paragraph{a) Point Features.} Our experiments show that removing radiometric features has a strong impact on performance, with a drop 
of $2.7$ to $4.0$ mIoU.
In contrast, removing geometric features 
results in a performance drop 
of $0.7$ on S3DIS, but $4.1$ on KITTI-360.

We observe that both outdoor datasets strongly benefit from local geometric features, which we hypothesize is due to their lower resolution and noise level.
These results indicate that radiometric features play an important role for all datasets and that geometric features may facilitate learning on noisy or subsampled datasets.

\paragraph{b) Adjacency Features.} The analysis of the impact of adjacency features on our model's performance indicates that they play a crucial role in leveraging contextual information from superpoints: removing all adjacency features leads to a significant drop of 
$3.0$ to $6.3$ mIoU points on the datasets, as shown in \textcolor{red}{4.3 b)} of the main paper. 
Among the different types of adjacency features, 
pose features appear particularly useful in characterizing the adjacency relationships between superpoints of S3DIS, while interface features have a smaller impact.
These results suggest that the relative pose 
of objects in the scene may have more influence on the 3D semantic analysis performed by our model than the precise characterization of their interface.
On the other hand, interface and ratio features seem to have more impact on outdoor datasets, while the pose information seems to be less informative in the semantic understanding of the scene.

\paragraph{c)  S3DIS Room Partition.} The S3DIS dataset is divided into individual rooms aligned along the $x$ and $y$ axes.
This setup simplifies the classification of classes such as walls, doors, or windows as they are consistently located at the edge of the room samples. Some methods also add normalized room coordinates to each points.
However, we argue that this partition may not generalize well to other environments, such as open offices, industrial facilities, or mobile mapping acquisitions, which cannot naturally be split into rooms.

To address this limitation, we use the absolute room positions to reconstruct the entire floor of each S3DIS area \cite{thomas2019kpconv,chaton2020torch}. This enables our model to consider large multi-room samples, resulting in a performance increase of $3.8$ points. This highlights the advantage of capturing long-range contextual information.
Additionally, we remark that \SHORTHAND performs better without using room-normalized coordinates, which may lead to overfitting and poor performance on layouts that deviate from the room-based structure of the S3DIS dataset such as large amphitheaters.

\section{Superpoint-Graphs Computation}
\label{sec:graphs}

The Superpoint Graph method by Landrieu and Simonovsky \cite{landrieu2018large} builds a graph from a point cloud using Delaunay triangulation, which can take a long time for large point clouds. In contrast, our approach connects two superpoints in $\cP_{i}$, where $i\geq1$ if their closest points are within a distance gap $\epsilon_i>0$. However, computing pairwise distances for all points is computationally expensive. We propose a heuristic to approximately find the closest pair of points for two superpoints, see Algorithm \ref{alg:pointnn}. We also accelerate the computation of adjacent superpoints by approximating only for superpoints with centroids closer than the sum of their radii plus the gap distance. This approximation helps to reduce the number of computations required for adjacency computation, which leads to faster processing times. All steps involved in the computation of our superpoint-graph are implemented on the GPU to further enhance computational efficiency.

\begin{algorithm}
\caption{Approximate Superpoint Gap}\label{alg:pointnn}
\begin{algorithmic}
\State{\textbf{Input:} superpoints $p_1$ and $p_2$, $\text{num\_steps}$}
    \State{$c_1 \gets  \text{centroid}(p_1)$ }
    \State{$c_2 \gets  \text{centroid}(p_2)$ }

    \For{$s \in \text{num\_steps}$}
        \State{$c_2 \gets \argmin_{p \in p2} \Vert c_1-p\Vert$}
        \State{$c_1 \gets \argmin_{p \in p1} \Vert c_2-p\Vert$}
    \EndFor
    
\Return{$\Vert c_1-c_2\Vert$}

\end{algorithmic}
\end{algorithm}

 \if 1 0

\begin{algorithm}
\caption{Superpoint-Graph Computation}\label{alg:segmentnn}
\begin{algorithmic}
\State{\textbf{Input:} $\cP_{i}$ partition, $\cP_{0}$ points, $\epsilon_i$ gap}

    \# compute the centroid and distance of superpoints
    \State{$\text{diam} \gets$ diameters of all superpoints of $\cP_{i}$ } 
    \State{$\text{center} \gets$ centroids of all superpoints of $\cP_{i}$}

    \State{$\text{max\_diam} \gets \text{max}(\text{diam})$}

    \# 
    \State{$\text{neigh}, dist \gets \text{radius\_nn}(center, max\_diam)$}
    
    \For{$(p_1, p_2), (d) \in zip(neigh, dist)$}
        \If{$\epsilon_i + (diam[p_1] + diam[p_2]) / 2 < d$}
            \State{$neigh, dist \gets$ trim out too-far $p_1, p_2$ pair }
        \EndIf
    \EndFor
    
    \For{$p_1, p_2 \in neigh$}
        \State{$d \gets \text{iterative\_smallest\_distance}(p_1, p_2, \cP_{0})$}
        \If{$\epsilon_i < d$}
            \State{$neigh, dist \gets$ trim out too-far $p_1, p_2$ pair }
        \EndIf
    \EndFor

\Return{neigh}

\end{algorithmic}
\end{algorithm}

\begin{algorithm}
\caption{Superpoint-Graph Computation}\label{alg:segmentnn}
\begin{algorithmic}
\State{\textbf{Input:} $\cP_{i}$ partition, $\cP_{0}$ points, $\epsilon_i$ gap}

    \State{$diam \gets$ array of size $i$ } 
    \State{$center \gets$ array of size $i$ }
    \For{$p \in \cP_{i}$}
        \State{$diam[p] \gets  \text{compute\_diameter}(p)$ }
        \State{$center[p] \gets  \text{compute\_centroid}(p)$ }
    \EndFor
    
    \State{$max\_diam \gets \text{max}(diam)$}
    
    \State{$neigh, dist \gets \text{radius\_nn}(center, max\_diam)$}
    
    \For{$(p_1, p_2), (d) \in zip(neigh, dist)$}
        \If{$\epsilon_i + (diam[p_1] + diam[p_2]) / 2 < d$}
            \State{$neigh, dist \gets$ trim out too-far $p_1, p_2$ pair }
        \EndIf
    \EndFor
    
    \For{$p_1, p_2 \in neigh$}
        \State{$d \gets \text{iterative\_smallest\_distance}(p_1, p_2, \cP_{0})$}
        \If{$\epsilon_i < d$}
            \State{$neigh, dist \gets$ trim out too-far $p_1, p_2$ pair }
        \EndIf
    \EndFor
\Return{neigh}

\end{algorithmic}
\end{algorithm}
\fi

Recovering the interface between two adjacent superpoints as evoked in \secref{sec:hf} involves a notion of visibility: we connect points from each superpoint which are \emph{facing} each other. This can be a challenging and ambiguous problem, which SuperPoint Graph \cite{landrieu2016cut} tackles using a Delaunay triangulation of the points. However, this method is impractical for large point clouds. To address this issue, we propose a heuristic approach with the following steps: (i) first, we use the Approximate Superpoint Gap algorithm to compute the approximate nearest points for each superpoint. Then, we restrict the search to only consider points within a certain distance of the nearest points. Finally, we match the points by sorting them along the principal component of the selected points.

\section{Details on Hierarchical Partitions}
\label{sec:pcp}
We present here a more detailed explanation of the hierarchical partition process. We define for each point $c$ of $\cC$ a feature $f_c$ of dimension $D$, and $G\eqdef(\cC,\cE,w)$ is the k-nn adjacency between the points, with $w \in \bR_+^\cE$ a nonnegative proximity value. Our goal is to compute a hierarchical multilevel partition of the point cloud into superpoints homogeneous with respect to $f$ at increasing coarseness. 

\paragraph{Piecewise Constant Approximation on a Graph.}
We first explain how to compute a single-level partition of the point cloud.
We consider the pointwise features $f_c$ as a $D$-dimensional signal $f \in \bR^{D \times \vert \cC \vert}$ defined on the nodes of the weighted graph $G\eqdef(\cC,\cE,w)$. 
We first define an energy $\cJ(e; f, \cG, \lambda)$ measuring the fidelity between a vertex-valued signal $e \in \bR^{D \times \vert \cC \vert}$ and the length of its contours, defined as the weight of the cut between its constant components \cite{landrieu2016cut}:
\begin{align}\label{eq:gmpp}
\cJ(e; f, \cG, \lambda)
\eqdef
\Vert e - {f}\Vert^2
+ \lambda\!\!\!
\sum_{(u,v) \in \cE}\!\!\!
w_{u,v}
\left[
e_u \neq e_v
\right]~,
\end{align}
with $\lambda \in \bR_+$ a regularization strength and $[a \neq b]$ the function equals to $0$ if $a=b$ and $1$ otherwise. Minimizers of $\cJ$ are approximations of $f$ that are piecewise constant with respect to a partition with simple contours in $\cG$.

We can characterize such signal $e \in \bR^{D \times \vert \cC \vert}$ by the coarsest partition $\cP^e$ of $\cP$ and its associated variable $f^e \in \bR^{D\times \vert \cP^e \vert}$ such that $e$ is constant within each segment $p$ of $\cP^e$ with value $f^e_p$. 
The partition $\cP^e$ also induces a graph $\hat{\cG}^e\eqdef(\cP^e,\cE^e,w^e)$ with $\cE^e$ linking the component of $\cP^e$ adjacent in $\cG$ and $w^e$ the weight of the cut between adjacent elements of $P^e$: 
\begin{align}
   &\cE^e \eqdef \{(U,V) \mid U,V \in \cP^e, (U \times V) \cap \cE \neq \varnothing \}\\
  &\text{For}\; (U,V) \in \cE^e,\; w^e_{U,V}\eqdef\sum_{(u,v) \in U \times V \cap \cE} w_{u,v} 
\end{align}

We denote by $\parti{e}$ the function mapping $e$ to these uniquely defined variables:  
\begin{align}
f^e, \cP^e, \hat{\cG}^e \eqdef \parti{e}~.
\end{align}

\paragraph{Point Cloud Hierarchical Partition.}
A set of partitions $\cP\eqdef[\cP_0, \cdots, \cP_i]$ defines a hierarchical partition of $\cC$ with $I$ levels if
$\cP_0=\cC$ and $P_{i+1}$ is a partition of $P_{i}$ for $i \in [0,I-1]$. 
We propose to use the formulations above to define a hierarchical partition of the point cloud $\cC$ characterized by a list $\lambda_1, \cdots, \lambda_{I}$ of nonnegative regularization strengths defining the coarseness of the successive partitions. In particular, We chose $\lambda_1$ such that $\vert \cP_1 \vert / \vert \cP_0 \sim 30$ in our experiments.
 
We first define $\hat{\cG}_0$ as the point-level adjacency graph $\hat{\cG}$ and $f_0$ as $f$. We can now define the levels of a hierarchical partition $\cP_i$ for $i \in [1,I]$:
\begin{align}\label{eq:nested}
    f_{i}, \cP_{i}, \hat{\cG}_{i}
    &\eqdef \partition (
        \argmin_{e \in \bR^{D \times \vert \cP_{i-1} \vert}}
        \cJ
        \left(
            e; f_{i-1}, \hat{\cG}_{i-1}, \lambda_{i-1}
        \right)
    ).
\end{align}
Given that the optimization problems defined in \eqref{eq:nested} for $i>1$ operate on the component graphs $\hat{\cG}_i$, which are smaller than $\hat{\cG}_0$, the first partition is the most demanding in terms of computation.

Note that we used the hat notation $\hat{\cG}_i$, because these graphs are only used for computing the hierarchical partitions $\cP_i$, and should be distinguished from the the superpoint graphs $\cG_i$ on which is based our self-attention mechanism, constructed from $\cP_i$ as explained in \secref{sec:graphs}.

\section{Parameterizing the Partition}
\label{sec:hyper}
We define $\cG$ as the $k=10$-nearest neighbor adjacency graph and set all edge weights $w$ to $1$. The point features $f_p$ whose piecewise constant approximation yields the partition are of three types: geometric, radiometric, and spatial.

Geometric features ensure that the superpoints are geometrically homogeneous and with simple shapes. We use the normalized dimensionality-based method described in \secref{sec:hf}. Radiometric features encourage the border of superpoints to follow the color contrast of the scene and are either RGB or intensity values; they must be normalized to fall in the [0,1] range. Lastly, we can add to each point their spatial coordinates with a normalization factor $\mu$ in $m^{-1}$ to limit the size of the superpoints. We recommend setting $\mu$ as the inverse of the maximum radius expected for a superpoint: the largest sought object (facade, wall, roof) or an application-dependent constraint.

The coarseness of the partitions depends on the regularization strength $\lambda$ as defined in \secref{eq:pcp}.
Finer partitions should generally lead to better results but to an increase in training time and memory requirement. We chose a ratio $\mid \cP_0 \mid / \mid \cP_1\mid \sim 30$ across all datasets as it proved to be a good compromise between efficiency and precision. Depending on the desired trade-off, different ratios can be chosen by trying other values of $\lambda$.

\if 1 0
\section{Influence of hyperparameters on partition}
This section must be \textbf{reassuring} to the reader: sure we are introducing new hyperparameters you are not familiar with but:
\begin{itemize}
    \item see how robust the model is to (most of) these
    \item here are fat-and-easy-to-reproduce experiments for choosing them and studying their importance on your dataset
\end{itemize}

The influence of hyperparameters is drawn by varying the regularization lambda for fixed values of a parameter, to obtain roughly ($10^2, 10^3, 10^4, 10^5, 10^6$) superpoints each time. Comparing Oracle mIoU, OA, compute speed 
\begin{itemize}
    \item voxel
    \item koutlier
    \item kfeat (20-50) (number of neighbors considered for pointwise features)
    \item radius knn (maximum radius for the neighbor search) 
    \item point feats (linearity, planarity, scattering, verticality, rgb, xyz\/sigma, ...)
    \item lambda xyz (supmat pour justif serieux handcrafted)
    \item lambda reg (0.05, 0.1, 0.4 viser 1 point par ordre de gdeur)
    \item split\_dampening\_ratio
    \item kadj (if it saturates fast, important message)
    \item cutoff 
    \item iterations
\end{itemize}

Remember that we want:
\begin{itemize}
    \item maximum semantic purity
    \item minimum preprocessing time
    \item minimum training time (ie minimum number of superpoints)
\end{itemize}
\fi 

\if 1 0
\section{Superpoint Dropout}
\label{sec:spdropout}

\begin{figure}
\centering
\includegraphics[width=\linewidth]{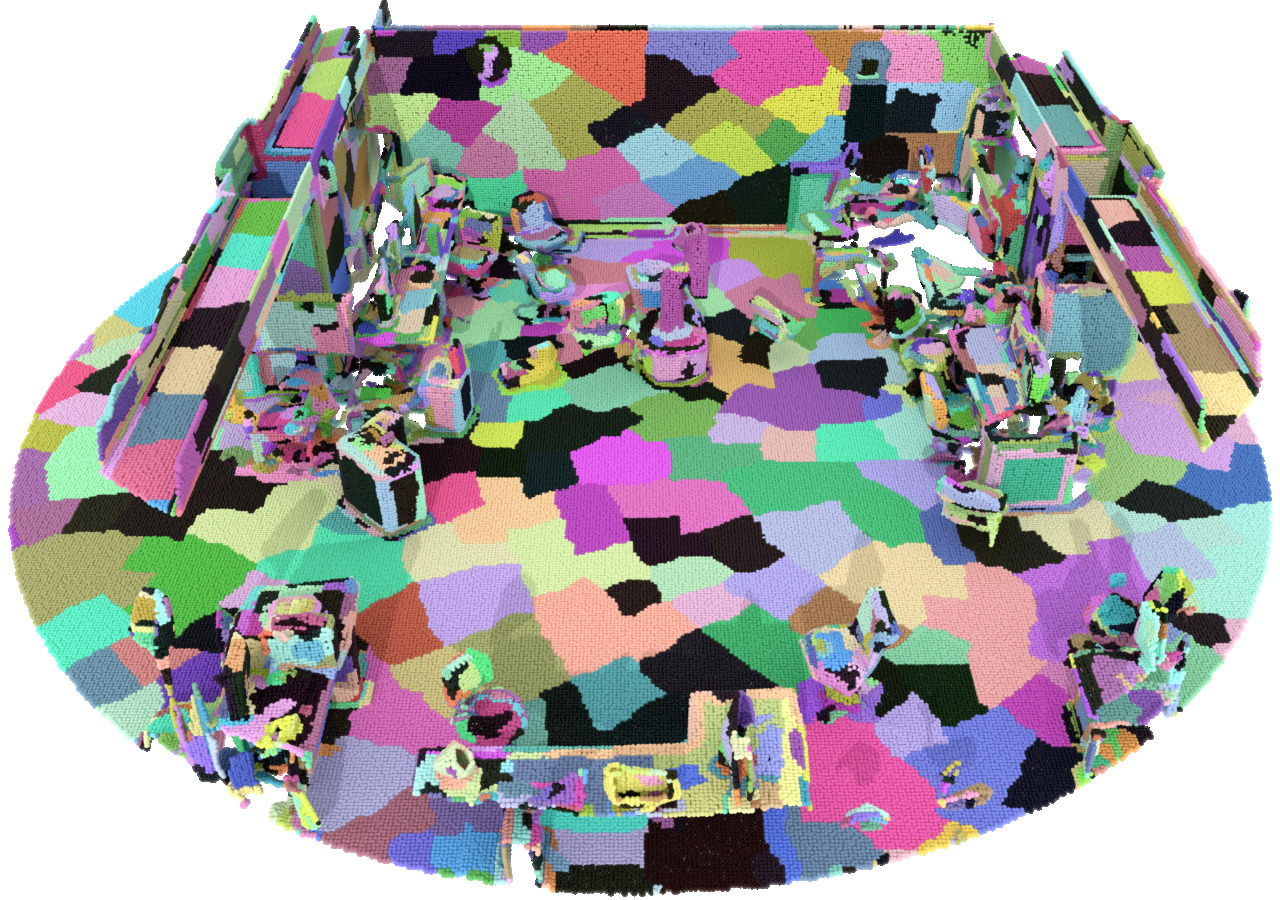}
\caption{{\bf Superpoint Dropout.} We can use the hierarchical partitions to guide augmentations such as masking geometrically-consistent parts of the considered scene. We represent in black the masked superpoints.}
\label{fig:dropout}
\end{figure}

We show the effect of our superpoint dropout on an S3DIS scene in \figref{fig:dropout}. As confirmed in \textcolor{red}{$4.3$}, removing geometrically-consistent parts of the scene offers a powerful augmentation for learning on relatively small datasets like S3DIS.

\fi

\section{Implementation Details}
\label{sec:implem}

\begin{table}
\caption{\textbf{Model Configuration.} We provide the detailed architecture of the
SPT-X architecture. In this paper, we use $X=64$ and $X=128$.
}
\label{tab:implementation}
\centering
\small{
\begin{tabular}{@{}lc@{}}
    \toprule
    Parameter & Value \\
    \midrule
     {\it Handcrafted features} \\
    $D_\text{point}^{\text{hf}}$ & $D_\text{point}^{\text{radio}} + D_\text{point}^{\text{geof}}$ \\
    $D_\text{adj}^{\text{hf}}$  & 18 \\
    \midrule
    {\it Embeddings sizes} \\
    $D_\text{point}$         & 128 \\
    $D_\text{adj}$           & 32 \\
    \midrule
    {\it Transformer blocks} \\
    $D_\text{val}$           & \underline{\textbf{X}} \\
    $D_\text{key}$           & 4 \\
    \# blocks encoder        & 3 \\
    \# blocks decoder        & 1 \\
    \# heads                 & 16 \\
    \midrule
    {\it MLPs} \\
    $\phi_\text{adj}^i$      & \footnotesize $[D_\text{adj}^{\text{hf}}, D_\text{adj}, D_\text{adj}, 3 D_\text{adj}]$  \\
    $\phi^0_\text{enc}$      & \footnotesize $[D_\text{point}^{\text{hf}} + D_\text{point}^{\text{pos}}, 32, 64, D_\text{point}]$ \\
    $\phi^1_\text{enc}$      & \footnotesize $[D_\text{point} + D_\text{point}^{\text{pos}}, D_\text{val}, D_\text{val}]$ \\
    $\phi^2_\text{enc}$      & \footnotesize $[D_\text{val} + D_\text{point}^{\text{pos}}, D_\text{val}, D_\text{val}]$ \\
    $\phi^1_\text{dec}$      & \footnotesize $[D_\text{val} + D_\text{val} + D_\text{point}^{\text{pos}}, D_\text{val}, D_\text{val}]$ \\

    \bottomrule
\end{tabular}}
\end{table}

\if 1 0
\begin{table*}
\caption{\textbf{Model Configuration.} Detailed description of the \SHORTHAND architecture parameters for each dataset.}
\label{tab:implementation}
\centering
\small{
\begin{tabular}{@{}lccc@{}}
    \toprule
     & S3DIS & KITTI-360 & DALES \\
 
    \midrule
    Activation & LeakyReLU & LeakyReLU & LeakyReLU \\
    Normalization & GraphNorm & GraphNorm & GraphNorm \\
    $D_\text{point}$         & 128 & 128 & 128 \\
    $D_\text{superpoint}$    & 64 & 32 & 32 \\
    $D_\text{adj}$           & 64 & 32 & 32 \\
    $D_\text{val}$           & 64 & 128 & 64 \\
    $D_\text{key}$           & 4 & 4 & 4 \\
    $\MLP_\text{point}$      & \footnotesize $[14,32,64,D_\text{point}]$      & \footnotesize $[11,32,64,D_\text{point}]$          & \footnotesize $[9,32,64,D_\text{point}]$ \\
    $\MLP_\text{superpoint}$ & \footnotesize $[20,D_\text{superpoint},D_\text{superpoint}]$          & \footnotesize $[17,D_\text{superpoint},D_\text{superpoint}]$              & \footnotesize $[15,D_\text{superpoint}, D_\text{superpoint}]$ \\
    $\MLP_\text{adj}$        & \footnotesize $[18,D_\text{adj},D_\text{adj}]$          & \footnotesize $[18,D_\text{adj},D_\text{adj}]$              & \footnotesize $[18,D_\text{adj},D_\text{adj}]$ \\
    $\phi^1_\text{enc}$      & \footnotesize $[D_\text{point}+D_\text{superpoint}+3,64,64]$                 & \footnotesize $[D_\text{point}+D_\text{superpoint}+3,128,128]$      & \footnotesize $[D_\text{point}+D_\text{superpoint}+3,64,64]$ \\
    $\phi^2_\text{enc}$      & \footnotesize $[64+D_\text{superpoint}+3,64,64]$                  & \footnotesize $[128+D_\text{superpoint}+3,128,128]$      & \footnotesize $[64+D_\text{superpoint}+3,64,64]$ \\
    $\phi^1_\text{dec}$      & \footnotesize $[64+64+D_\text{superpoint}+3,64,64]$               & \footnotesize $[128+128+D_\text{superpoint}+3,128,128]$  & \footnotesize $[64+64+D_\text{superpoint}+3,64,64]$ \\

    \# blocks encoder        & 3 & 3 & 3 \\
    \# blocks decoder        & 1 & 1 & 1 \\
    \# heads                 & 16 & 16 & 16 \\
    
    \bottomrule
\end{tabular}}
\end{table*}
\fi

We provide the exact parameterization of the \SHORTHAND architecture used for our experiments.   
All MLPs in the architecture use LeakyReLU activations and GraphNorm \cite{cai2021graphnorm} normalization. For simplicity, we represent an MLP by the list of its layer widths: $[\text{in\_channels}, \text{hidden\_channels}, \text{out\_channels}]$. 

\paragraph{Point Input Features.} {
We refer here to the dimension of point positions, radiometry, and geometric features as $D_\text{point}^{\text{pos}}=3$, $D_\text{point}^{\text{radio}}$, and $D_\text{point}^{\text{geof}}=4$ respectively. 
As seen in \secref{sec:hf}, S3DIS and KITTI-360 use $D_\text{point}^{\text{radio}}=3$, while DALES uses $D_\text{point}^{\text{radio}}=1$.
}

\paragraph{Model Architecture.}{
The exact architecture SPT-64 used for S3DIS and DALES is detailed in \tabref{tab:implementation}.
The other models evaluated are SPT-16, SPT-32, SPT-128 (used for KITTI-360), and SPT-256, which use the same parameters except for $D_\text{val}$. 

\paragraph{SPT-nano.} For SPT-nano, we use and $D_\text{val}=16$, $D_\text{adj}=16$, and $D_\text{key}=2$. As SPT-nano does not compute point embedding, it does not use $\phi^0$, and we set up $\phi^1_\text{enc}$ as $[D_\text{point}^{\text{hf}} + D_\text{point}^{\text{pos}}, D_\text{val}, D_\text{val}]$.
}

\section{Model Scalability}
\label{sec:scalingablation}

We study the scalability of SPT by comparing models with different parameter counts on each dataset. 
It is important to note that the superpoint approach drastically compresses the training set, which can lead to overfitting, see \secref{sec:limitations}. 
For example, as illustrated in \tabref{tab:scalingablation}, SPT-128 with $D_\text{val}={128}$ (777k param.) performs $1.4$ points below $D_\text{val}=64$ on S3DIS. 

We report a similar behavior for other hyperparameters: in \tabref{tab:qkdimablation}, $D_\text{key}=8$ instead of $4$ incurs a drop of $1.0$, while in \tabref{tab:headsablation}, $N_\text{heads}=32$ instead of $16$ a drop of $0.1$ point. For the larger KITTI-360 dataset ($13$m nodes), $D_\text{val}={128}$ performs $1.9$ points above $D_\text{val}=64$, but $5.4$ points above $D_\text{val}=256$ (2.7m param.).

\begin{table}[H]
\caption{\textbf{Impact of Model Scaling.} Impact of model size for each dataset.}
\label{tab:scalingablation}
\centering
\small{
\begin{tabular}{@{}lcccc@{}}
    \toprule
    Model & Size & S3DIS & KITTI & DALES \\
     & $\times10^6$ & 6-Fold & 360 Val & \\
    \midrule
    SPT-32  & 0.14 & 74.5 & 60.6 & 78.7 \\
    SPT-64  & 0.21 & \bf 76.0 & 61.6 & \bf 79.6 \\
    SPT-128 & 0.77 & 74.6 & \bf 63.5 & 78.8 \\
    SPT-256 & 1.80 & 74.0 & 58.1 & 77.6 \\
    \bottomrule
\end{tabular}}
\end{table}
\begin{table}[H]
\caption{\textbf{Impact of Query-Key Dimension.} Impact of $D_\text{key}$ on S3DIS 6-Fold.}
\label{tab:qkdimablation}
\centering
\small{
\begin{tabular}{@{}ccccc@{}}
    \toprule
    $D_\text{key}$  & 2 & 4 & 8 & 16 \\
    \midrule
    SPT-64 & 75.6 & \bf 76.0 & 75.0 & 74.7 \\
    \bottomrule
\end{tabular}}
\end{table}
\begin{table}[H]
\caption{\textbf{Impact of Heads Count.} Impact of the number of heads $N_\text{head}$ on the S3DIS 6-Fold performance.}
\label{tab:headsablation}
\centering
\small{
\begin{tabular}{@{}ccccc@{}}
    \toprule
    $N_\text{head}$ & 4 & 8 & 16 & 32 \\
    \midrule
   SPT-64 &  74.3 & 75.2 & \bf 76.0 & 75.9 \\
    \bottomrule
\end{tabular}}
\end{table}

\section{Hierarchical Supervision}
\label{sec:lossablation}

We explore, in \tabref{tab:lossablation}, alternatives to our hierarchical supervision introduced in Section \textcolor{red}{$3.3$} : predicting the most frequent label for $\cP_1$ and the distribution for $\cP_2$.
We use ``\text{freq}-$\cP_i$'' to refer to the prediction of the most frequent label applied  the $\cP_i$ partition. Similarly, ``\text{dist}-$\cP_i$" denotes the prediction of the distribution of labels within each superpoint of the partition $\cP_i$. 

We observe a consistent improvement across all datasets by adding the \text{dist}-$\cP_i$ supervision. 
This illustrates the benefits of supervising higher-level partitions, despite their lower purity.
Moreover, supervising $\cP_1$ with the distribution rather than the most frequent label leads to a further performance drop.
This validates our choice to consider $\cP_1$ superpoints as sufficiently pure to be supervised using their dominant label.

\begin{table}[H]
\caption{\textbf{Ablation on Supervision.} Impact of our hierarchical supervision for each dataset.}
\label{tab:lossablation}
\centering
\small{
\begin{tabular}{@{}lcccc@{}}
    \toprule
    Loss & S3DIS & KITTI & DALES \\
     & 6-Fold & 360 Val & \\
    \midrule
    \text{freq}-$\cP_i$-$\cP_1$ \text{dist}-$\cP_i$-$\cP_2$ & \bf 76.0 & \bf 63.5 & \bf 79.6 \\
    \midrule
    \text{freq}-$\cP_1$    & -0.2 & -0.8 & -0.8 \\
    \text{dist}-$\cP_i$-$\cP_1$    & -0.8 & -1.3 & -0.8 \\
    \bottomrule
\end{tabular}}
\end{table}

\section{Detailed Results}
\label{sec:classwise}

\begin{table*}[t]
\caption{{\bf Class-wise Performance.} Class-wise mIoU across all datasets for our \METHOD. }
\label{tab:classwise} 
\begin{center}
\footnotesize{

    \begin{tabular}{lc*{14}{c}}
        
        \multicolumn{15}{c}{S3DIS Area~5}\\
        Method & mIoU & ceiling & floor & wall & beam & column & window & door & chair & table & bookcase & sofa & board & clutter\\
        \midrule
        PointNet \cite{qi2017pointnet} & 41.1 & 88.8 & 97.3 & 69.8 & \bf 0.1 & 3.9 & 46.3 & 10.8 & 52.6 & 58.9 & 40.3 & 5.9 & 26.4 & 33.2 \\
        SPG \cite{landrieu2018large} & 58.4 & 89.4 & 96.9 & 78.1 & 0.0 & 42.8 & 48.9 & 61.6 & 84.7 & 75.4 & 69.8 & 52.6 & 2.1 & 52.2 \\
        MinkowskiNet \cite{choy20194d} & 65.4 & 91.8 & \bf 98.7 & 86.2 & 0.0 & 34.1 & 48.9 & 62.4 & 81.6 & \bf 89.8 & 47.2 & 74.9 & 74.4 & 58.6 \\
        {SPG + SSP \cite{landrieu2019point}} & 61.7 & 91.9 & 96.7 & 80.8 & 0.0 & 28.8 & 60.3 & 57.2 & 85.5 & 76.4 & 70.5 & 49.1 & 51.6 & 53.3 \\
        KPConv \cite{thomas2019kpconv} & 67.1 & 92.8 & 97.3 & 82.4 & 0.0 & 23.9 & 58.0 & 69.0 & 91.0 & 81.5 & 75.3 & 75.4 & 66.7 & 58.9 \\
        PointTrans.\cite{zhao2021point} & 70.4 & 94.0 & 98.5 & \bf 86.3 & 0.0 & 38.0 & 63.4 & 74.3 & 89.1 & 82.4 & 74.3 & 80.2 & 76.0 & 59.3 \\
        DeepViewAgg \cite{robert2022learning} & 67.2 & 87.2 & 97.3 & 84.3 & 0.0 & 23.4 & \bf 67.6 & 72.6 & 87.8 & 81.0 & 76.4 & 54.9 & \bf 82.4 & 58.7 \\
        Stratified PT \cite{lai2022stratified} & \bf 72.0 & \bf 96.2 & \bf 98.7 & 85.6 & 0.0 & \bf 46.1 & 60.0 & \bf 76.8 & \bf 92.6 & 84.5 & \bf 77.8 & 75.2 & 78.1 & \bf 64.0\\
        \midrule
        \SHORTHAND &  68.9 & 92.6 & 97.7 & 83.5 & \bf 0.2 & 42.0 & 60.6 & 67.1 & 88.8 & 81.0 & 73.2 & \bf 86.0 & 63.1 & 60.0 \\
        \SHORTHANDNANO & 64.9 & 92.4 & 97.1 & 81.6 &  0.0 & 38.2 & 56.4 & 58.6 & 86.3 & 77.3 & 69.6 & 82.5 & 50.5 & 53.4 \\
        \midrule~\\

        \multicolumn{15}{c}{S3DIS 6-FOLD}\\
        \midrule~\\
        PointNet \cite{qi2017pointnet} & 47.6 & 88.0 & 88.7 & 69.3 & 42.4 & 23.1 & 47.5 & 51.6 & 42.0 & 54.1 & 38.2 & 9.6 & 29.4 & 35.2 \\
        SPG \cite{landrieu2018large} & 62.1 & 89.9 & 95.1 & 76.4 & 62.8 & 47.1 & 55.3 & 68.4 & 73.5 & 69.2 & 63.2 & 45.9 & 8.7 & 52.9 \\
        ConvPoint \cite{boulch2020convpoint} & 68.2 & \bf 95.0 & \bf 97.3 & 81.7 & 47.1 & 34.6 & 63.2 & 73.2 & 75.3 & 71.8 & 64.9 & 59.2 & 57.6 & 65.0 \\
        MinkowskiNet \cite{choy20194d,robert2022learning} & 69.5 & 91.2 & 90.6 & 83.0 & 59.8 & 52.3 & 63.2 & 75.7 & 63.2 & 64.0 & 69.0 & 72.1 & 60.1 & 59.2 \\
        {SPG + SSP \cite{landrieu2019point}} & 68.4 & 91.7 & 95.5 & 80.8 & 62.2 & 54.9 & 58.8 & 68.4 & 78.4 & 69.2 & 64.3 & 52.0 & 54.2 & 59.2 \\
        KPConv \cite{thomas2019kpconv} & 70.6 & 93.6 & 92.4 & 83.1 & 63.9 & 54.3 & 66.1 & 76.6 & 57.8 & 64.0 & 69.3 & 74.9 & 61.3 & 60.3 \\
        DeepViewAgg \cite{robert2022learning} & 74.7 & 90.0 & 96.1 & \bf 85.1 & 66.9 & 56.3 & \bf 71.9 & \bf 78.9 & 79.7 & 73.9 & \bf 69.4 & 61.1 & \bf 75.0 & \bf 65.9 \\
        \midrule
        \SHORTHAND & \bf 76.0 & 93.9 & 96.3 & 84.3 & \bf 71.4 & \bf 61.3 & 70.1 & 78.2 & \bf 84.6 & \bf 74.1 & 67.8 & \bf 77.1 & 63.6 & 65.0 \\
        \SHORTHANDNANO & 70.8 & 93.1 & 96.0 & 80.9 & 68.4 & 54.0 & 62.2 & 71.3 & 76.3 & 70.8 & 63.3 & 74.3 & 51.9 & 57.6 \\
        \midrule
    \end{tabular}\\~\\~\\

    \begin{tabular}{lc*{16}{c}}
        \multicolumn{17}{c}{KITTI-360 Val}\\
        Method & mIoU & \rotatebox{90}{road} & \rotatebox{90}{sidewalk} & \rotatebox{90}{building} & \rotatebox{90}{wall} & \rotatebox{90}{fence} & \rotatebox{90}{pole} & \rotatebox{90}{traffic lig.} & \rotatebox{90}{traffic sig.} & \rotatebox{90}{vegetation} & \rotatebox{90}{terrain} & \rotatebox{90}{person} & \rotatebox{90}{car} & \rotatebox{90}{truck} & \rotatebox{90}{motorcycle} & \rotatebox{90}{bicycle}\\
        \midrule
        MinkowskiNet \cite{choy20194d,robert2022learning} & 54.2 & 90.6 & 74.4 & 84.5 & 45.3 & 42.9 & 52.7 & 0.5 & 38.6 & 87.6 & 70.3 & 26.9 & 87.3 & 66.0 & 28.2 & 17.2 \\
        DeepViewAgg \cite{robert2022learning} & 57.8 & \bf 93.5 & 77.5 & 89.3 & 53.5 & \bf 47.1 & \bf 55.6 & 18.0 & 44.5 & \bf 91.8 & 71.8 & 40.2 & 87.8 & 30.8 & 39.6 & \bf 26.1 \\
        \midrule
        \SHORTHAND  & \bf 63.5 & 93.3 & \bf 79.3 & \bf 90.8 & \bf 56.2 & 45.7 & 52.8 & \bf 20.4 & \bf 51.4 & 89.8 & \bf 73.6 & \bf 61.6 & \bf 95.1 & \bf 79.0 & \bf 53.1 & 10.9 \\
        \SHORTHANDNANO & 57.2 & 91.7 & 74.7 & 87.8 & 49.3 & 38.8 & 49.0 & 12.2 & 39.2 & 88.0 & 69.5 & 39.9 & 94.2 & 80.1 & 33.7 & 10.4 \\
        \midrule
    \end{tabular}\\~\\~\\
    
    \begin{tabular}{lc*{10}{c}}
        \multicolumn{11}{c}{DALES}\\
        Method & mIoU & ground & vegetation & car & truck & power line & fence & pole & building \\
        \midrule
        PointNet++ \cite{qi2017pointnetpp}   & 68.3 & 94.1 & 91.2 & 75.4 & 30.3 & 79.9 & 46.2 & 40.0 & 89.1 \\
        ConvPoint \cite{boulch2020convpoint} & 67.4 & 96.9 & 91.9 & 75.5 & 21.7 & 86.7 & 29.6 & 40.3 & 96.3 \\
        SPG \cite{landrieu2018large}         & 60.6 & 94.7 & 87.9 & 62.9 & 18.7 & 65.2 & 33.6 & 28.5 & 93.4 \\
        PointCNN \cite{li2018pointcnn}       & 58.4 & \bf 97.5 & 91.7 & 40.6 & 40.8 & 26.7 & 52.6 & 57.6 & 95.7 \\
        KPConv \cite{thomas2019kpconv}       & \bf 81.1 & 97.1 & \bf 94.1 & 85.3 & 41.9 & \bf 95.5 & \bf 63.5 & \bf 75.0 & 96.6 \\
        \midrule
        \SHORTHAND & 79.6 & 96.7 & 93.1 & \bf 86.1 & \bf 52.4 & 94.0 & 52.7 & 65.3 & \bf 96.7 \\
        \SHORTHANDNANO & 75.2 & 96.5 & 92.6 & 78.1 & 35.8 & 92.1 & 50.8 & 59.9 & 96.0 \\
        \midrule
    \end{tabular}\\
    
}\end{center}
\end{table*}

We report in \tabref{tab:classwise} the class-wise performance across all datasets for \SHORTHAND and other methods for which this information was available. As previously stated, \SHORTHAND performs close to state-of-the-art methods on all datasets, while being significantly smaller and faster to train. 
By design, superpoint-based methods can capture long-range interactions and their predictions are more spatially regular than point-based approaches. This may explain the performance of \SHORTHAND on S3DIS, which encompasses large, geometrically homogeneous objects or whose identification requires long-range context understanding, such as ceiling, floor, columns, and windows.
For all datasets, results show that some progress could be made in analyzing smaller objects with intricate geometries. This suggests that a more powerful point-level encoding may be beneficial.

\balance

\begin{figure*}[!b]
    \centering
    \begin{tabular}{c}
        \toprule
        \large{S3DIS}
        \\\midrule
        \begin{tabular}{@{}rlrlrlrlrl@{}}
            \definecolor{tempcolor}{rgb}{0.91,0.90,0.41}
            \tikz \fill[fill=tempcolor, scale=0.3, draw=black] (0, 0) rectangle (1.2, 1.2);
            & \small{ceiling} 
            &
            \definecolor{tempcolor}{rgb}{.37,0.61,0.77}
            \tikz \fill[fill=tempcolor, scale=0.3, draw=black] (0, 0) rectangle (1.2, 1.2); 
            & \small{floor}
            &
            \definecolor{tempcolor}{rgb}{0.70,0.45,0.31}
            \tikz \fill[fill=tempcolor, scale=0.3, draw=black] (0, 0) rectangle (1.2, 1.2); 
            & \small{wall}
            &
            \definecolor{tempcolor}{rgb}{0.95,.58,0.51}
            \tikz \fill[fill=tempcolor, scale=0.3, draw=black] (0, 0) rectangle (1.2, 1.2);
            & \small{beam} &
            \definecolor{tempcolor}{rgb}{0.31,0.63,.58}
            \tikz \fill[fill=tempcolor, scale=0.3, draw=black] (0, 0) rectangle (1.2, 1.2);
            & \small{column} 
            \\
            \definecolor{tempcolor}{rgb}{0.30,0.68,.32}
            \tikz \fill[fill=tempcolor, scale=0.3, draw=black] (0, 0) rectangle (1.2, 1.2); 
            & \small{window}
            &
            \definecolor{tempcolor}{rgb}{.42,0.52,0.29}
            \tikz \fill[fill=tempcolor, scale=0.3, draw=black] (0, 0) rectangle (1.2, 1.2); 
            & \small{door}
            &
            \definecolor{tempcolor}{rgb}{.16,0.19,0.39}
            \tikz \fill[fill=tempcolor, scale=0.3, draw=black] (0, 0) rectangle (1.2, 1.2);
            & \small{chair}
            &
            \definecolor{tempcolor}{rgb}{.30,0.30,0.30}
            \tikz \fill[fill=tempcolor, scale=0.3, draw=black] (0, 0) rectangle (1.2, 1.2);
            & \small{table} 
            &
            \definecolor{tempcolor}{rgb}{.88,0.20,0.20}
            \tikz \fill[fill=tempcolor, scale=0.3, draw=black] (0, 0) rectangle (1.2, 1.2); 
            & \small{bookcase}
            \\
            \definecolor{tempcolor}{rgb}{0.35,.18,0.37}
            \tikz \fill[fill=tempcolor, scale=0.3, draw=black] (0, 0) rectangle (1.2, 1.2); 
            & \small{sofa}
            &
            \definecolor{tempcolor}{rgb}{.32,0.43,.45}
            \tikz \fill[fill=tempcolor, scale=0.3, draw=black] (0, 0) rectangle (1.2, 1.2);
            & \small{board}
            &
            \definecolor{tempcolor}{rgb}{0.91,0.91,.90}
            \tikz \fill[fill=tempcolor, scale=0.3, draw=black] (0, 0) rectangle (1.2, 1.2);
            & \small{clutter} 
            &
            \definecolor{tempcolor}{rgb}{0.,0.,0}
            \tikz \fill[fill=tempcolor, scale=0.3, draw=black] (0, 0) rectangle (1.2, 1.2);
            & \small{unlabeled} 
        \end{tabular}
        \\
         \toprule
        \large{KITTI-360}
        \\\midrule
        \begin{tabular}{rlrlrlrlrl}
            \definecolor{tempcolor}{rgb}{0.50, 0.25, 0.50}
            \tikz \fill[fill=tempcolor, scale=0.3, draw=black] (0, 0) rectangle (1.2, 1.2);
            & \small{road} 
            &
            \definecolor{tempcolor}{rgb}{0.95, 0.13, 0.90}
            \tikz \fill[fill=tempcolor, scale=0.3, draw=black] (0, 0) rectangle (1.2, 1.2);
            & \small{sidewalk} 
            &
            \definecolor{tempcolor}{rgb}{0.27, 0.27, 0.27}
            \tikz \fill[fill=tempcolor, scale=0.3, draw=black] (0, 0) rectangle (1.2, 1.2);
            & \small{building} 
            &
            \definecolor{tempcolor}{rgb}{0.4 , 0.4 , 0.61}
            \tikz \fill[fill=tempcolor, scale=0.3, draw=black] (0, 0) rectangle (1.2, 1.2);
            & \small{wall} 
            &
            \definecolor{tempcolor}{rgb}{0.74, 0.6 , 0.6 }
            \tikz \fill[fill=tempcolor, scale=0.3, draw=black] (0, 0) rectangle (1.2, 1.2);
            & \small{fence} 
            \\
            \definecolor{tempcolor}{rgb}{0.6 , 0.6 , 0.6 }
            \tikz \fill[fill=tempcolor, scale=0.3, draw=black] (0, 0) rectangle (1.2, 1.2);
            & \small{pole} 
            &
            \definecolor{tempcolor}{rgb}{0.98, 0.66, 0.11}
            \tikz \fill[fill=tempcolor, scale=0.3, draw=black] (0, 0) rectangle (1.2, 1.2);
            & \small{traffic light} 
            &
            \definecolor{tempcolor}{rgb}{0.86, 0.86, 0.  }
            \tikz \fill[fill=tempcolor, scale=0.3, draw=black] (0, 0) rectangle (1.2, 1.2);
            & \small{traffic sign} 
            &
            \definecolor{tempcolor}{rgb}{0.41, 0.55, 0.13}
            \tikz \fill[fill=tempcolor, scale=0.3, draw=black] (0, 0) rectangle (1.2, 1.2);
            & \small{vegetation} 
            &
            \definecolor{tempcolor}{rgb}{0.59, 0.98, 0.59}
            \tikz \fill[fill=tempcolor, scale=0.3, draw=black] (0, 0) rectangle (1.2, 1.2);
            & \small{terrain} 
            \\
            \definecolor{tempcolor}{rgb}{0.86, 0.07, 0.23}
            \tikz \fill[fill=tempcolor, scale=0.3, draw=black] (0, 0) rectangle (1.2, 1.2);
            & \small{person} 
            &
            \definecolor{tempcolor}{rgb}{0.  , 0.  , 0.55}
            \tikz \fill[fill=tempcolor, scale=0.3, draw=black] (0, 0) rectangle (1.2, 1.2);
            & \small{car} 
            &
            \definecolor{tempcolor}{rgb}{0.  , 0.  , 0.27}
            \tikz \fill[fill=tempcolor, scale=0.3, draw=black] (0, 0) rectangle (1.2, 1.2);
            & \small{truck} 
            &
            \definecolor{tempcolor}{rgb}{0.  , 0.  , 0.90}
            \tikz \fill[fill=tempcolor, scale=0.3, draw=black] (0, 0) rectangle (1.2, 1.2);
            & \small{motorcycle} 
            &
            \definecolor{tempcolor}{rgb}{0.46, 0.04, 0.12}
            \tikz \fill[fill=tempcolor, scale=0.3, draw=black] (0, 0) rectangle (1.2, 1.2);
            & \small{bicycle} 
            \\
            \definecolor{tempcolor}{rgb}{0.  , 0.  , 0.  }
            \tikz \fill[fill=tempcolor, scale=0.3, draw=black] (0, 0) rectangle (1.2, 1.2);
            & \small{ignored} 
            &
        \end{tabular}\\
         \toprule
            \large{DALES}
        \\\midrule
        \begin{tabular}{rlrlrlrlrl}
            \definecolor{tempcolor}{rgb}{0.95, 0.84, 0.67}
            \tikz \fill[fill=tempcolor, scale=0.3, draw=black] (0, 0) rectangle (1.2, 1.2);
            & \small{ground} 
            &
            \definecolor{tempcolor}{rgb}{0.27, 0.45, 0.26}
            \tikz \fill[fill=tempcolor, scale=0.3, draw=black] (0, 0) rectangle (1.2, 1.2); 
            & \small{vegetation}
            &
            \definecolor{tempcolor}{rgb}{0.91, 0.20, 0.94}
            \tikz \fill[fill=tempcolor, scale=0.3, draw=black] (0, 0) rectangle (1.2, 1.2); 
            & \small{car}
            &
            \definecolor{tempcolor}{rgb}{0.95, 0.93, 0.}
            \tikz \fill[fill=tempcolor, scale=0.3, draw=black] (0, 0) rectangle (1.2, 1.2);
            & \small{truck} &
            \definecolor{tempcolor}{rgb}{0.75, 0.6,0.6  }
            \tikz \fill[fill=tempcolor, scale=0.3, draw=black] (0, 0) rectangle (1.2, 1.2);
            & \small{power line} 
            \\
            \definecolor{tempcolor}{rgb}{0., 0.91,  0.04}
            \tikz \fill[fill=tempcolor, scale=0.3, draw=black] (0, 0) rectangle (1.2, 1.2); 
            & \small{fence}
            &
            \definecolor{tempcolor}{rgb}{0.94,  0.45, 0.}
            \tikz \fill[fill=tempcolor, scale=0.3, draw=black] (0, 0) rectangle (1.2, 1.2); 
            & \small{pole}
            &
            \definecolor{tempcolor}{rgb}{0.84, 0.26, 0.21}
            \tikz \fill[fill=tempcolor, scale=0.3, draw=black] (0, 0) rectangle (1.2, 1.2);
            & \small{building}
            &
            \definecolor{tempcolor}{rgb}{0., 0.03, 0.45}
            \tikz \fill[fill=tempcolor, scale=0.3, draw=black] (0, 0) rectangle (1.2, 1.2);
            & \small{unknown}
            
        \end{tabular}\\
        \midrule~\\
    \end{tabular}
    \caption{{\bf Colormaps.} }
    \label{fig:colormaps}
\end{figure*}

}{}

\end{document}